\documentclass[%
reprint,
amsmath,amssymb,
aps,
]{revtex4-2}
\usepackage{amssymb}
\usepackage{graphicx}
\usepackage{dcolumn}
\usepackage{bm}

\usepackage{braket}
\usepackage{amsmath}
\usepackage{algorithm}
\usepackage{algpseudocode}
\usepackage{amsfonts}

\begin{document}
	
	\preprint{APS/123-QED}
	
	\title{Multi-Omic and Quantum Machine Learning Integration for Lung Subtypes Classification}

	\author{Mandeep Kaur Saggi$^{1}$}
	\email{drmandeepsaggi@gmail.com, msaggi@purdue.edu}
	
	\author{Amandeep Singh Bhatia$^{1,2}$}		
	\email{drasinghbhatia@gmail.com}
	
	\author{Mensah Isaiah$^{3}$}
	
	\author{Humaira Gowher$^{3}$}
	\email{hgowher@purdue.edu}
	
	\author{Sabre Kais$^{2}$}
	\email{skais@ncsu.edu}
	
    \email{Corresponding author1: skais@ncsu.edu}
    \email{Corresponding author2: hgowher@purdue.edu}
   	

	\affiliation{$^{1}$ Department of Chemistry and Purdue Quantum Science and Engineering Institute, Purdue University, IN, USA}		
	\affiliation{$^{2}$ Department of Electrical and Computer Engineering, North Carolina State University, NC, USA}
	\affiliation{$^{3}$ Department of Biochemistry, Purdue University, IN, USA}

\begin{abstract}

Quantum Machine Learning (QML) is a red-hot field that brings novel discoveries and exciting opportunities to resolve, speed up, or refine the analysis of a wide range of computational problems. In the realm of biomedical research and personalized medicine,  the significance of multi-omics integration lies in its ability to provide a thorough and holistic comprehension of complex biological systems. This technology links fundamental research to clinical practice.  The insights gained from integrated omics data can be translated into clinical tools for diagnosis, prognosis, and treatment planning. The fusion of quantum computing and machine learning holds promise for unraveling complex patterns within multi-omics datasets, providing unprecedented insights into the molecular landscape of lung cancer. Due to the heterogeneity, complexity, and high dimensionality of multi-omic cancer data, characterized by the vast number of features (such as gene expression, micro-RNA, and DNA methylation) relative to the limited number of lung cancer patient samples, our prime motivation for this paper is the integration of multi-omic data, unique feature selection, and diagnostic classification of lung subtypes: lung squamous cell carcinoma (LUSC-I) and lung adenocarcinoma (LUAD-II) using quantum machine learning. We developed a method for finding best differentiating features between LUAD and LUSC datasets, which has the potential for biomarker discovery.
In this paper, we show the efficacy of Quantum Neural Networks (QNNs) with three dimensions of feature encoding for the classification of multi-Omics human lung data from The Cancer Genome Atlas, comparing the framework to a variety of classical machine learning methods. Our results indicate that the Multi-omic Quantum Machine Learning Lung Subtype Classification (MQML-LungSC) framework offers superior classification performance with smaller training datasets as significant and non-significant based on p-value, thus providing compelling empirical evidence for the potential future application of unconventional computing approaches in the biomedical sciences.  In comparing the performance of our proposed models based on the number of encoded features (256, 64, 32), it is evident that the model with 256 encoded features exhibits superior results across several metrics. It achieves a training accuracy of 0.95 and a testing accuracy of 0.90, which are higher than those of the models with 64 and 32 encoded features. Specifically, the model with 64 encoded features scores 0.92 for training accuracy and 0.86 for testing accuracy, while the model with 32 encoded features scores 0.88 for training accuracy and 0.85 for testing accuracy. The analyses of large-scale molecular data are beneficial for many aspects of oncology research, including the classification of possible subtypes, stages, and grades of cancer. 

	\end{abstract}
	
	\maketitle


\section{INTRODUCTION}
\noindent Quantum machine learning is pioneering a new era in computational biology, unleashing powerful tools that redefine the possibilities for tackling intricate biological challenges. The analyses of large-scale molecular data are beneficial for many aspects of oncology research, including the classification of possible subtypes, stages, and grades of cancer. Several approaches have been proposed to train networks using quantum computing technology more accurately, robustly, and efficiently \cite{Link1} \cite{1}. However, hybrid strategies have recently emerged given the current limitations of quantum machines and the constrained number of available qubits. These strategies leverage existing technology to achieve practical and usable solutions \cite{2}.
Quantum-enhanced AI and machine learning methods are gaining attention as promising solutions to medical challenges. Recent advancements in quantum computing and quantum AI have demonstrated their wide-ranging applicability in healthcare. These methods have shown promise in various healthcare and drug-discovery domains and predicting ADME-Tox properties in drug discovery \cite{3} \cite{4} \cite{5}, including rapid genome analysis \cite{6} and sequencing \cite{7}, disease detection \cite{8}, reference crop evapotranspiration classification \cite{9}, for chemistry and electronic structure calculations \cite{30}, \cite{31} \cite{32} \cite{33}. 
Moreover, quantum computing models play a significant role in predicting gene mutations critical for the pathogenesis and diagnosis of specific cancer types, such as the Glioma Tumor Classification \cite{10}.
Cancer sub-type classification is crucial for understanding cancer pathogenesis and developing targeted treatments that can benefit patients the most \cite{11}. 
Lung cancer is the most commonly diagnosed malignant tumor and is a leading cause of cancer-associated mortality. It is the second highest cause of new cancer cases in both genders in the United States and is the second leading cause of cancer deaths in females globally. The most common subtypes of lung cancers are lung squamous cell carcinoma (LUSC) and lung adenocarcinoma (LUAD), classified together as non-small cell lung cancer (NSCLC). 
The GDC-TCGA dataset provides diverse omic data crucial for cancer research and subtype classification. Specifically, for LUSC and LUAD cancer subtype the dataset includes
\textbf{DNA Methylation (DNAme):} This dataset, derived from the Illumina Human Methylation 450 platform, provides beta values indicating DNA methylation levels across various genes. This data helps identify methylation patterns that are specific to LUAD and LUSC subtypes, revealing regulatory changes associated with each cancer type (which means DNA methylation features are present (indicative of significant regulatory changes) or absent (no significant changes) in each subtype to understand their role in tumor biology). Each sample in this dataset represents tumor tissue from an individual patient, with data formatted as rows of gene identifiers and columns of sample beta values. 
\textbf{RNA Sequencing (RNA-seq)}: Measures gene expression levels by analyzing RNA from tumor tissues using the HTSeq platform, helping to identify genes with differential expression between LUAD and LUSC subtypes. The RNA-seq data are presented as log2(count + 1) values, reflecting the relative expression of genes. 
\textbf{MicroRNA Sequencing (miRNA-seq)}: This dataset quantifies microRNA expression levels using stem-loop expression technology. The miRNA-seq data are also transformed into log2(RPM + 1) values, providing insights into the regulatory roles of microRNAs in LUAD and LUSC. Each sample represents tumor tissue from an individual patient, and the data are organized into rows of miRNA identifiers and columns of expression values.
For LUAD vs. LUSC subtype classification, the dataset includes primary tumor and solid tissue normal samples from which these multiomic features are derived. This approach allows for a thorough examination of molecular differences between the lung subtypes, and enhancing diagnostic accuracy.

Traditional studies often analyze individual omic features in isolation, focusing on discrete datasets such as DNA Methylation patterns \cite{42}, gene expression profiles \cite{43}, or microRNA levels. In these contexts, a "sample" refers to a biological specimen collected from a patient, such as a piece of tumor tissue or a blood sample. Each sample provides specific molecular data representing that patient's unique biological characteristics. This isolated approach, which examines data from each sample separately, can miss valuable insights from the interactions and correlations between molecular features across different samples.

In contrast, our MQML-LungSC framework integrates DNA-Methylation, RNA-seq, and miRNA-seq data to explore the interconnectedness of molecular features across a wide range of tumor samples. By examining the relationships between gene expression levels, methylation patterns, and microRNA profiles both within individual samples and across a population of samples, we can identify complex interactions and patterns that differentiate LUAD from LUSC. This integrated analysis allows for a more comprehensive understanding of the molecular mechanisms underlying these lung subtypes, potentially leading to more accurate classification and the discovery of significant features.

In Fig. 1, the development of lung cancer is illustrated by comparing normal cells to tumor cells. However, recent studies have suggested that LUAD and LUSC should be classified and treated as different cancers \cite{44}. Previous studies have utilized traditional feature selection and machine learning methods for cancer diagnosis, detection, and classification, but few have extended them to study potential features and biological pathways to discriminate between LUAD and LUSC \cite{12}. To improve cancer classification accuracy, novel machine learning and feature selection methods have been developed. However, few studies have used overlapping features from different methods for classification, gene expression analysis, and molecular features \cite{13} \cite{14}.

This work proposes a novel lung sub-type classification method that integrates classical feature selection techniques with a quantum classifier. This hybrid approach aims to enhance the accuracy and robustness of sub-type classification, thereby potentially contributing to the development of more effective cancer therapies. We propose a set of hybrid quantum computing and advanced machine learning approaches to apply machine learning on small datasets, such as (primary tumor and normal) sample types in The Cancer Genome Atlas (TCGA). These approaches aim to address the well-known "big n, small m" problem in multi-cancer analysis (Lung subtype diagnosis or classification), including  (LUSC) and (LUAD). 

The cancer datasets named (LUSC) and (LUAD) are taken from the Multi-Omics Cancer Benchmark TCGA  (http://cancergenome.nih.gov/) Pre-processed Data, publicly available at the Multi-Omics Cancer Benchmark repository. For our study we employed omics datasets for transcriptome profiling by RNA-seq and miRNA-seq (micro-RNA) DNA methylation analysis by Methylation Array, , and  associated clinical outcomes using associated patient information. These multi-omics data were extracted for a case study of LUSC and LUAD. To the best of our knowledge, this study is the first to utilize GDC-TCGA data within a Quantum-Classical framework using combination of omics data (DNA methylation, transciptome profiling from RNA-seq, and miRNA-seq, and clinical outcomes) data for lung subtype classification and molecular features identification. 

\noindent \textbf{Problem Formulation and Motivation:}
Accurate diagnostic classification of subtypes can greatly help physicians to choose surveillance and treatment strategies for patients. Following the explosive growth of huge amounts of biological data, the shift from traditional bio-statistical methods to computer-aided means has made machine-learning methods an integral part of today’s cancer prognosis and diagnosis.  Integration of multi-omics data allows more advanced comprehensive and systematic analysis of biological changes, providing a new biomarkers for the early diagnosis of diseases. However, there are several challenges in integrating and analyzing cancer multi-omics data on a large scale. TGCA database consists of several cancer omics datasets, including transciptome and proteome profiling, copy number variable, somatic structural, and DNA methylation. Mostly, multi-omics is characterized by high noise, high multidimensionality, and multidimensional heterogeneity, such as (big “n” and small “m”) affecting the efficiency of classification tasks, where “n” refers to the number of features/genes and “m” samples ((i) DNA: 503 samples and 485,577 genes, (ii) RNA: 585 samples and 60,488 genes, (iii) miRNA: 564 samples and 1,881 genes) for LUAD. (i) DNA: 412 samples and 485,577 genes, (ii) RNA: 550 samples and 60,488 genes, (iii) miRNA: 523 samples and 1,881 genes) for LUSC. 

\noindent \textbf{Main contributions:}
This study introduces a hybrid quantum classification model for distinguishing between LUSC-I tumor and LUAD-II tumor subtypes of lung datasets. In the classical section, we aim to identify the most important combination of multi-omic molecular markers using the feature selection technique. The multi-omic integrated features are encoded in the quantum section with different dimensions to find the best combination and hit list using quantum-classical model weights for tumor classification in LUAD.

To summarize, this paper makes the following contributions:
\begin{itemize}
	\item  Developed a pipeline for data pre-processing, feature engineering based on p-value t-test employs significant and non-significant using high dimensions of omic datasets which include  DNAme, miRNA-seq and RNA-seq in LUAD and LUSC subtype dataset.  
	\item  Proposed a classical machine learning and filter methods for feature selection using random forest, mutual information, chi-square, and PCA on single omics and select top 85-85-86 features from each omic for further integration and selection of clinical attributes such as age, gender, pathological stage, survival status etc using GDC TCGA dataset.
	\item Developed the quantum neural network algorithm to encode the 32 features, 64 features, and 256 features with amplitude encoding for subtype-I and subtype-II classification and find the top-hit features with the qnn layer/dense layer for further visualization plots such as violin, dot plot, heatmap clustering, and PCA.   
	\item Evaluated the accuracy and strength of the proposed quantum machine learning framework using three encoding structures of dimensions with fewer parameters and many epochs. 
\end{itemize}

The remainder of the article is organized as follows. Section II presents the Materials and Methods, including the dataset, processes, and methods. Section III outlines the proposed methodology of our framework, detailing the implementation process and describing the multi-omic quantum machine learning approach. Section IV provides the results and performance analysis, including experimental results on multi-omic integrations across three dimensions, comparisons with classical classifiers, and single-omic datasets. Section V discusses the findings and outcomes. Finally, Section VI concludes the paper and explores future research directions.

\section{MATERIAL AND METHODS}
This section provides a comprehensive overview of the methods utilized in our study. We describe the feature engineering process, feature selection methods, and proposed quantum neural network model for diagnosing lung dataset subtypes LUAD and LUSC. Each subsection below elaborates on these methods in detail. 

\subsection{\textbf{Study Population}}
TCGA is a cancer multi-omics database generated by the National Institutes of Health Our proposed framework comprises four types of omics datasets: Gene expression, miRNA expression, DNA methylation, and Clinical patient information of lung squamous cell carcinoma. The number of samples acquired for selecting features that distinguish between LUSC and LUAD datasets is presented in Fig 1 and well detailed in Algorithm 1.

\begin{figure*}	[!ht]
	\centering
	\includegraphics[scale=0.38]{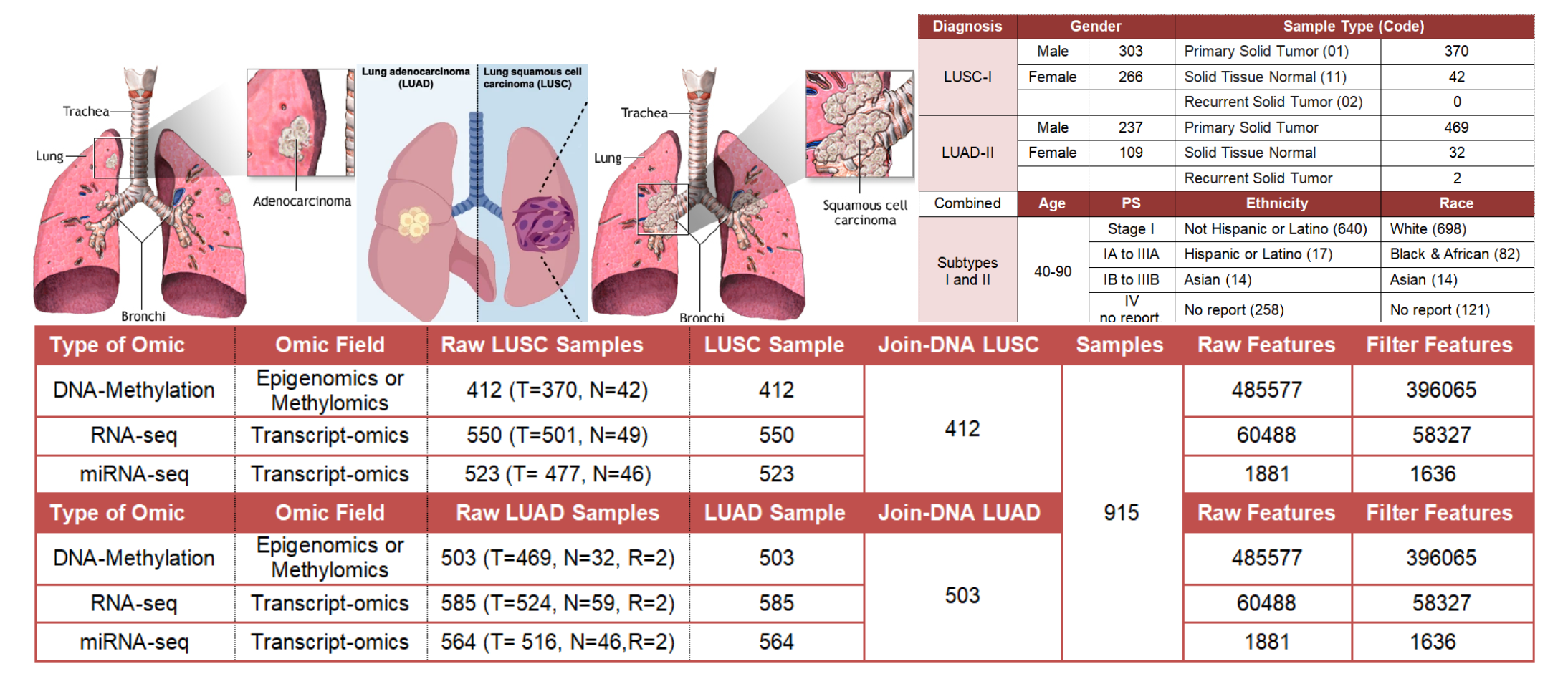}
	\caption{\textbf{Summary of Multi-Omic Modalities and Clinical information} (a) \textbf{Development of Non-small cell lung cancer (NSCLC) lung cancer}. Overview of Non-Small Cell Lung Cancer (NSCLC) subtypes, highlighting the three main subtypes: adenocarcinoma (LUAD), squamous cell carcinoma (LUSC), and large cell carcinoma. This study focuses on two NSCLC subtypes, LUAD and LUSC. (b) Summary of Clinical information of subtypes diagnosis class with gender, sample type. Also combined subtypes patient’s age, Pathological stages, ethnicity, and race (c) information from GDC-TCGA with each entry indicating  lung dataset subtype-I (LUSC) and subtype II (LUAD) 915 patient samples and high dimensional features of each omic}
\end{figure*}

The original dataset comprised genomic data from various modalities, including DNAme, miRNA-seq, and RNA-seq, obtained from a cohort of patients with (LUAD) and (LUSC) lung. Specifically, the DNAme array dataset consisted from 503 patients with 485,577 features; the miRNA dataset contained samples from 564 patients with 1,881 features, and the RNA dataset encompassed samples from 585 patients with 60,488 features of the lung dataset. A summary of their clinical information is provided in Fig 1, with more comprehensive details available on the GDC-TCGA website. As described in Algorithm 1, We have selected three omics modalities for this study, and a summary of clinical samples is shown in Fig 1. The original dataset comprised genomic data from various modalities, including DNAme, miRNA-seq, and RNA-seq, obtained from a cohort of patients with (LUAD) and (LUSC) lung. Specifically, the DNAme array dataset consisted from 503 patients with 485,577 features; the miRNA dataset contained samples from 564 patients with 1,881 features, and the RNA dataset encompassed samples from 585 patients with 60,488 features of the lung dataset. A summary of their clinical information is provided in Fig 1, with more comprehensive details available on the GDC-TCGA website. As described in Algorithm 1, We have selected three omics modalities for this study, and a summary of clinical samples is shown in Fig 1.

\subsection{\textbf{Data Loading and Data Pre-processing}}
In the First phase, MQML takes any number of omic measures such as genomic, epigenomic, and transcriptomic datasets as input. Due to the inherent complexity and size of the dataset, pre-processing steps were necessary to ensure data quality and reduce computational burden. The LUAD and LUSC, Lung datasets are acquired from the Multi-Omics Cancer Benchmark GDC-TCGA Pre-processed Data, It consists of four types of omics datasets for a case study of lung squamous cell carcinoma.: RNA-seq, miRNA-seq, DNAme, and Clinical patient information.    In the second phase, each omic has several gene features column-wise and patient samples row-wise. The raw data of each Omic1$_{Subtype-I}$ and Omic1$_{Subtype-II}$ is combined column-wise, with patient samples row-wise. Then, features that contain a sum of 0 are removed. To extract and select the survival/clinical attributes of both subtypes, combine the survival and clinical attributes based on sample type, i.e., diagnostic subtype-I and subtype-II.

\subsection{ \textbf{Feature Engineering}}

Then, the selected clinical samples of subtype-I and subtype-II are used to split each omic into two parts, i.e., Omic1.1$_{Subtype-I}$ and Omic1.2$_{Subtype-II}$. Each omic data is processed through a feature engineering process, including a statistical t-test, to determine if there is a significant difference between the means of the two groups.

This process involves analyzing two omic data frames, Omic1.1$_{Subtype-I}$ and Omic1.2$_{Subtype-I}$, to compare their mean values and compute statistical significance. First, the mean values for each column in both data frames are calculated. Then, a t-test is conducted for each column to determine the p-values, which assess the statistical significance of the differences between the two datasets. The mean values and p-values are then appended to their respective data frames. Finally, the updated data frames contain the original data mean values and p-values.
This process integrates multiple omic datasets using minimum patient samples of Omic (Omic$_{1}$, Omic$_{2}$, and Omic$_{3}$) for the subtypes separately using a join operation. The integrated dataset combines each Omic1-2-3$_{Subtype-I}$ and Omic1-2-3$_{Subtype-II}$ mean values and p-values at the bottom of the row. Then, the entire dataset is sorted based on p-values for each Omic1-3$_{Subtype-I}$ and Omic1-3$_{Subtype-II}$ to identify the most and least significant values in the data frame. Common significance levels include \(\alpha = 0.05\) (most commonly used, indicating a 5\% risk of a Type I error), \(\alpha = 0.01\) (more stringent, indicating a 1\% risk), and \(\alpha = 0.10\) (less stringent, indicating a 10\% risk). Then, each omic combined Omic1-2-3$_{Subtype}$ data frame is isolated into independent Omic1$_{Subtype}$, Omic2$_{Subtype}$, and Omic3$_{Subtype}$ data frame datasets for data analysis steps. Finally, the prepared clinical and survival attributes are combined with each Omic$_{Subtype}$ as shown in Fig. 2.
The \textbf{t-test }is a statistical method used to determine if there is a significant difference between the means of two groups. In this context, it analyzes multi-omic data from two lung dataset subtypes, LUSC and LUAD. By calculating the t-statistic, we can identify significant features that differentiate the two dataset subtypes. The t-statistic is calculated as:

\begin{equation}
	t_{\text{statistic}} = \frac{\bar{X}_{\text{LUSC}} - \bar{X}_{\text{LUAD}}}{s_{\text{LUSC}} + s_{\text{LUAD}}}
\end{equation}
where \(\bar{X}_{\text{LUSC}}\) is the mean of the LUSC group, \(\bar{X}_{\text{LUAD}}\) is the mean of the LUAD group, \(s_{\text{LUSC}}\) is the standard deviation of the LUSC group, and \(s_{\text{LUAD}}\) is the standard deviation of the LUAD group.

\begin{figure}[!ht]
	\centering
	\includegraphics[scale=0.5]{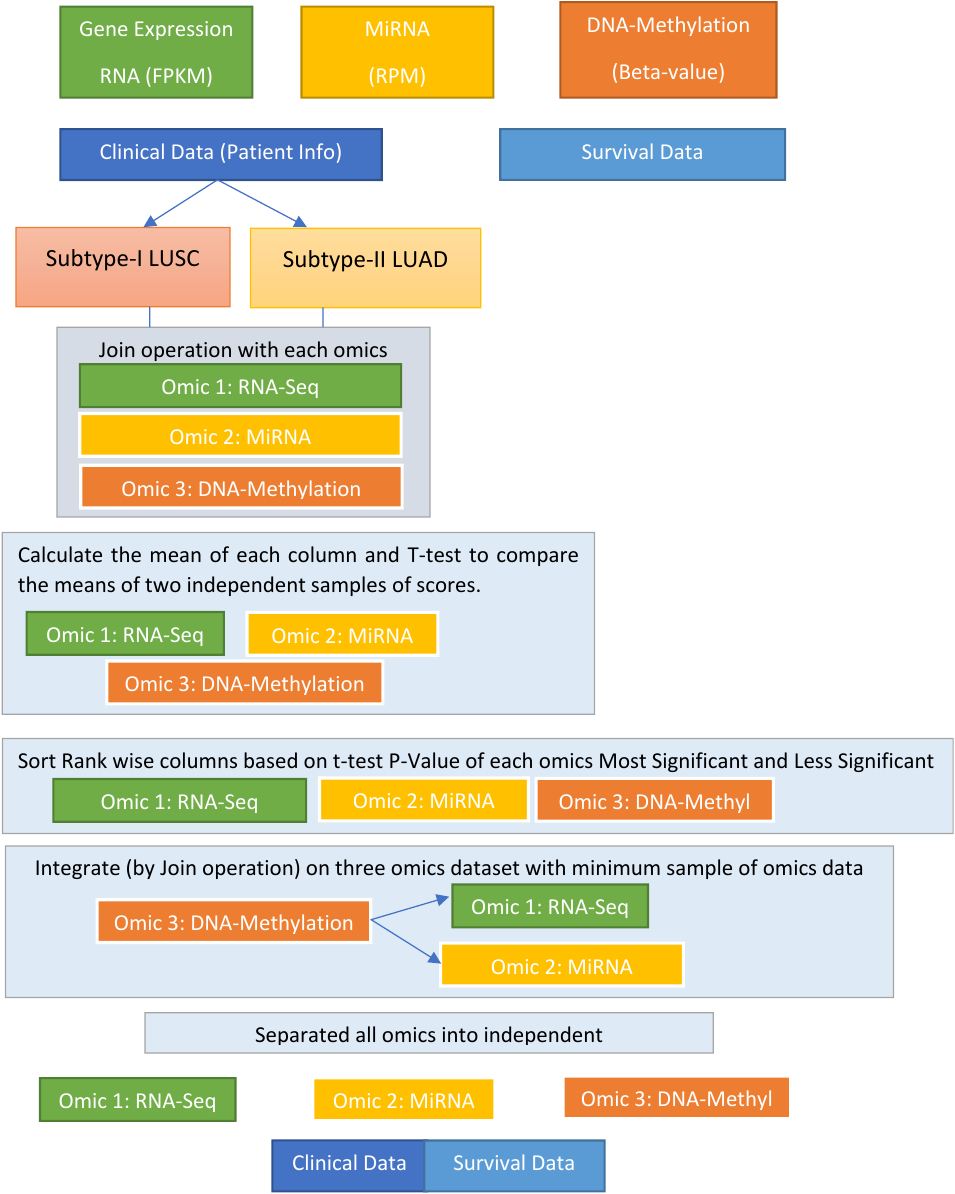}
	\caption{\textbf{Flowchart of Data Engineering process}. This diagram illustrates the steps involved in analyzing and integrating multiple omic datasets, including mean value calculation, statistical t-tests, integration of datasets, and combination with clinical and survival attributes. }
\end{figure}

\subsection{ \textbf{Feature Selection}}
In the third phase, each omic is divided into three subsets of samples: Omic\(_{S1}\), Omic\(_{S2}\), and Omic\(_{S3}\), based on p-values as shown in Table IV and Algorithm 1. We employed three steps in the feature selection process. 
In the first step, we determined the important features using four methods (two filter methods, one supervised method, and one unsupervised method) with the Select K Best function. Using a Venn diagram, we selected the unique features by identifying common and uncommon features. In the second step, we employed a random forest to classify the selected features from each selection process. We then combined the features based on AUC-ROC analysis to choose the best features with an AUC greater than 0.80. In the third step, each omic's combined and reduced subset is further processed to reduce the features using hierarchical clustering based on distance.

\subsubsection{\textbf{Feature Selection Process}}
In this section, we detail the feature selection process steps to identify the most relevant features for our analysis and benchmark the performance.
\noindent \textbf{Model 1 Mutual Information (M33I):} measures the amount of information obtained about one random variable through another random variable. In feature selection, it quantifies the dependency between each feature and the target variable. Features with high mutual information scores are considered more informative for predicting the target variable. Equation 2 \cite{34} defines the mutual information \( I(X_i; Y) \) as follows:

\begin{equation}
	I(X_i; Y) = \sum_{x_i \in X_i} \sum_{y \in Y} p(x_i, y) \log \left( \frac{p(x_i, y)}{p(x_i) p(y)} \right)
\end{equation}
\noindent where $X_i$ is the $i$-th feature, $Y$ is the target variable, $p(x_i, y)$ is the joint probability distribution function of $X_i$ and $Y$, $p(x_i)$ and $p(y)$ are the marginal probability distribution functions of $X_i$ and $Y$, respectively.

\noindent \textbf{Model 2 Chi-Square:} The test measures the independence between categorical variables. Feature selection quantifies the association between each categorical feature and the target variable. Features with high chi-square scores indicate a strong association with the target variable. The Chi-square value mathematical equation of joint probability genes can be calculated according to (3) \cite{35}
\begin{equation}
	\chi^2(X_i, Y) = \sum_{x_i \in X_i} \sum_{y \in Y} \frac{(O_{x_i, y} - E_{x_i, y})^2}{E_{x_i, y}}
\end{equation}
\noindent where $X_i$ is the $i$-th feature, $Y$ is the target variable, $O_{x_i, y}$ is the observed frequency of occurrence of $X_i$ and $Y$, and $E_{x_i, y}$ is the expected frequency of occurrence of $X_i$ and $Y$ under the null hypothesis of independence between them.

\noindent \textbf{Model 3 Random Forest Model:} is an ensemble learning method that constructs multiple decision trees during training and outputs the mode of the classes (classification) or mean prediction (regression) of the individual trees. It calculates feature importance based on how much the model's accuracy decreases when each feature is randomly shuffled. Features with higher importance scores contribute more to the predictive power of the random forest model. For classification, with the majority vote, the predicted class \(\hat{y}\) is given by \cite{36}:
\[
\hat{y} = \text{mode}\left(\{ T_b(\mathbf{x}) \}_{b=1}^B \right)
\]
where \( B \) is the total number of trees in the forest, and \( T_b(\mathbf{x}) \) is the prediction of the \( b \)-th tree for input \(\mathbf{x}\).

\noindent \textbf{Feature Importance Calculation} in RF can be measured by evaluating how much the model's prediction error increases when the values of a feature are permuted while keeping the other features unchanged. The feature importance for feature \( j \), \( \text{FI}_j \), can be computed as:
\[
\text{FI}_j = \frac{1}{B} \sum_{b=1}^B \sum_{t \in T_b} \Delta i_t \cdot {1}(j \in t)
\]
where: \( B \) is the total number of trees in the forest, \( T_b \) is the \( b \)-th tree, \( t \) represents a node in the tree,  \( \Delta i_t \) is the decrease in impurity at node \( t \) (e.g., reduction in Gini impurity or entropy), \( {1}(j \in t) \) is an indicator function that is 1 if feature \( j \) is used in node \( t \). To select the top \( k \) features based on their importance scores: (i) Compute the feature importances \( \{ \text{FI}_1, \text{FI}_2, \ldots, \text{FI}_p \} \). (ii) Sort the features by their importance scores in descending order. (iii) Select the top \( k \) features with the highest importance scores.

\noindent \textbf{Model 4 Principal Component Analysis (PCA):} reduces the dimensionality of a dataset while retaining most of the variance by transforming original features into principal components \cite{37}. The process involves standardizing the data (\( z_{ij} = \frac{x_{ij} - \mu_j}{\sigma_j} \)), computing the covariance matrix (\( \mathbf{C} = \frac{1}{n-1} \mathbf{Z}^T \mathbf{Z} \)), and performing eigenvalue decomposition (\( \mathbf{C} \mathbf{e}_i = \lambda_i \mathbf{e}_i \)) to obtain eigenvalues and eigenvectors. Principal components are formed by projecting standardized data onto eigenvectors (\( \mathbf{PC} = \mathbf{Z} \mathbf{E} \)). The top \( k \) components that explain the most variance are then selected. This transformation preserves essential data variability while reducing dimensionality.

\subsubsection{\textbf{AUC-ROC Analysis}}
Further, we applied the second feature selection process step to identify the best features using AUC-ROC analysis for each feature.  This approach was combined with the first feature selection process, where we obtained features from RF, MI, PCA, and Chi. We set a threshold (Th) and selected features if their AUC-ROC scores for both the training and testing datasets exceeded 0.80(Th). \[ \text{Best\_features} =\\
\{ X_i \mid \text{AUC}_i^\text{train} > \text{Th} \land \text{AUC}_i^\text{test} > \text{Th} \}
\] 
\[
\text{combined\_features} = \bigcup_{\text{method}} \text{best\_features}_{\text{method}}
\]
\[
\text{unique\_features} = \text{unique}(\text{combined\_features})
\]
\begin{equation}
	S_{\text{unique}} = \{ x \mid x \in S_{\text{all}} \}
\end{equation}
\begin{itemize}
	\item \( S_{\text{all}} \) represents the set of all selected features.
	\item \( S_{\text{unique}} \) represents the set of selected features with duplicates removed.
\end{itemize}

The Best features are selected from different methods, combined into a single list, and remove any duplicate features from the combined list. The $\text{combined\_features}$ denote the combined list of best features. The $\text{unique\_features}$ denote the list of unique features after removing duplicates.

\subsubsection{\textbf{Hierarchical Clustering}}
Further, we have applied the third feature selection process step i.e. hierarchical clustering in two ways. Where one approach is based on the dendrogram clusters to select the best features using clusters and second approach is to compute the pairwise distance between features with distance to select the features. The approaches are:
\noindent \textbf{1. Dendrogram Clusters}. Given a dataset represented by \( OMIC_{Subset1 \text{ to } 3} \), where each column represents a feature, we perform hierarchical dendrogram clustering to identify clusters of similar features. We create a dendrogram from the linkage matrix L and Assign cluster IDs to the features using the function.  
\begin{equation}
	D = \text{dendrogram}(L) C_i = \text{fcluster}(L, t)
\end{equation}
In order to extract features by each cluster, we store features in a dictionary called \text{clusters}, where each key is a cluster label \( C_i \), and the corresponding value is a list of features belonging to that cluster. For each cluster \( C_i \), we compute the feature importance \( I(C_i) \) by summing the absolute values of the features in the cluster.
\begin{equation}
	I(C_i) = \sum_{f \in F(C_i)} |f|
\end{equation}
\begin{equation}
	\text{selected\_features\_with\_cluster} = \bigcup_{i} \text{Top}_K(I(C_i))
\end{equation}
\begin{equation}
	\text{best\_features\_unique} = \text{set}(\text{selected\_features\_with\_cluster})
\end{equation}
Then, selected the top \( K \) important features from each cluster to form \text{selected\_features\_with\_cluster} and merge the selected features to get a unique set of best features.

\noindent \textbf{2. Pairwise Distances}
In the second approach, the hierarchical clustering is performed on the pairwise distance matrix \( D \) using the Ward linkage method, which aims to minimize the variance when forming clusters. We compute the pairwise distance matrix \( D \) using the Euclidean metric between the transpose of the scaled feature matrix \( OMIC_{SUBSET}[\text{S\_unique}] \). The Euclidean distance between two points \( p \) and \( q \) in \( n \)- dimensional space and set a distance threshold \( \theta \) to a specific value i.e. \( \theta = 3.5 \). is given by:
\begin{equation}
	\text{Euclidean distance}(p, q) = \sqrt{\sum_{i=1}^{n} (q_i - p_i)^2} 
\end{equation}
The Ward linkage criterion is calculated as \cite{38}: \[ Z = \text{linkage}(D, \text{method}='ward') \]
\begin{equation}
	d(u,v) = \sqrt{\frac{|v|+|s|}{|T|}d(v,s)^2 + \frac{|v|+|t|}{|T|}d(v,t)^2 - \frac{|v|}{|T|}d(s,t)^2} 
\end{equation}
where, \( d(u,v) \) represents the distance between clusters \( u \) and \( v \), and \( |v| \), \( |s| \), \( |t| \), and \( |T| \) denote the number of points in clusters and subclusters, and \( d(v,s) \), \( d(v,t) \), and \( d(s,t) \) are the distances between clusters and subclusters.

\subsection{\textbf{Diagnostic Classical and Quantum Classifiers}}
In the fourth phase, we define the machine learning and quantum machine learning methods employed for subtype-I and subtype-II lung datasets diagnostic classification. In our experimental analysis, we used four distinct machine-learning algorithms. This section comprehensively overviews these classifiers and their respective theoretical implementations.

\textbf{Logistic Regression (LR)} is a linear model used for binary classification problems. Using the logistic function, it estimates the probability that an instance belongs to a particular class. The model is based on the linear combination of input features \cite{39}.

\begin{equation}
	\begin{aligned}
		&\text{Logistic Function :}  \quad \sigma(z) = \frac{1}{1 + e^{-z}} \\
		&\text{Prediction:}  \quad \hat{y} = \sigma(\mathbf{w}^T \mathbf{x} + b) \\
		&\text{Log loss:} -\frac{1}{N} \sum_{i=1}^N \left[ y_i \log(\hat{y}_i) + (1 - y_i) \log(1 - \hat{y}_i) \right]
	\end{aligned}
\end{equation}

Where $\mathbf{w}$ represents the weights, $\mathbf{x}$ the input features, $b$ the bias term, $\hat{y}$ the predicted probability, and $y$ the actual label. Regularization can be applied to prevent overfitting, and in this case, L2 regularization is used with a penalty parameter $C=0.1$.

{ \textbf{Multi-Layer Perceptron (MLP)} for a MLP with \( L \) layers, the activation \( a_j^{(l)} \) of neuron \( j \) in layer \( l \) is computed recursively from the input layer \( l=1 \) to the output layer \( l=L \) \cite{40}:
	\[
	a_j^{(l)} = f\left( \sum_{i=1}^{n^{(l-1)}} w_{ij}^{(l)} a_i^{(l-1)} + b_j^{(l)} \right)
	\]
	where:\( a_j^{(l)} \): Activation of neuron \( j \) in layer \( l \), \( w_{ij}^{(l)} \): Weight connecting neuron \( i \) in layer \( l-1 \) to neuron \( j \) in layer \( l \), \( a_i^{(l-1)} \): Activation of neuron \( i \) in the previous layer \( l-1 \),\( b_j^{(l)} \): Bias term for neuron \( j \) in layer \( l \), \( f \): Activation function applied element-wise to the linear combination.
	
	\textbf{Support Vector Machine (SVM)} is a supervised learning model used for classification tasks. It finds the optimal hyperplane that maximizes the margin between different classes. The kernel trick allows SVMs to perform non-linear classification by mapping input features into higher-dimensional space \cite{41}
	\begin{equation}
		\begin{aligned}
			&\text{Decision Function:} \quad f(\mathbf{x}) = \mathbf{w}^T \phi(\mathbf{x}) + b \\
			&\text{Optimization Problem:} \\
			&\min_{\mathbf{w}, b} \frac{1}{2} \|\mathbf{w}\|^2 + C \sum_{i=1}^N \max(0, 1 - y_i (\mathbf{w}^T \phi(\mathbf{x}_i) + b))
		\end{aligned}
	\end{equation}
	
	Where $\phi$ is the feature mapping function (RBF kernel in this case), $\mathbf{w}$ the weight vector, $b$ the bias term, $C$ is the regularization parameter, and $y$ the actual label.
	
	{\textbf{Random Forest (RF)} is an ensemble learning method that constructs multiple decision trees during training and outputs the class that is the mode of the classes (classification) or mean prediction (regression) of the individual trees \cite{36}. It is robust to over-fitting and performs well on many datasets.
		\begin{align}
			\text{Prediction:} \quad \hat{y} &= \frac{1}{T} \sum_{t=1}^T h_t(\mathbf{x})
		\end{align}
		Where, $T$ is the number of trees, and $h_t$ is the prediction of the $t$-th tree. Parameters such as the number of estimators (trees), criterion (e.g., entropy), max depth, and others can be tuned for optimal performance.
		
		{ \textbf{Quantum Neural Network (QNN):} In this section, we will provide an overview of the hybrid quantum neural network model, detailing the methodology used for three different models. Each model varies in the number of features (256, 64, and 32) and the corresponding number of qubits (8, 6, and 5) used for encoding, respectively. The models are named as 
			QNN$_{1}$ for 256 features, QNN$_{2}$ for 64 features and QNN$_{3}$ for 32 features. The summary of hybrid models is given in Table I. The hybrid model leverages a combination of QNNs and classical neural networks (CNN). 
			
			\noindent \textbf{Feature Encoding:} The initial step involves encoding a classical vector into a quantum state. To efficiently simulate quantum circuits, each feature vector N-dimensional  \textit{x} is normalized and embedded into a quantum state $\ket{\psi}$ using amplitude encoding as 
			\begin{equation}
				\ket{\psi_x}=\frac{1}{\parallel x \parallel} \sum_{j=1}^{N} x_j \ket{j}
			\end{equation}

\begin{table}[!ht]
	\centering
	\begin{scriptsize}
		\caption{Summary of hybrid quantum-classical models}
		\begin{tabular}{|l|c|c|c|c|c| }
			\hline
			Models & Features & Qubits & Depth & Gates & Dense\\
			\hline
			QNN$_{1}$  & 256 & 8 & 5 &  8 Rot, 7 CZ& 256  \\
			QNN$_{2}$  & 64  & 6 &  5 & 6 Rot, 5 CZ&  64  \\
			QNN$_{3}$ & 32  & 5 &  5 & 5 Rot, 4 CZ & 32    \\
			\hline
		\end{tabular}
	\end{scriptsize}
\end{table}

\noindent \textbf{Quantum Layer Ansatz:} Subsequently, a sequence of parameterized quantum gates, denoted as ${U}(\theta)$, is applied to the quantum state. This unitary transformation ${U}(\theta)$ consists of several local quantum gates, such as two-qubit unitaries. The entire quantum circuit 
\textit{U} is made up of \textit{n} unitary blocks, expressed as $\mathcal{U}(\overrightarrow{\theta}) = U_1U_2...U_n$, where ith unitary block is:
\begin{equation}
	U_i(\theta_i)=exp(-i\theta_iP)
\end{equation}
\noindent where \textit{P} consists of Pauli operators, and $\theta_i$ denotes a gate parameter vector of $U_i(\theta_i)$.

\begin{algorithm}[H]
	\begin{scriptsize}
		\caption{MQML-LungSC: Multi-Omic Quantum Machine Learning for Lung Subtype Classification Framework Analysis Workflow}
		\begin{algorithmic}[1]
			\State \textbf{Input:} GDC-TCGA dataset Type: LUSC and LUAD Lung dataset, 915 patients
			\State \hspace{\algorithmicindent} DNA-OMIC 1: DNA$_{S1}$, DNA$_{S2}$, DNA$_{S3}$, RNA-OMIC 2: RNA$_{S1}$, RNA$_{S2}$, RNA$_{S3}$, miRNA-OMIC 3: miRNA$_{S1}$, miRNA$_{S2}$
			\State \hspace{\algorithmicindent} Dataset Patients: (i) DNA: 503 samples \& 485,577 genes, (ii) RNA: 585 samples \& 60,488 genes, 
			(iii) miRNA: 564 samples \& 1,881 genes.
			\State \hspace{\algorithmicindent} Feature Selection: (i) Mutual Information(MI), (ii) Chi-square, (iii) Principal Component Analysis(PCA), (iv) Random Forest(RF)
			
			\State \hspace{\algorithmicindent} Expressions: $X'$ is the resulting matrix. $x$ represents a row vector in $X$. $x_i$ represents the $i$th element of the row vector $x$. $n$ is the number of elements in each row vector $x$.
			
			\State \textbf{PHASE 1- \textit{Data Acquisition and Data Preprocessing:}}
			\State \hspace{\algorithmicindent} \textit{Multi-Omics:} (i) DNA: 503 samples \& 485,577, (ii) RNA: 585 samples \& 60,488 genes, (iii) miRNA: 564 samples \& 1,881 genes.
			\State \hspace{\algorithmicindent} \textit{Pre-processing:} \textit{Combined LUAD and LUSC:} \textit{Concatenate Raw Omics} $\text{OMIC-Data}_{\text{LUSC}} \cup \text{OMIC-Data}_{\text{LUAD}}$  
			\State \hspace{\algorithmicindent} Remove rows where the sum of elements is less than or equal to 0: $X' = \{ x \in X \, | \, \sum_{i=1}^{n} x_i > 0 \}$
			
			\State \textbf{PHASE 2- Feature Engineering: t-test Statistics}
			\For{Each OMIC (DNA$_{1}$, RNA$_{2}$, miRNA$_{3}$)}
			\State \textit{Combine Patient Attributes:} Combine clinical-survival patient data with $OMIC[1]_{\text{LUSC}}$ and $OMIC[1]_{\text{LUAD}}$ 
			\State \textit{Subset Splitting and Join Operation:} Rank the clinical data order based on sample sub-type and isolate LUAD and LUSC into separate data frames. $\text{Subset}_{\text{LUSC-I}} \cup \text{Subset}_{\text{LUAD-II}}$
			\State \textit{Mean Calculation:} Calculate the mean for each column feature, i.e., column-wise genes: 
			$\bar{X} = \frac{1}{n} \sum_{i=1}^{n} X_i$ 
			\State \textit{Calculate t-test:} 
			$t = \frac{\bar{X}_{\text{LUSC}} - \bar{X}_{\text{LUAD}}}{s_{\text{LUSC}} + s_{\text{LUAD}}}$
			\State \textit{Sorting rank-wise features (P-value) to make significant and non-significant features}
			\State \textit{Integrate (by join operation) with the minimum sample of omic:} 
			$\text{Join\_Operation:}(\text{OMIC-Data}_1, \text{OMIC-Data}_2, \text{OMIC-Data}_3)$
			\State \textit{Separate all omics into independent sets of data, i.e., row-wise TCGA-ID samples and column-wise features}
			\State \textit{Combine:} $(\text{Omic\_Dataset}, \text{Clinical\_Features}, \text{Survival\_Features})$ \text{of LUAD and LUSC subtypes}
			\EndFor
			\State \textbf{PHASE 3-Feature Selection: PROCESS 1-ML methods}
			\For{Each OMIC (DNA$_{LUSC-LUAD}$, RNA$_{LUSC-LUAD}$, miRNA$_{LUSC-LUAD}$)}
			\State Divide into subsets (Omic$_{1}$: S$_{1}$, S$_{2}$, S$_{3}$) based on the p-value and drop the features with unnamed GENE
			\State Apply the K-best selection method with a different no of features: MI, Chi, PCA, and RF on each subset.
			\State Split data into training and testing sets: $X_{\text{train}}, X_{\text{test}}, y_{\text{train}}, y_{\text{test}} \gets \text{train\_test\_split}(X_{\text{scaled}}, y_{\text{label}}, 0.2, \text{seed}=42)$
			\State Define selected features: $\text{selected\_features} \gets [\text{selected\_MI}, \text{selected\_PCA}, \text{selected\_Chi}, \text{selected\_RF}]$
			\State Define selected scores: $\text{selected scores} \gets [\text{scores MI}, \text{scores PCA}, \text{scores Chi}, \text{scores RF}]$
			
			\State Initialize dictionary: $\text{best\_features} \gets \{\}$ \text{and} Initialize list: $\text{selected\_features\_all} \gets []$
			\State The number of unique features and common across all the methods is given by:
			\[
			|S_{\text{Unique}}| = |S_{\text{MI}} \cup S_{\text{PCA}} \cup S_{\chi^2}| - |S_{\text{Common}}|
			\]
			\[
			|S_{\text{Common}}| = |S_{\text{MI}} \cap S_{\text{PCA}} \cap S_{\chi^2}|
			\]
			\EndFor
			
		\end{algorithmic}
	\end{scriptsize}
\end{algorithm}

\begin{algorithm}[H]
	\begin{scriptsize}
		\caption{MQML-LungSC: (Contd.)}
		\begin{algorithmic}[1]
			\State \textbf{PROCESS 2- Perform AUC-ROC analysis with RF classifier:} Apply RF classifier (n-estimator=250) with conditions on each feature to obtain the best features.
			\For{$i$ in $\{0, 1, 2, 3\}$}
			\State Get method, features, and scores: $method, features, scores \gets \text{methods}[i], \text{best\_selected\_features}[i], \text{selected\_scores}[i]$
			\State Sort features by scores
			\For{each feature in sorted features}
			\State Evaluate ROC AUC for feature of TR and TS
			\If{ROC AUC $> 0.80$ }
			\State Append feature to $\text{best\_selected\_features}[method]$
			\EndIf
			\EndFor
			\EndFor
			\State Calculate total count of features, Combine best features and Remove duplicate features.
			\State \textbf{PROCESS 3- \textit{Hierarchical Clustering Selection:}}
			\State $(1) \text{pairwise\_distances} \gets \text{pdist}(X^T, \text{metric}=\text{'euclidean'})$ \text{and} $ \text{distance\_threshold} = \text{Max\_number}$
			\State $(2) Z \gets \text{linkage}(\text{pairwise\_distances}, \text{method}=\text{'ward'})$
			\State $(3) \text{cluster} \gets \text{fcluster}(Z, \text{t}=\text{distance\_threshold}, \text{criterion}=\text{'maxclust'})$
			\State $(4) \text{Split Cluster Group:} (i)\text{No. of clusters}, (ii)\text{Select top clusters}$
			\State (5) \text{Apply heatmap and dendrogram analysis for clustering}
			\State \textbf{PROCESS 4- Apply UMAP for Dimensionality Reduction:}
			\State $X_{\text{reduced}} \gets \text{UMAP}(\text{n\_components}=2, \text{min\_dist}=0.1, \text{n\_neighbors}=15)$
			\State \textbf{PROCESS 5- Apply Quantum Machine Learning:} Train the QML model on the selected features for subtype classification and calculate accuracy metrics.
		\end{algorithmic}
	\end{scriptsize}
\end{algorithm}

In quantum ansatz, we have applied parameterized rotations \(R(\theta, \phi, \lambda)\), and Controlled-Z (CZ) gates for entanglement between adjacent qubits. Each qubit undergoes a rotation defined by three parameters, \( \theta \), \( \phi \), and \( \lambda \). The rotation operation \( R(\theta, \phi, \lambda) \) can be expressed as:
\begin{equation}
	R(\theta, \phi, \lambda) = R_z(\lambda) R_y(\theta) R_z(\phi)
\end{equation}
The controlled-Z gates introduce entanglement between adjacent qubits, which a phase flip if both qubits are in the \(|1\rangle\) state. For instance, a single  layer of ansatz applies the following unitary transformation on the 8-qubit system as:
\begin{equation}
	U(W) = \prod_{i=0}^{7} R(\theta_i, \phi_i, \lambda_i) \prod_{i=0}^{6} CZ_{i,i+1}
\end{equation}

\begin{figure*}[!ht]	
	\includegraphics[scale=0.23]{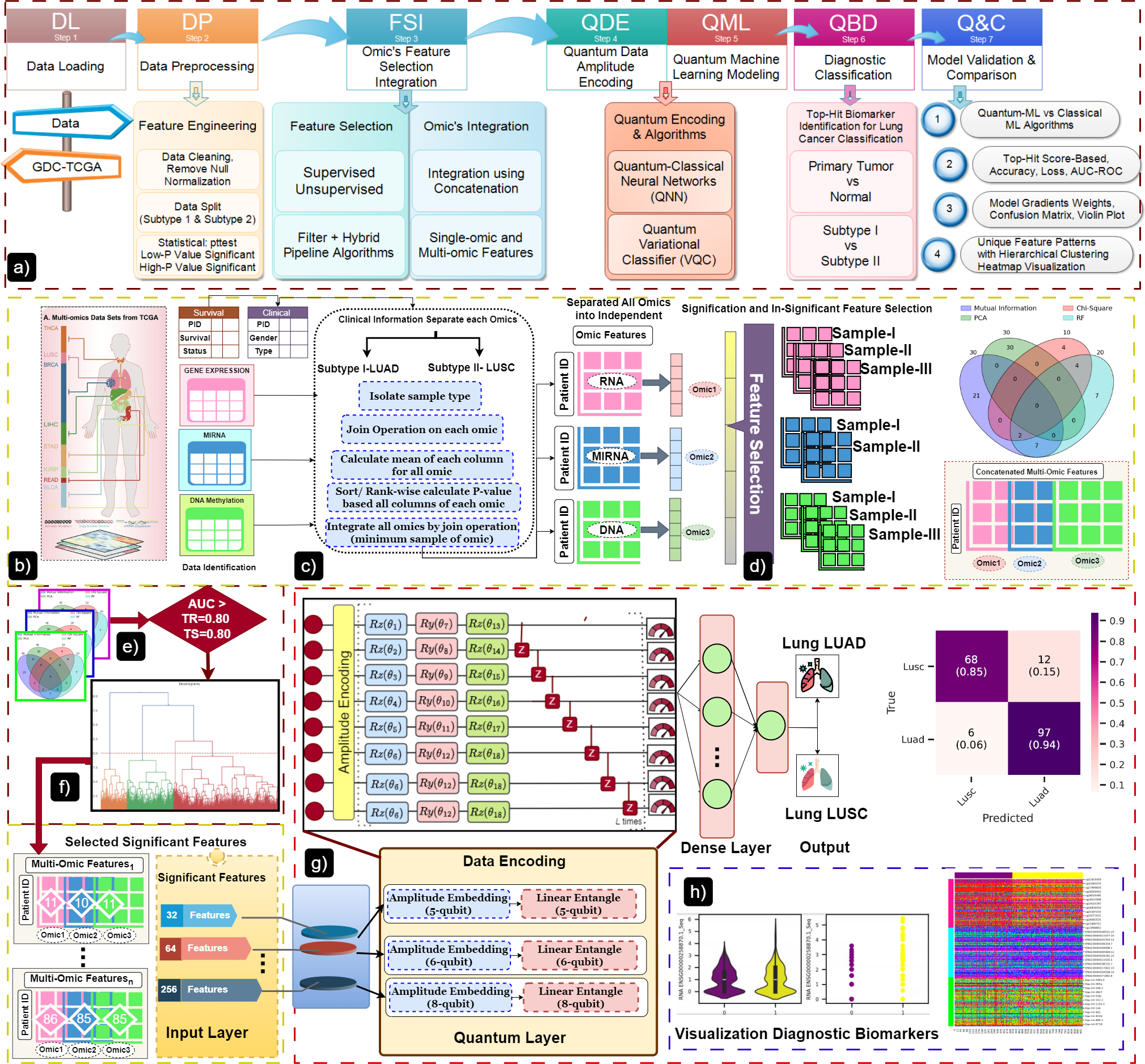}
	\caption{\textbf{Workflow of multi-omics integration and classification using quantum neural networks}. 
		\textbf{(a)} Schematic representation of the overall pipeline of MQML-QNN framework. 
		\textbf{(b)} Data acquisition from GDC-TCGA, including (i) DNAme, (ii) RNA-seq, (iii) miRNA-seq, and (iv) clinical and survival attributes of patients. 
		\textbf{ (c)} Data preprocessing and feature engineering using t-test p-values for each omic data type to differentiate between subtypes I and II.
		\textbf{(d)}-\textbf{(e)} Feature selection process involving: (i) Four feature selection models: Random Forest (RF), Mutual Information (MI), Principal Component Analysis (PCA), and Chi-Squared (Chi). (ii) AUC-ROC analysis with thresholding for feature selection. (iii) Hierarchical clustering based on sample vs. feature and pairwise feature similarity using distance metrics.
		\textbf{(f)} Integration of multi-omic data resulting in 256 features from 915 common patients.
		\textbf{(g)} Quantum amplitude encoding with three feature dimensions: 32, 64, and 256, and the quantum circuit with a dense layer for diagnostic classification of LUAD versus LUSC lung datasets.
		\textbf{(h)} Visualization of features through violin plots, confusion matrices, and heatmaps.}
\end{figure*}

\noindent \textbf{Measurement:} Finally, a quantum measurement ($\hat{O}$) is applied using the expectation value of the Pauli-Z operator for each qubit
\begin{equation}
	\tilde{y}_i=\braket{\psi_i|\mathcal{U}(\theta)^{\intercal}\hat{O}\mathcal{U}(\theta)|\psi_i}
\end{equation}
The measurement results from the quantum circuit are then fed into a layer of a classical neural network, which is used to predict the label of the input state. 
\noindent \textbf{Objective Function:} In diagnostic classification, a binary classification task distinguishing between Subtype-I and Subtype-II using multi-omic data aims to minimize the binary cross-entropy loss. The binary cross-entropy loss function is defined as follows:
\begin{equation}
	\mathcal{L} = -\frac{1}{N} \sum_{i=1}^{N} \left[ y_i \log(\hat{y}_i) + (1 - y_i) \log(1 - \hat{y}_i) \right]
\end{equation}

\noindent where \textit{N} is the number of samples (915 in our case), \(y_i\) is the true label for the \(i\)-th sample (\(y_i \in \{0, 1\}\)) with (0) representing LUSC$_{Subtype-I}$ and (1) representing LUAD$_{Subtype-II}$, and \(\hat{y}_i\) is the predicted probability for the \(i\)-th sample.

\subsection{ \textbf{Multi-Omic Integration and Development}}

In the last phase, the subsets of Omic\(_{S1}\), Omic\(_{S2}\), and Omic\(_{S3}\) are combined. The multi-omic integration process is then followed to integrate all the Omic data to train the quantum neural network model using different feature dimensions with amplitude encoding, such as Multi-Omic\(_{256}\), Multi-Omic\(_{64}\), and Multi-Omic\(_{32}\).

\begin{table*}
	\caption{Range of p-values used for selecting most significant and less significant features.}
	\begin{scriptsize}
		\begin{tabular}{ |p{0.2cm} | p{1.5cm}|p{2.2cm}|  p{2.6 cm}|  p{1.8cm}| p{2.65cm}| p{1.2cm}|}
			\hline
			\multicolumn{1}{|c|}{} & \multicolumn{2}{|c|}{OMIC1: DNAme} & \multicolumn{2}{|c|}{OMIC2: RNA-seq} & \multicolumn{2}{|c|}{OMIC3: miRNA-seq} \\
			\hline
			Set & P-value & Index  & P-value & Index   & P-value & Index  \\
			\hline
			1	&	8.42e-25 to 4.99e-02& 140000 - 299885  &	5.19e-24 to 4.99e-02&	7000 - 36900 & 1.17e-25 to 4.99e-02 & 100-821   \\
			2	&	2.89e-06 to 4.99e-02&	241000 - 299885	&	2.53e-05 to 4.99e-02&	22300 - 36900	& 2.35e-03 to 9.93e-01 &  600-1586\\
			3	&	4.99e-02 to 9.99e-01	&	299885 - 344344	    &	4.99e-02 to 9.99e-01	        &	36900 - 58324  &  -   &- \\
			\hline
		\end{tabular}
	\end{scriptsize}
\end{table*}

\begin{table*}
	\caption{Feature Selection Process using P-value based data}
	\begin{scriptsize}
		\begin{tabular}{ |p{0.3cm} | p{1.2 cm}|p{1.7cm} |p{1cm}| p{1.2 cm}|  p{1.7cm} |p{1cm}| p{1.2cm}| p{1.7cm}|p{0.85 cm}|}
			\hline
			\multicolumn{1}{|c|}{} & \multicolumn{3}{|c|}{OMIC1: DNA} & \multicolumn{3}{|c|}{OMIC2: RNA} & \multicolumn{3}{|c|}{OMIC3: miRNA} \\
			\hline
			Set & Features & FS Methods & FS Clt & Features & FS Methods &FS Clt  & Features & FS Methods & FS Clt  \\
			\hline
			1	&	159885 & 40       &	10  &29900	 & 39 (37) & 20  &721 & 70(63) &35 \\
			2	&	44459 & 70 (68)   &	50  &14600	 &58 (51)  & 32  &986 & 76 (63) & 50\\
			3	&	58885 & 47 (46)   &	25  &21424	 &97(71)  & 34   & -& -&-  \\
			\hline
		\end{tabular}
	\end{scriptsize}
\end{table*}

\section{METHODOLOGY}
\subsection{\textbf{Overview of MQML framework}}
We present a pioneering framework of multi-omic-quantum machine learning (MQML-LungSC) tailored for integrating data, selecting unique features, classifying diagnostics, and identifying features associated with two distinct subtypes of lung datasets: LUSC primary tumor type-I and  LUAD primary tumor type-II. Due to the datasets' inherent heterogeneity, complexity, and high dimensionality, our study aims to assess the efficacy of hybrid-QML in selecting, reducing, encoding features, and classifying diagnostics between subtype-I and subtype-II patients across integrated multi-omic subtype datasets. 
The whole process is divided into a few phases as described in Fig 3. (i) Data loading, (ii) Data Pre-processing- Feature engineering, (iii) Feature Selection- split into three steps, (iv) quantum data encoding and circuit, (v) quantum-classical hybrid model development phase, testing phase and (vi) identification of molecular top-hit features genes.
MQML processes various omic datasets (epigenomic and transcriptomic) by pre-processing to ensure data quality and reduce the computational burden. Omic features are columns, and patient samples are rows. In the first data pre-processing phase, (Omic[1]$_{Subtype-I}$ and Omic[1]$_{Subtype-II}$) are combined and patient survival and clinical attributes are extracted based on the diagnostic subtype i.e. sample-type. In the second phase, we conducted a feature engineering process using the statistical t-test method to identify significant and non-significant features within each omic. Then, we split each Omic[1]$_{subset1}$ to Omic[1]$_{subset3}$ subset into three parts based on p-values: two subsets with p-values less than 0.05 and one subset with p-values greater than 0.05. Table IV shows the features selected using a feature selection process based on p-value for further diagnostic classification in multi-omic modalities. In the third phase, we applied a feature selection process on each Omic[1]$_{subset1-3}$ into three steps: (i) four machine learning feature selection methods, (ii) AUC-ROC analysis, and (iii) hierarchical clustering with distance-based methods to obtain the most optimal and unique features for each omic subset. 

In the first selection, we applied four feature selection methods: (MI), Chi-square, (PCA), and (RF). Each subset of omics/modality was then analyzed to select the best features using the four selection methods, considering both common and uncommon features. In the second step of feature selection, the total selected features were further processed based on a random forest classifier, considering only features with AUC-ROC scores higher than 0.80 in both training and testing datasets for further analysis. In the third step, then we applied the hierarchical clustering approach, selecting additional features based on two criteria: (i) Pearson-Euclidean distance between features and (ii) Euclidean distance using the ward method of clustering. The top cluster features were obtained on each Omic$_{Subset}$. Finally, the selected features based on distances were processed to visualize each modality heatmap and other visualization plots. Further, this feature selection process was ap- plied to select the best and unique features from each omic modality (DNA-me, RNA-Seq, miRNA-Seq) based on their subsets. Then, each subset of modality was combined among single omic, such as three subsets for Omic$_{DNA-Single}$ and Omic$_{RNA-Single}$, and two subsets for Omic$_{miRNA-Single}$.
\begin{table}[!ht]
	\centering
	\caption{List of P-value Based Conducted Encoding and Simulations}
	\begin{scriptsize}
		\begin{tabular}{|c|c|c|c|c|}
			\hline
			\textbf{Set} & \textbf{P-value} & \textbf{DNAme} & \textbf{RNA-seq} & \textbf{miRNA-seq} \\
			\hline
			$S_{1}$ & $<0.05$ & 10 & 20 & 35 \\
			\hline
			$S_{2}$ & $<0.05$ & 25 & 32 & - \\
			\hline
			$S_{3}$ & $>0.05$ & 50 & 34 & - \\
			\hline
			$S_{4}$ & $<0.05$ & - & - & 51 \\
			\hline
			\multicolumn{2}{|c|}{\textbf{Multi-Omic Data}} & 85 & 86 & 85 \\
			\hline
		\end{tabular}
	\end{scriptsize}
	\label{tab:simple_table}
\end{table}
Further, in the third phase, the multi-omic integration process is followed to combine each subset of Omic1 = Omic[1]$_{subset1}$+Omic[1]$_{subset2}$+Omic[1]$_{subset3}$ and then integrate all the single Omic such as Omic$_{1}$, Omic$_{2}$, and Omic$_{3} $ to train the QNN model using different dimensions of features with amplitude encoding such as (Multi-Omic$_{256}$, Multi-Omic$_{64}$ and Multi-Omic$_{32}$). In this step, all the single omics were integrated using hierarchical integration column-wise among 915 patient samples of LUAD and LUSC. This combined integrated dataset consists of 85 features for Omic$_{DNA}$, 86 for Omic$_{RNA}$, and 85 for Omic$_{miRNA}$, totaling 256 features for multi-omic analysis. 
Now, to evaluate the performance of the quantum model, we evaluate three QNN models on three sets of datasets. We randomly selected features to create datasets with 32 and 64 features, forming integrated multi-omic data. This allowed us to perform the QNN model classification across three data dimensions: 32, 64, and 256.

In the Fourth Phase, we developed a MQML-QNN model for the diagnostic classification of lung subtype-I and subtype-II. We compared the quantum model with classical machine learning classifiers to evaluate the performance as depicted in Table X (LR, MLP, and SVM ). We benchmarked the performance of the best results with Quantum neural network and Random Forest and evaluated the comparison with all existing classical models for the 256, 64, and 32 features of integrated multi-omics as shown in Table X and Figure 13. 

Additionally, In the last phase, we extracted top-hit multi-omic molecular features' importance features using the QNN model weights for further analysis. The workflow of the proposed (MQML-LungSC) framework is shown in Fig. 3.

\section{RESULTS and PERFORMANCE}
In this section, we will present the performance of our proposal MQML-LungSC framework in detail. The performance of MQML was designed to compare with three dimensions of QNN models that perform diagnostic subtype classification. The metrics used to compare the classification performance of (MQML-LungSC) were accuracy (ACC), loss, precision, recall, and F1-score (F1). The F1 score was calculated by the mean F1 score of each class, weighted by the size of that class. Diagnostic and Identification of LUSC and LUAD subtypes related gene after the omic data integration. For each dataset subtype, RNA-seq, DNAme and miRNA-seq are used for data integration with common patients.

\begin{figure*}[!ht]
	\centering
	\begin{tabular}{c}
		\textbf{(a)} 
		\includegraphics[scale=0.5]{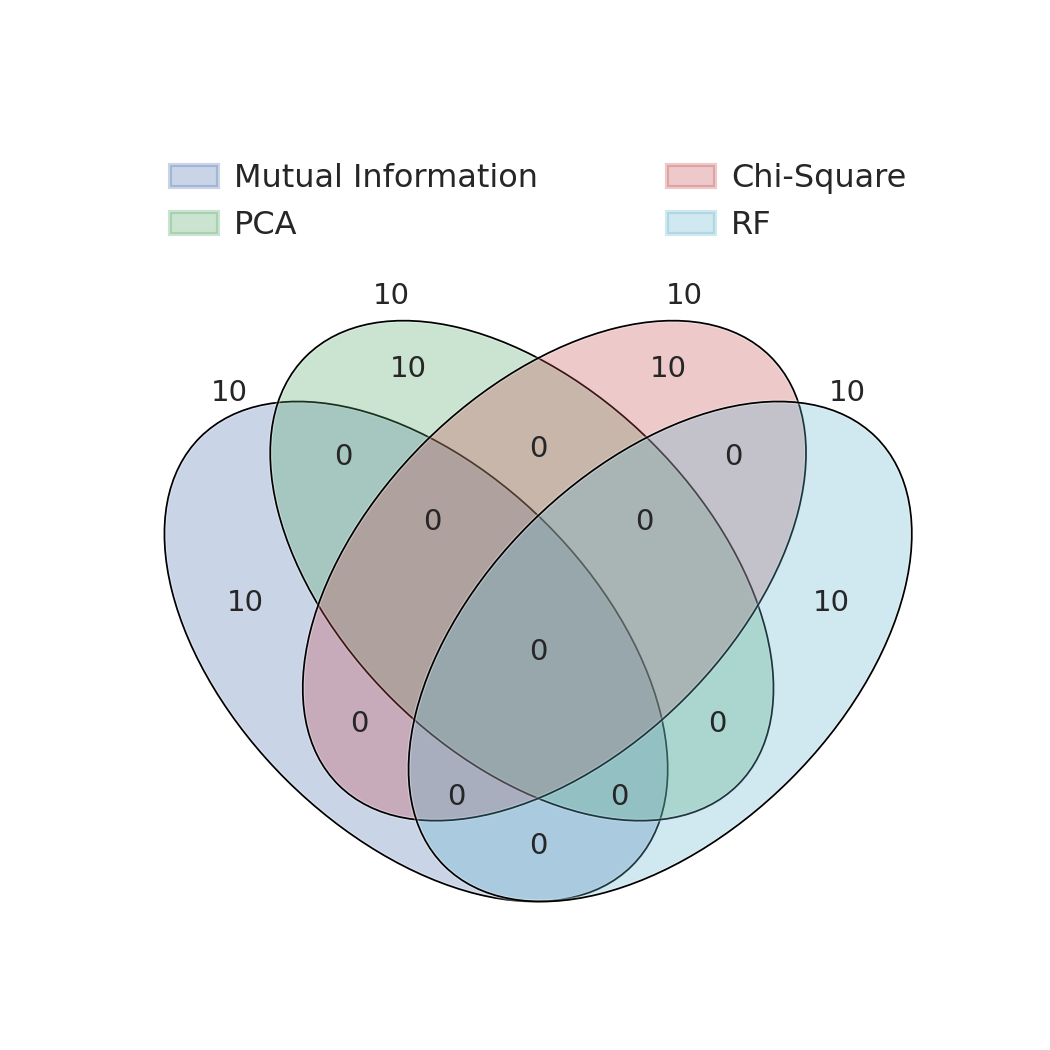} 
		\textbf{(b)} 
		\includegraphics[scale=0.5]{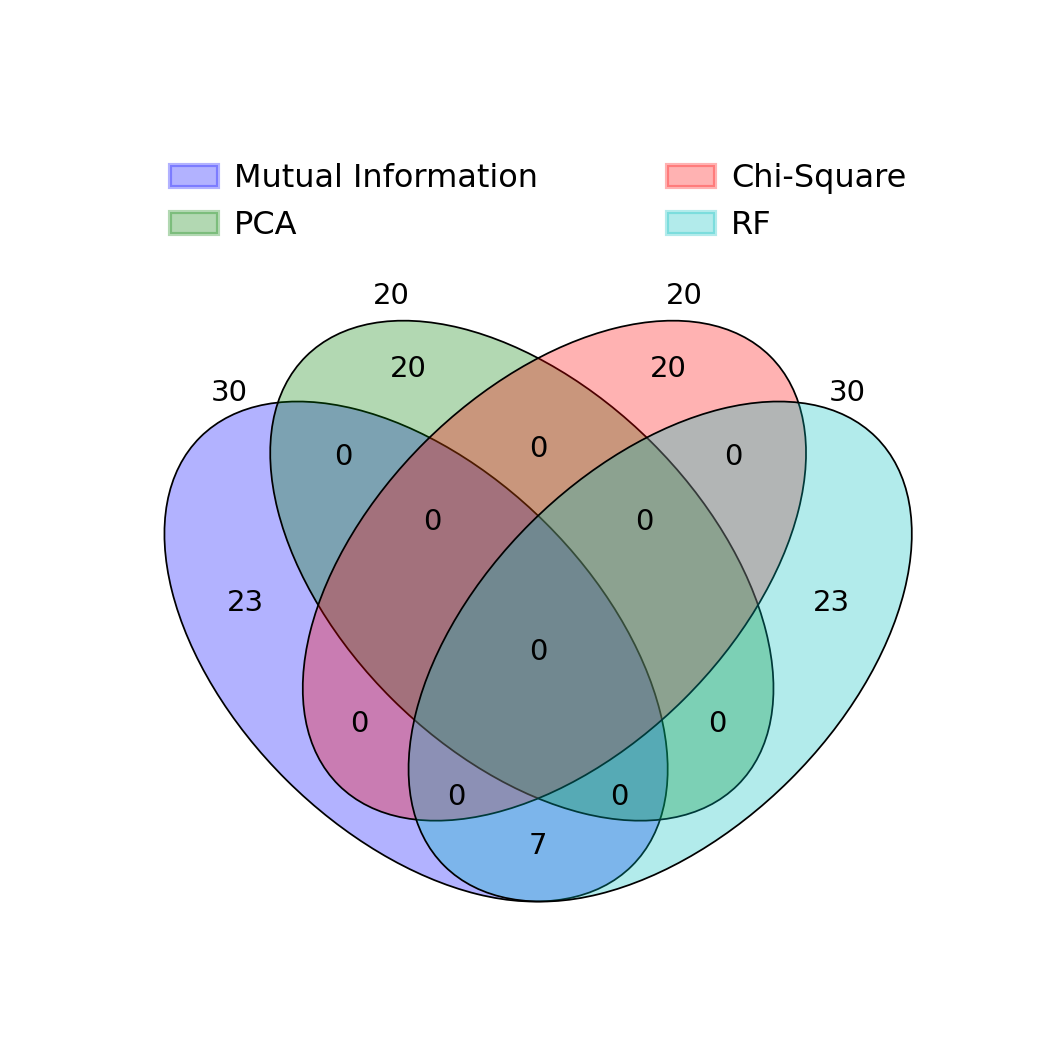} 
		\textbf{c)} 
		\includegraphics[scale=0.5]{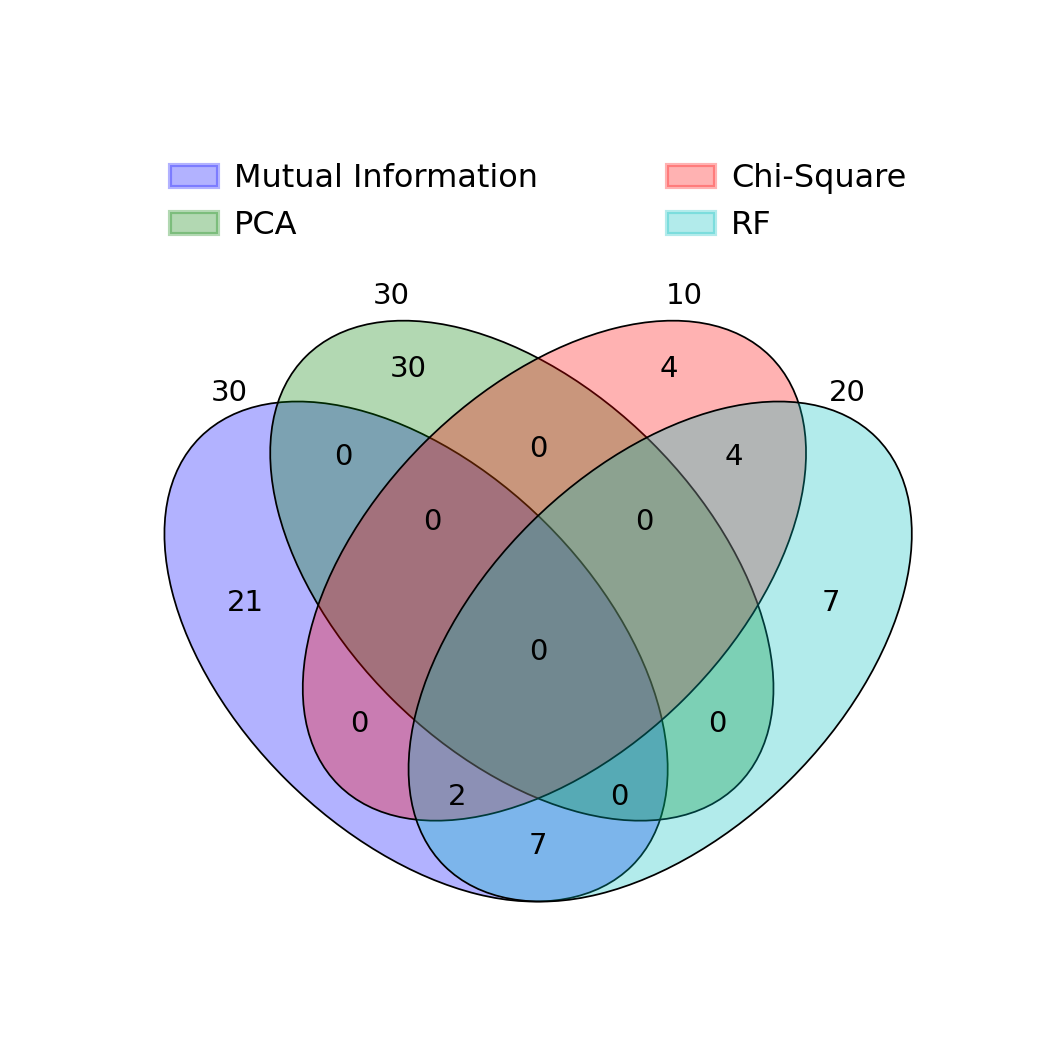} 
		
	\end{tabular}
	\caption{\textbf{Visualization of Venn Diagram and \textbf{Represents the Feature Importance Plot with score }} Represent the visualization through venn diagram for the features selection process using four machine learning methods on DNA methylation (DNAme), RNA transcript level (RNA-seq), and miRNA-seq levels (miRNA) modalities. (a) DNA Sample$_{1}$ to Sample$_{3}$, (b) RNA Sample$_{1}$ to Sample$_{3}$ and (c) miRNA Sample$_{1}$ to Sample$_{2}$,  Visualization of feature selection using four models i.e. i) Mututal information, (ii) Chi-square, (iii) Principal component (iv) Random forest of three modalities (d) DNA S$_{1}$ to S$_{3}$, (e) RNA S$_{1}$ to S$_{3}$, and (f) miRNA S$_{1}$ to S$_{2}$}
\end{figure*}

\begin{figure*}[!ht]
	\centering
	\includegraphics[scale=0.15]{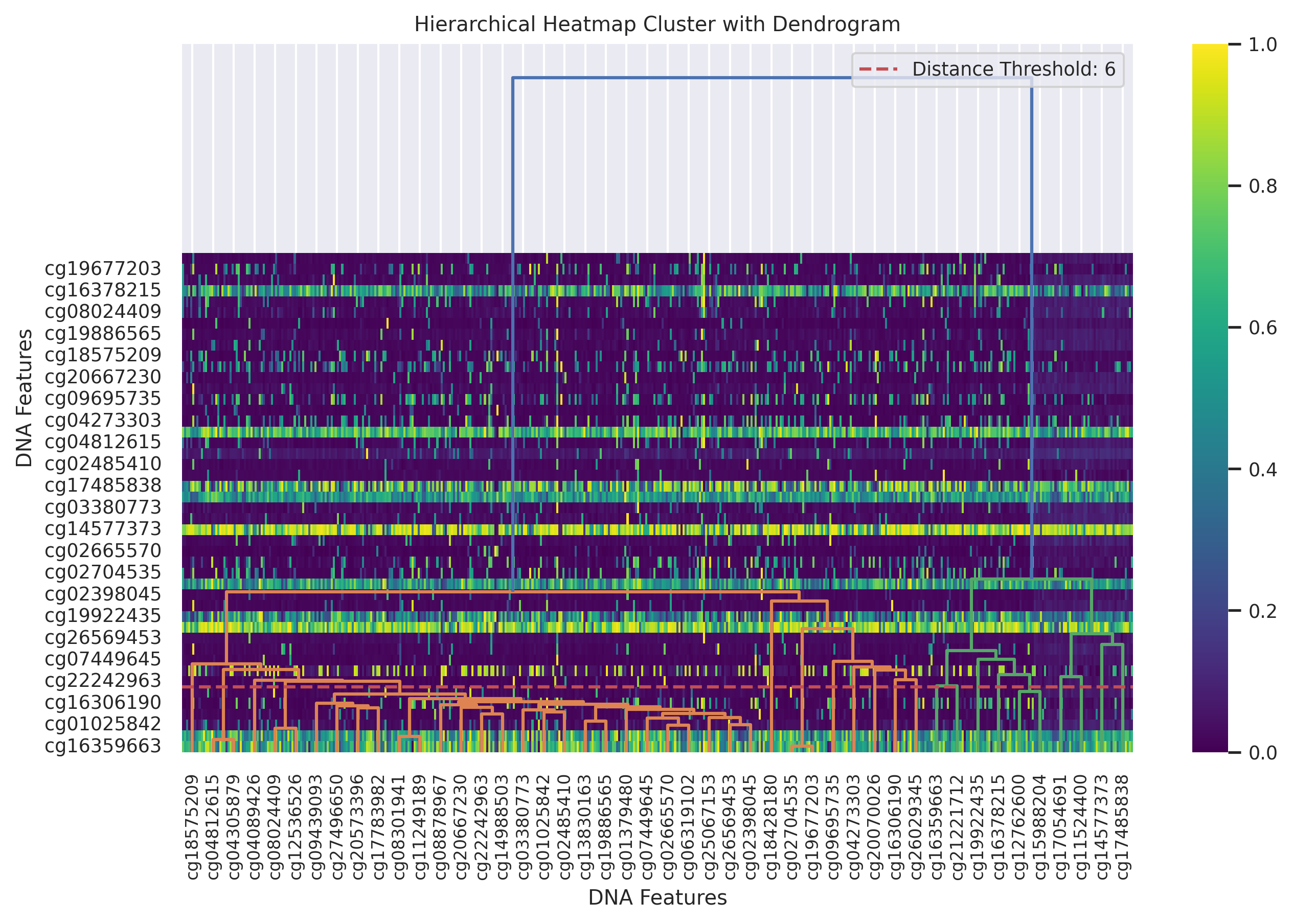}
	\includegraphics[scale=0.15]{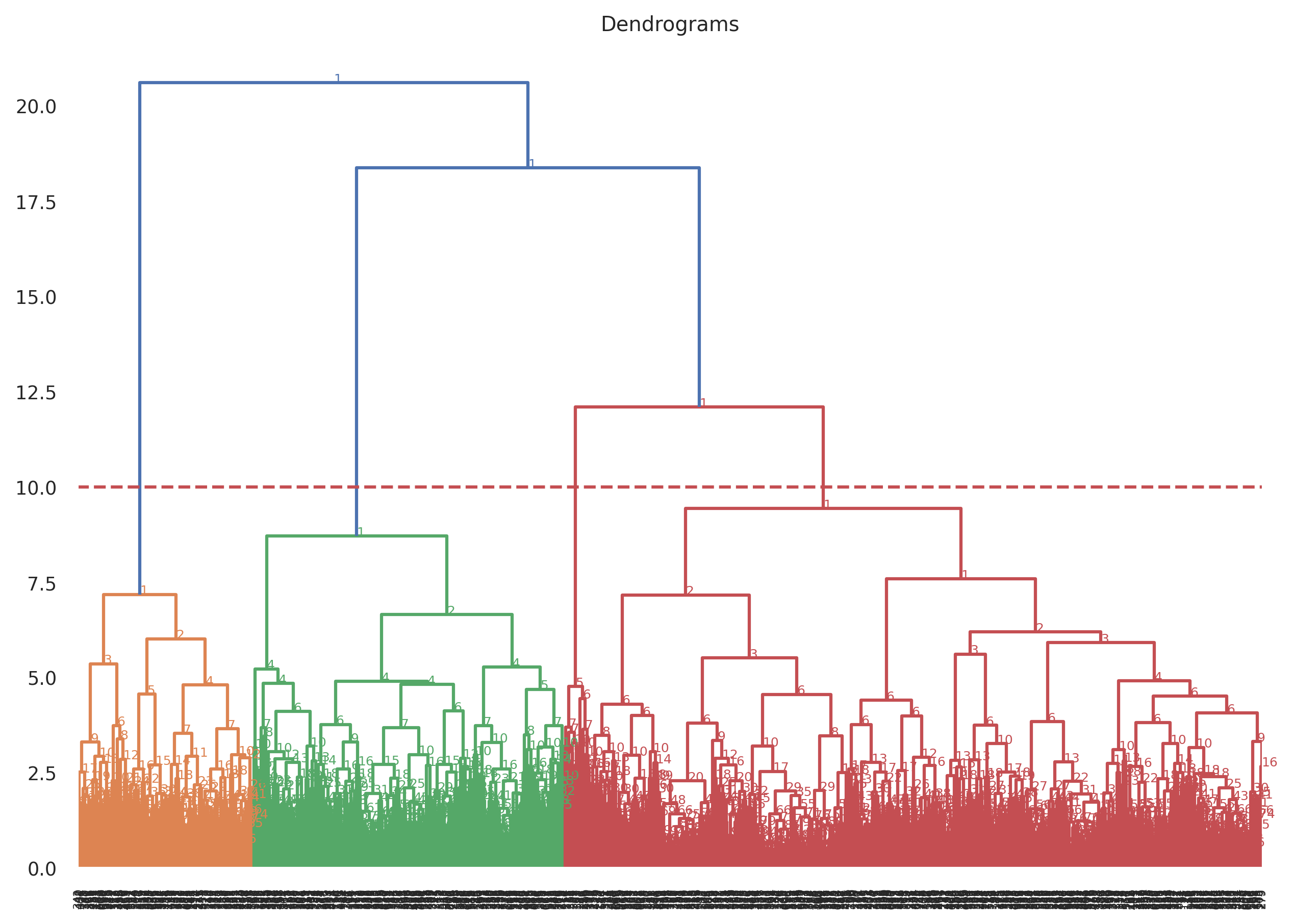}
	\includegraphics[scale=0.15]{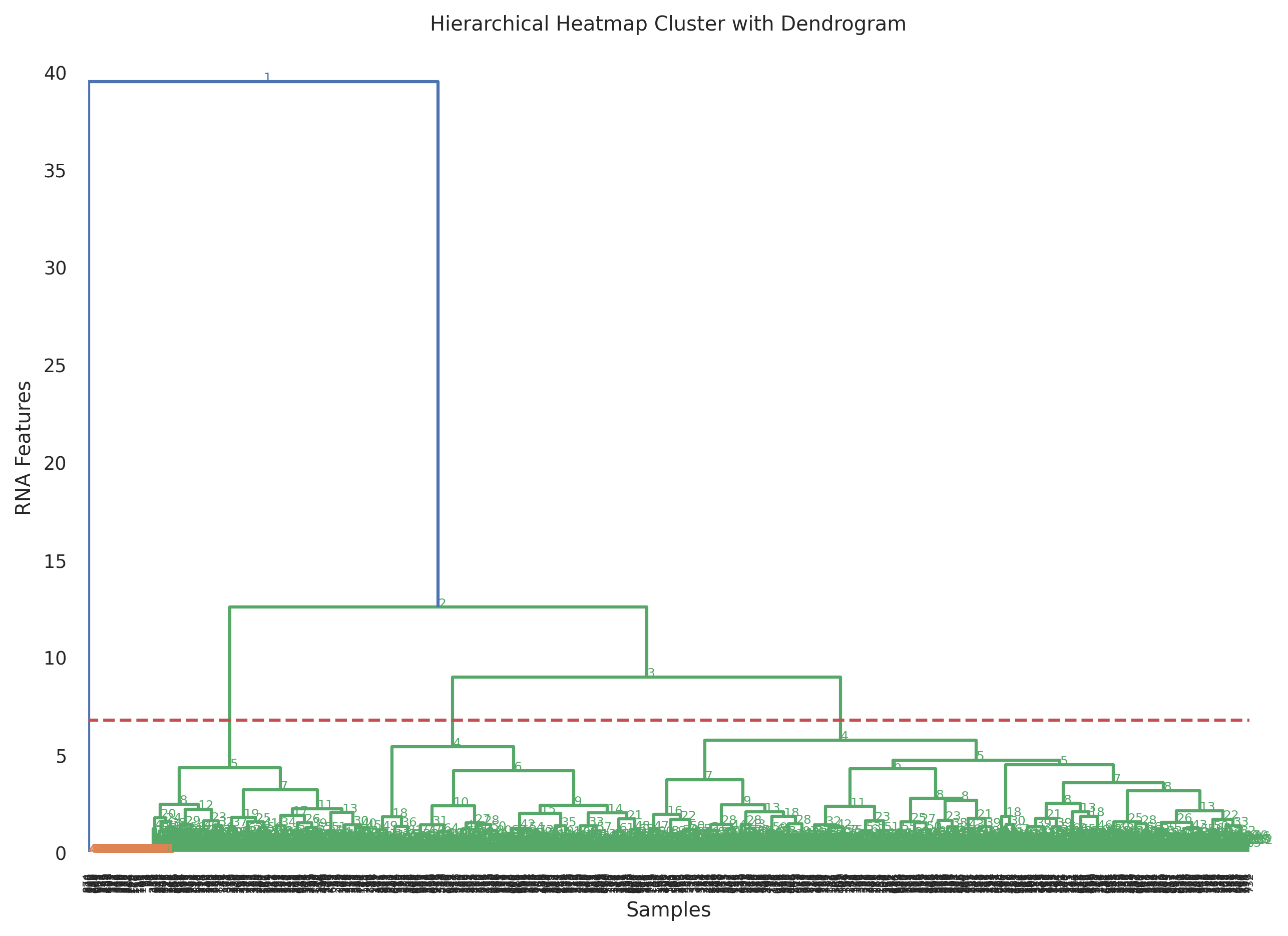}\\
	\includegraphics[scale=0.15]{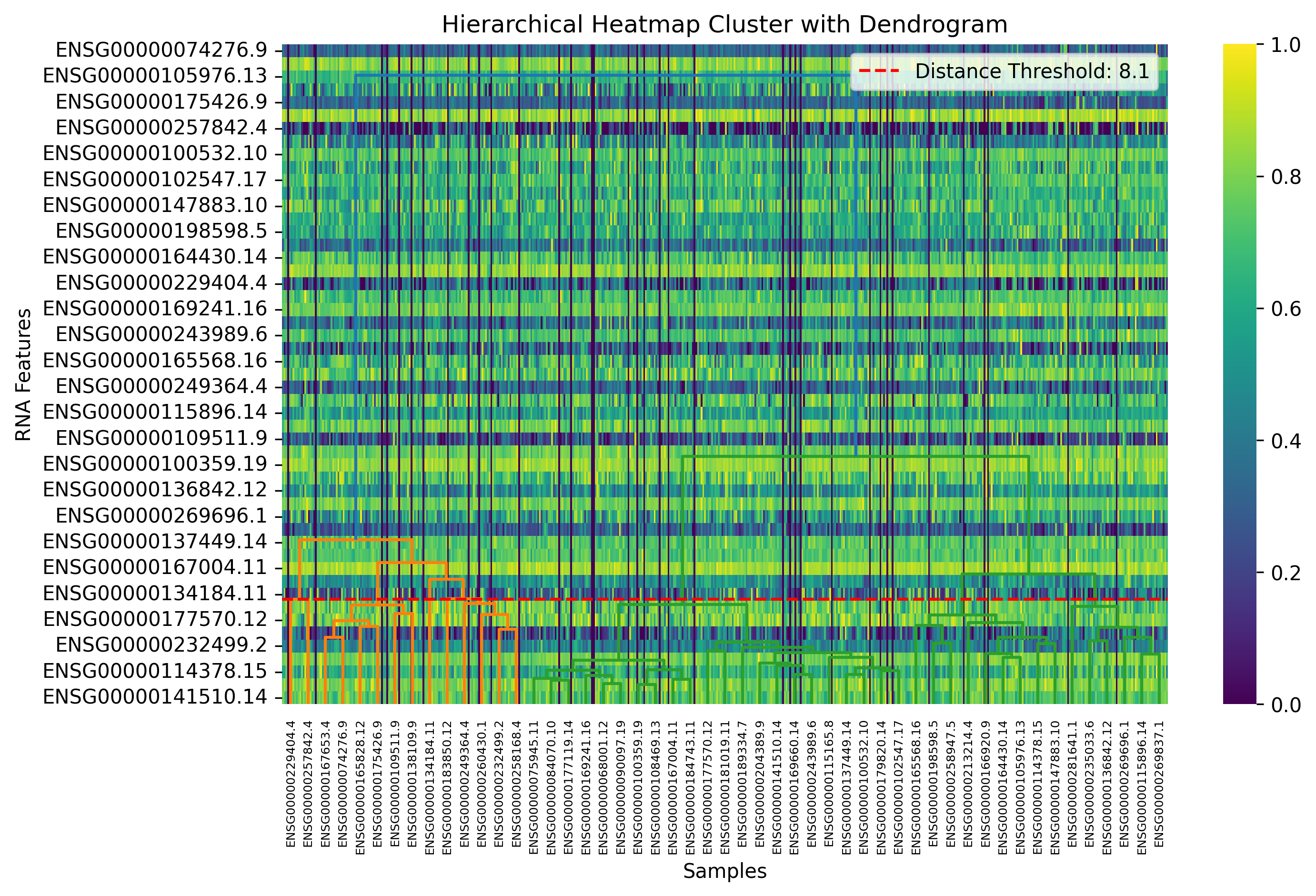}
	\includegraphics[scale=0.15]{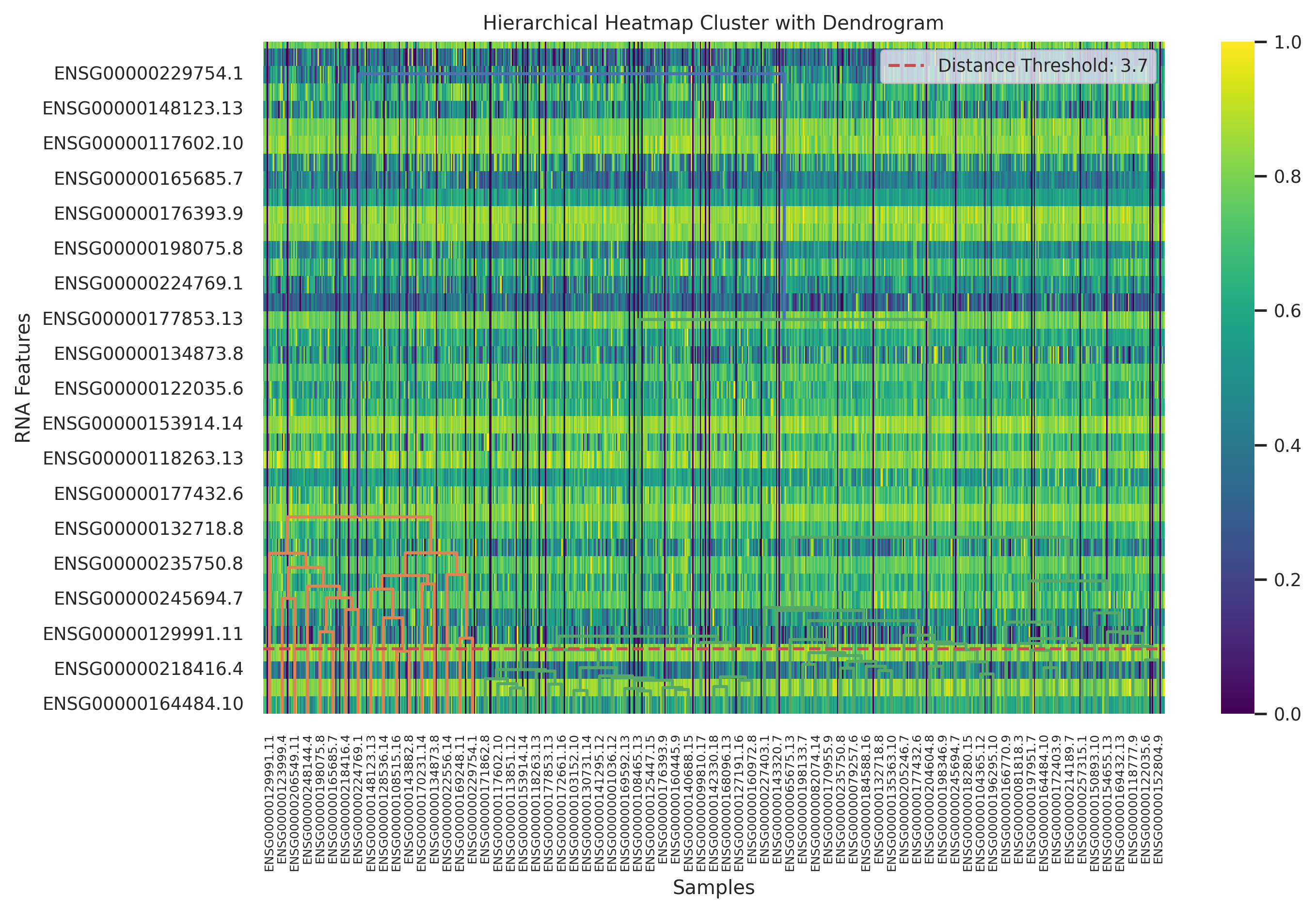}
	\includegraphics[scale=0.15]{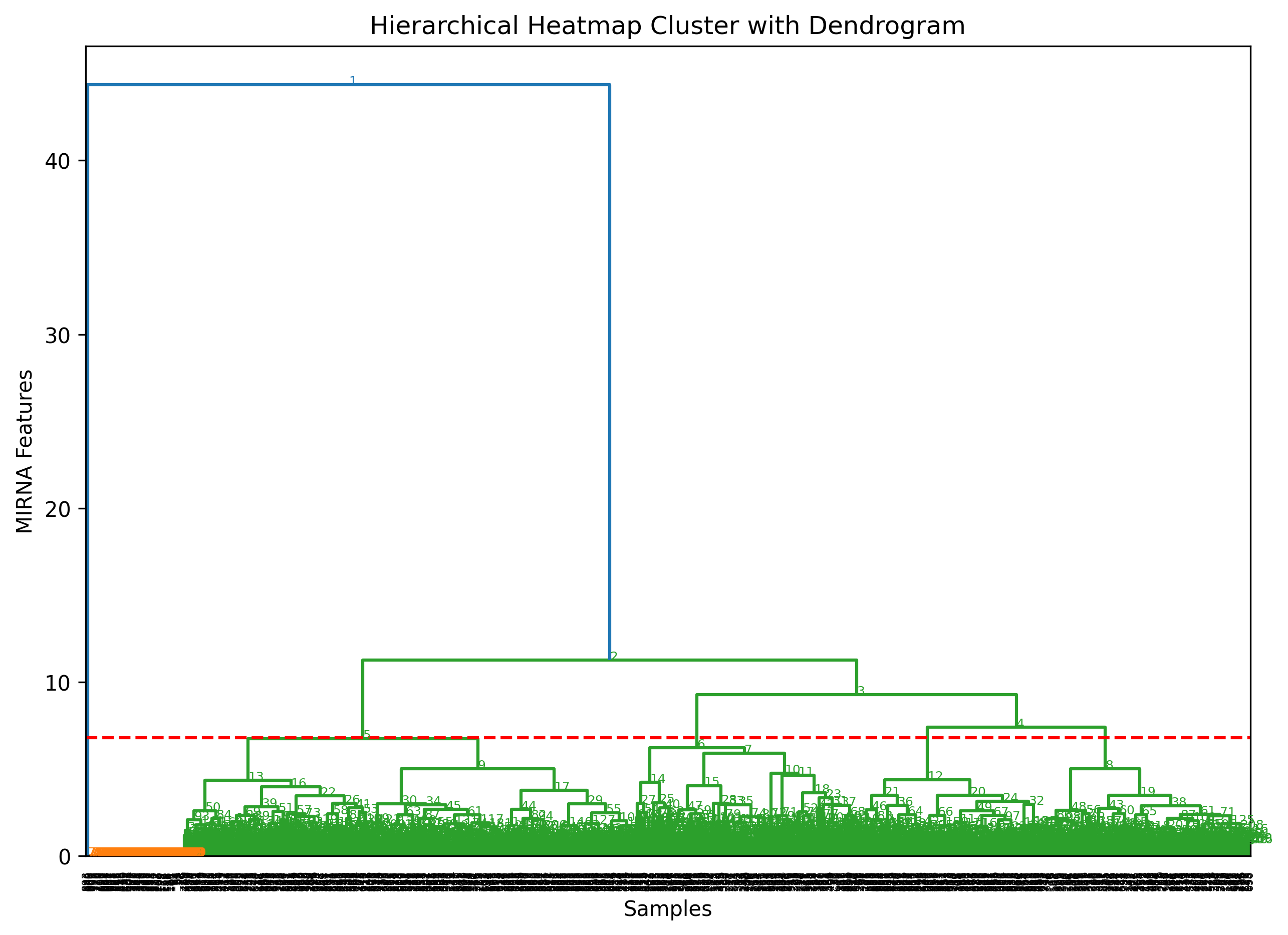}
	\caption{\textbf{Represents the Hierarchical Clustering Dendogram based pairwise feature selection and distance-based selection of features of each omic subsets based on  (a)-(c) DNA S$_{1}$ to S$_{3}$, (d)-(f) RNA S$_{1}$ to S$_{3}$, and (g)-(h) miRNA S$_{1}$ to S$_{3}$,} }
\end{figure*}
\begin{figure*}[!ht]
	\centering
	\begin{tabular}{c}
		\textbf{(a)} 
		\includegraphics[scale=0.12]{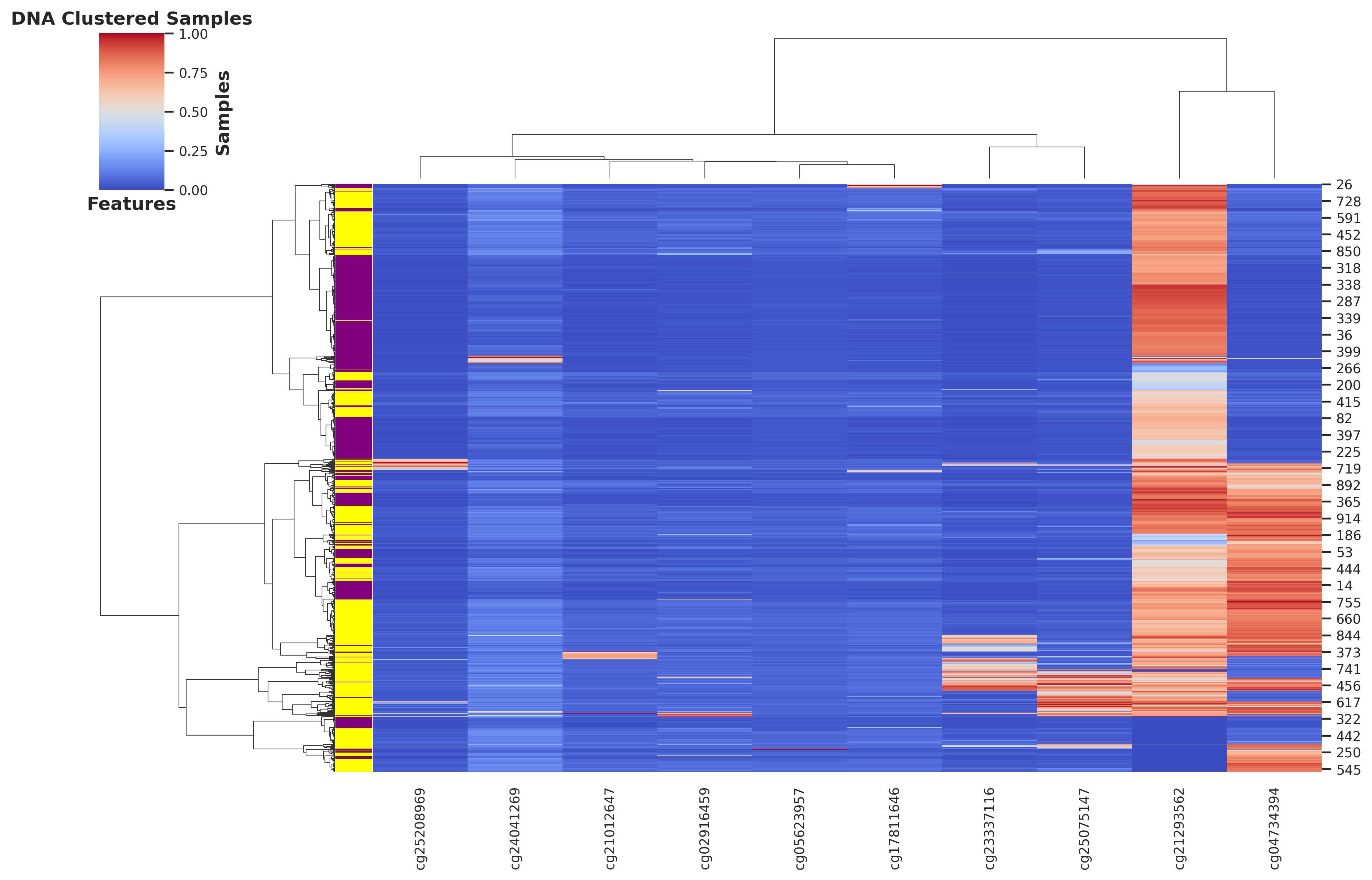}
		\textbf{(b)} 
		\includegraphics[scale=0.12]{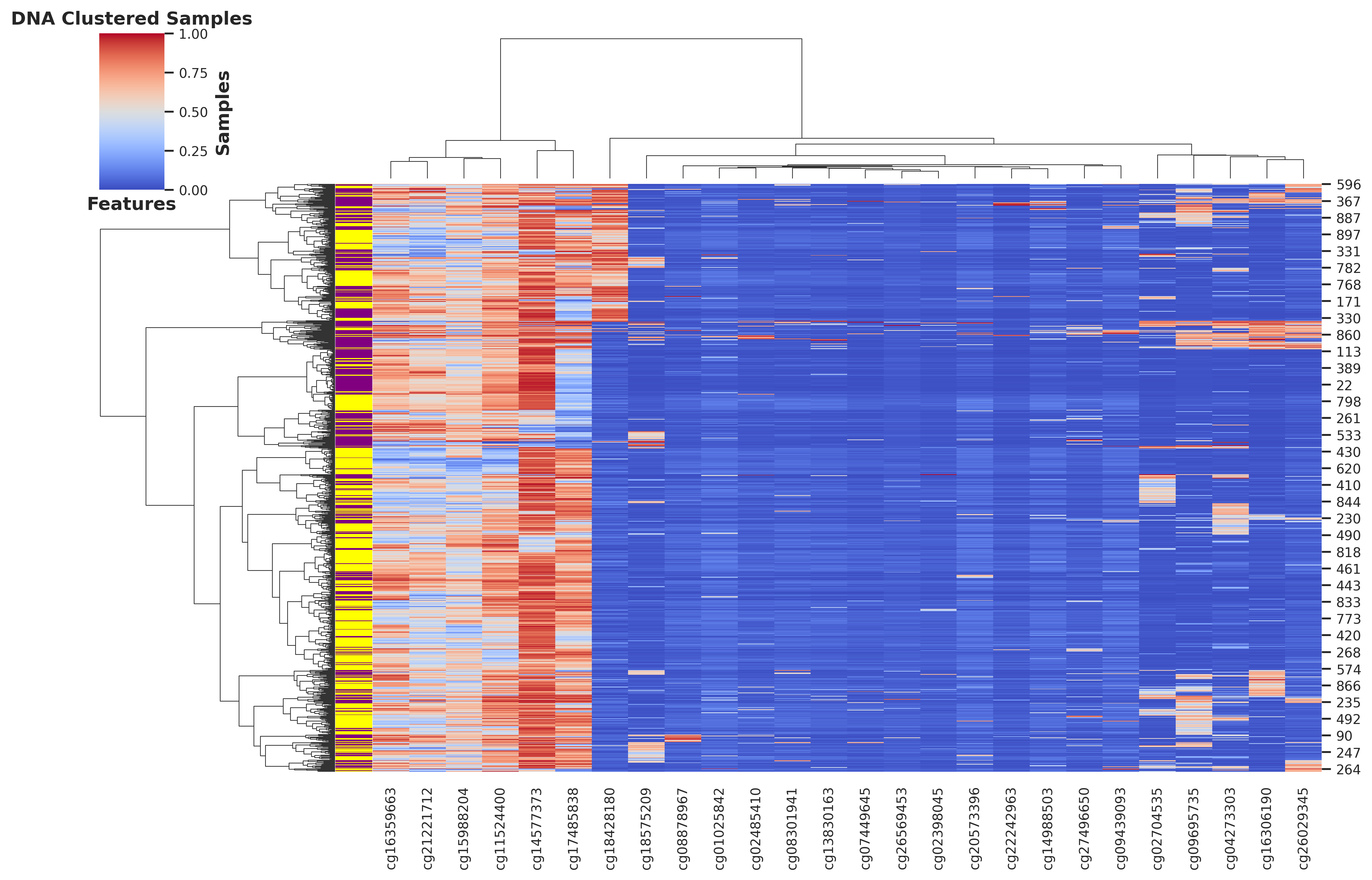}
		\textbf{(c)} 
		\includegraphics[scale=0.12]{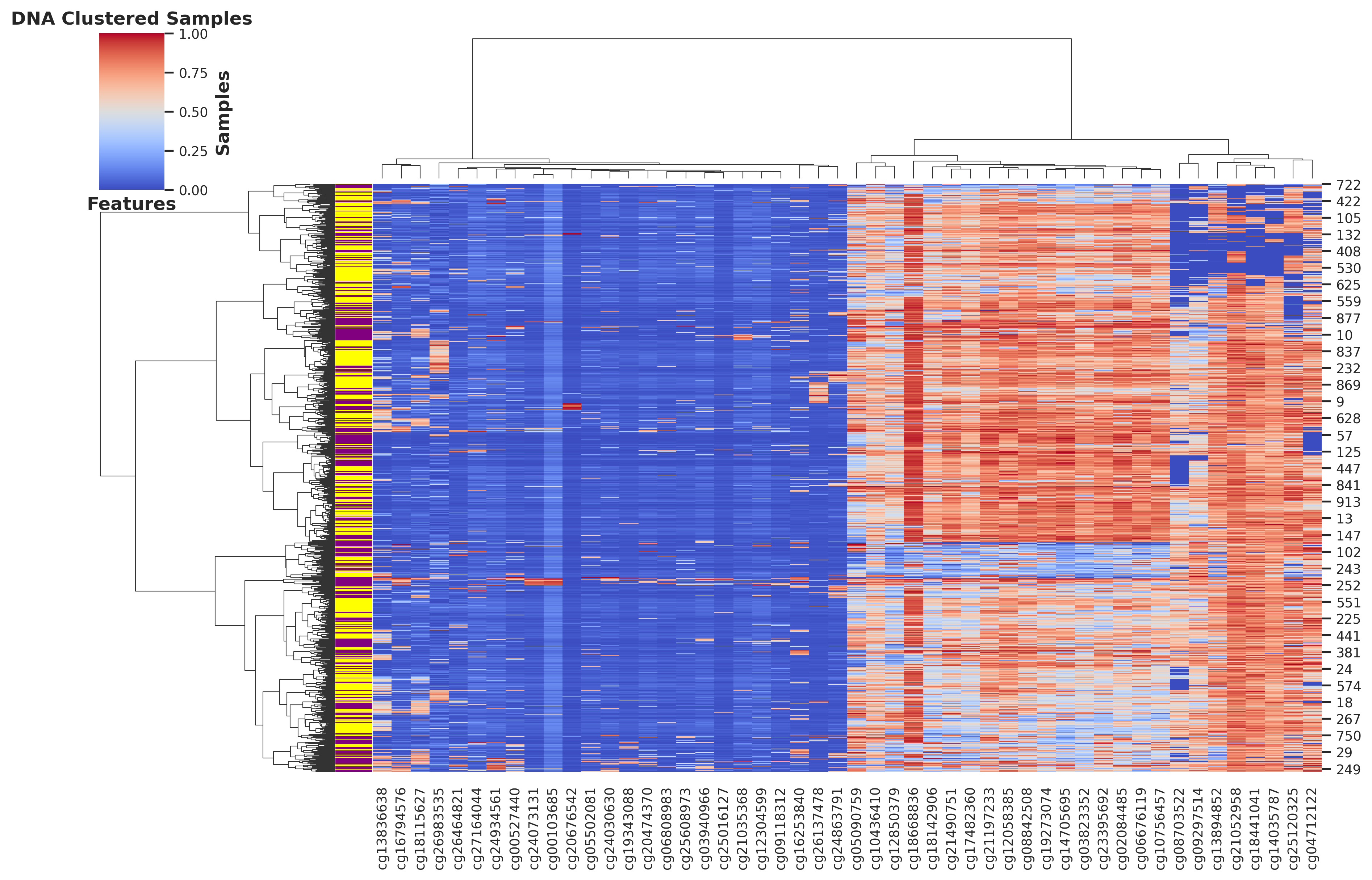} \\
		\textbf{(d)} 
		\includegraphics[scale=0.12]{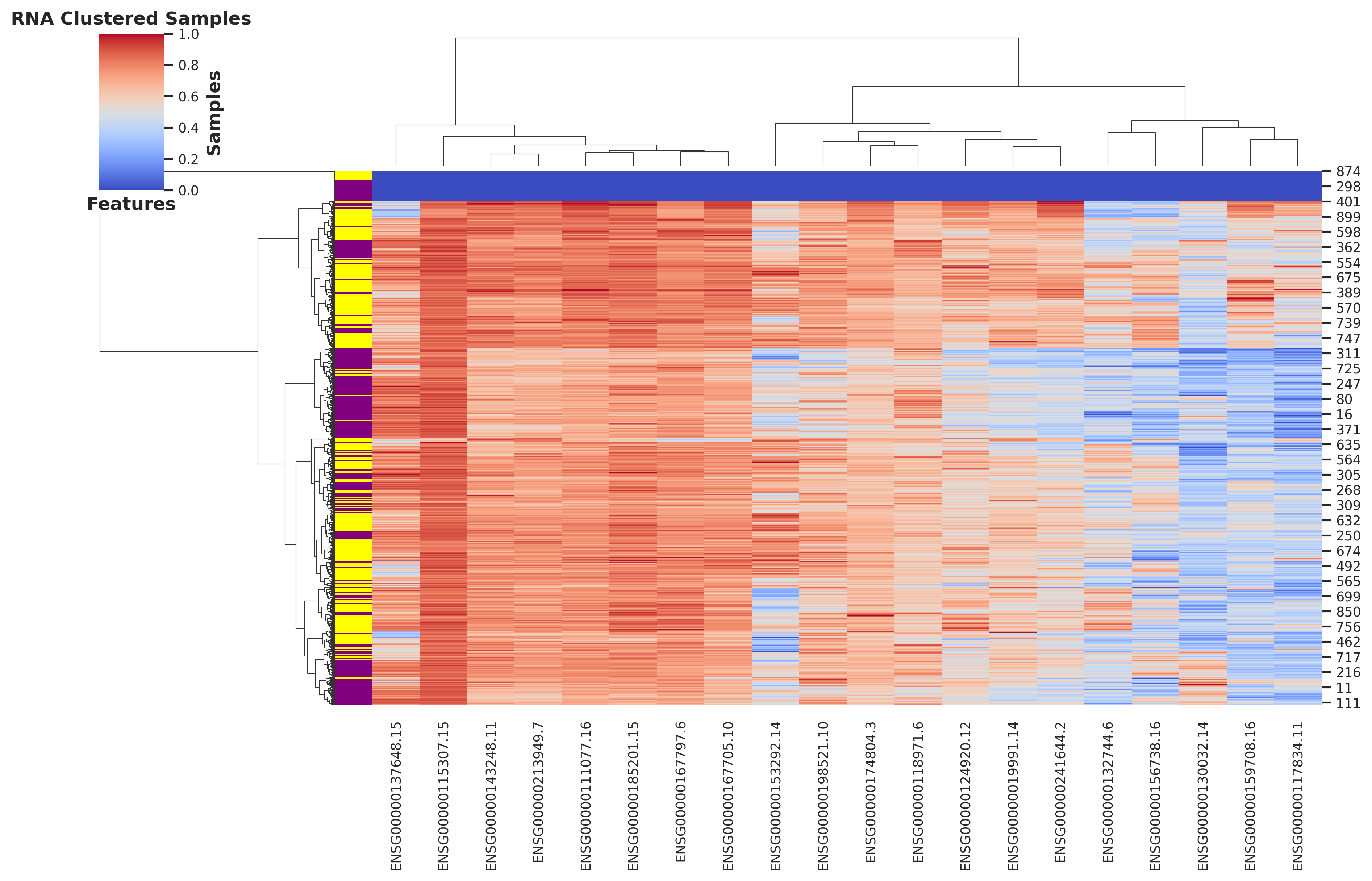} 
		\textbf{(e)} 
		\includegraphics[scale=0.12]{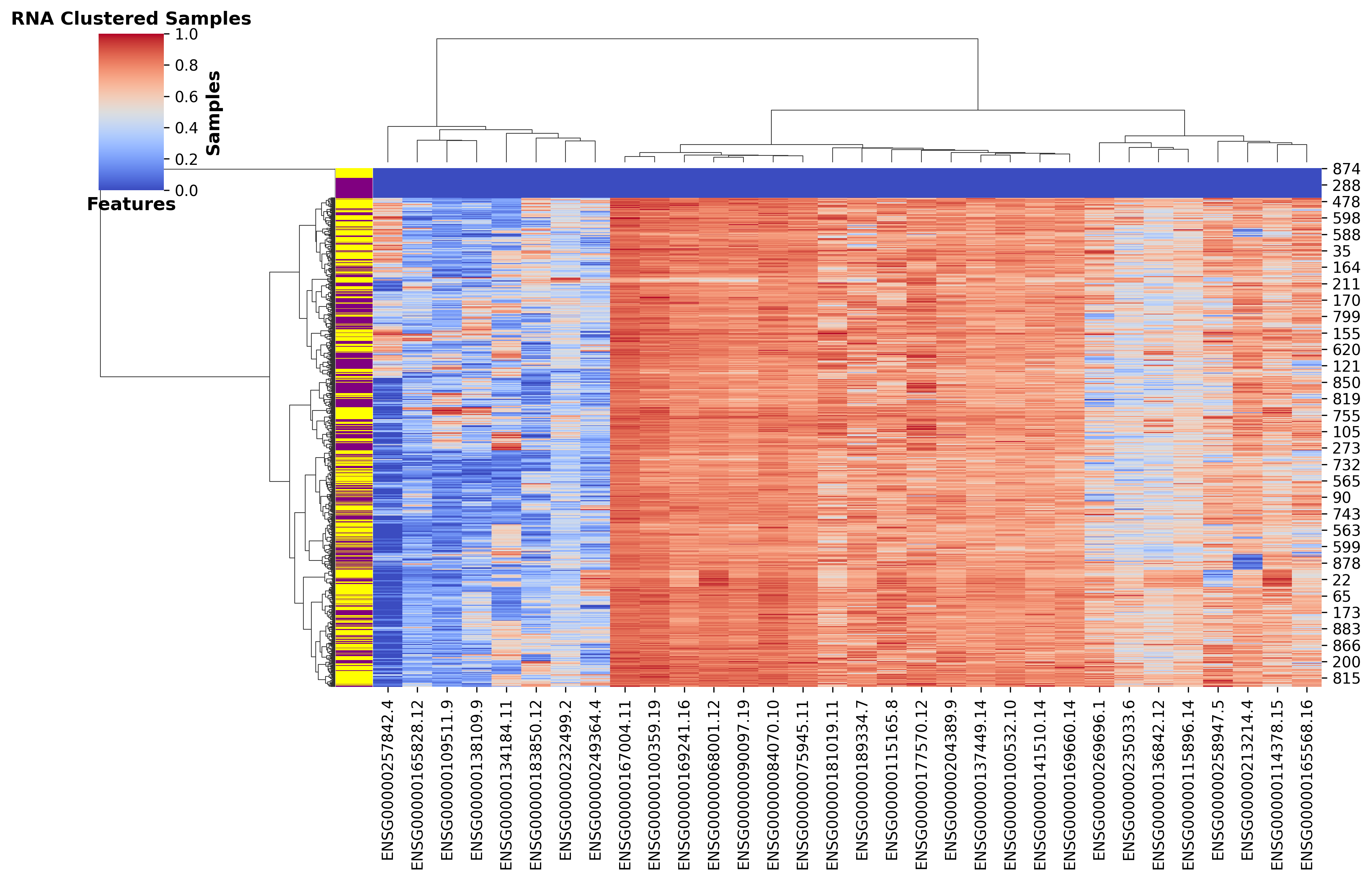}
		\textbf{(f)} 
		\includegraphics[scale=0.12]{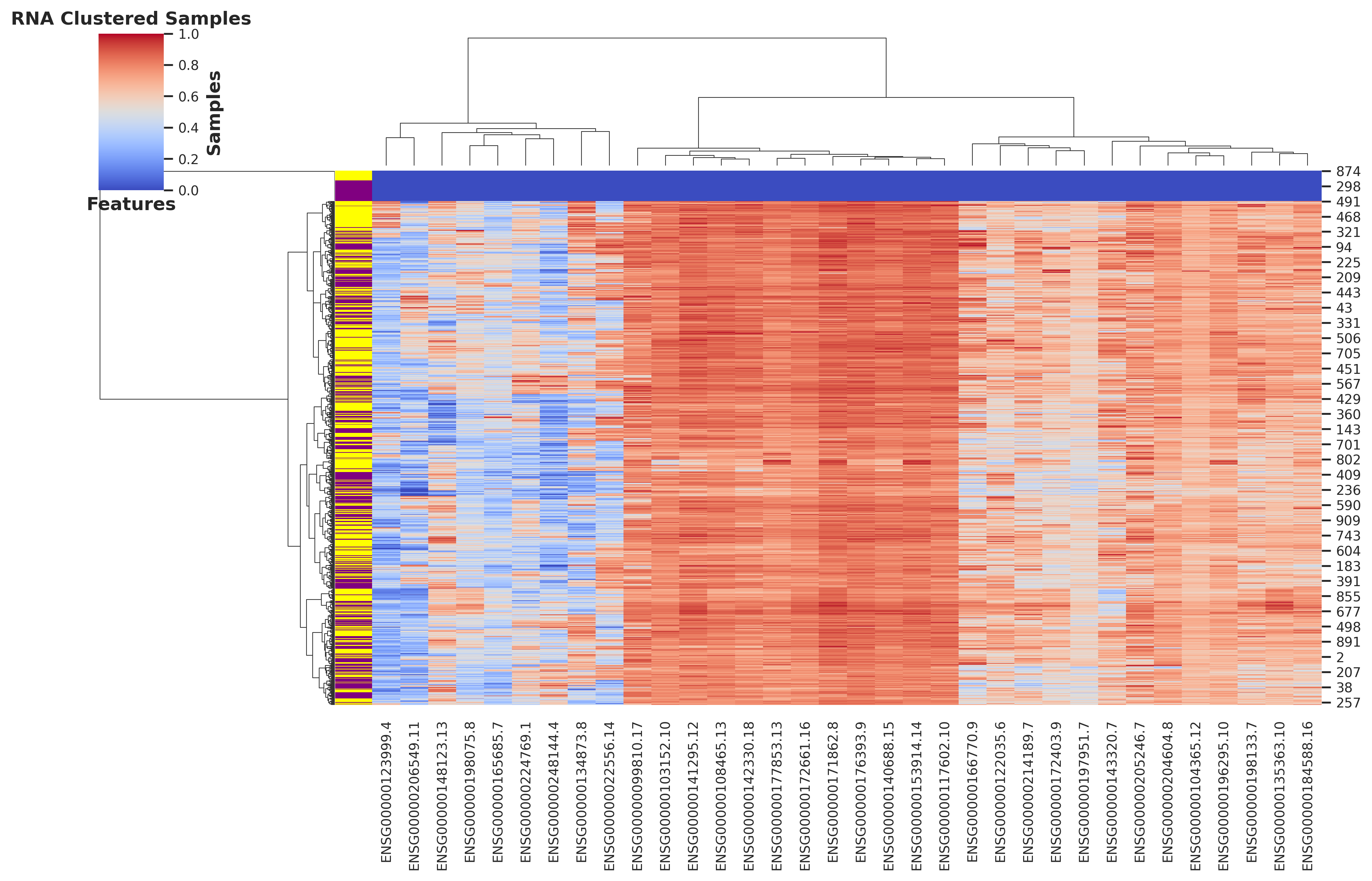}\\
		\textbf{(g)} 
		\includegraphics[scale=0.12]{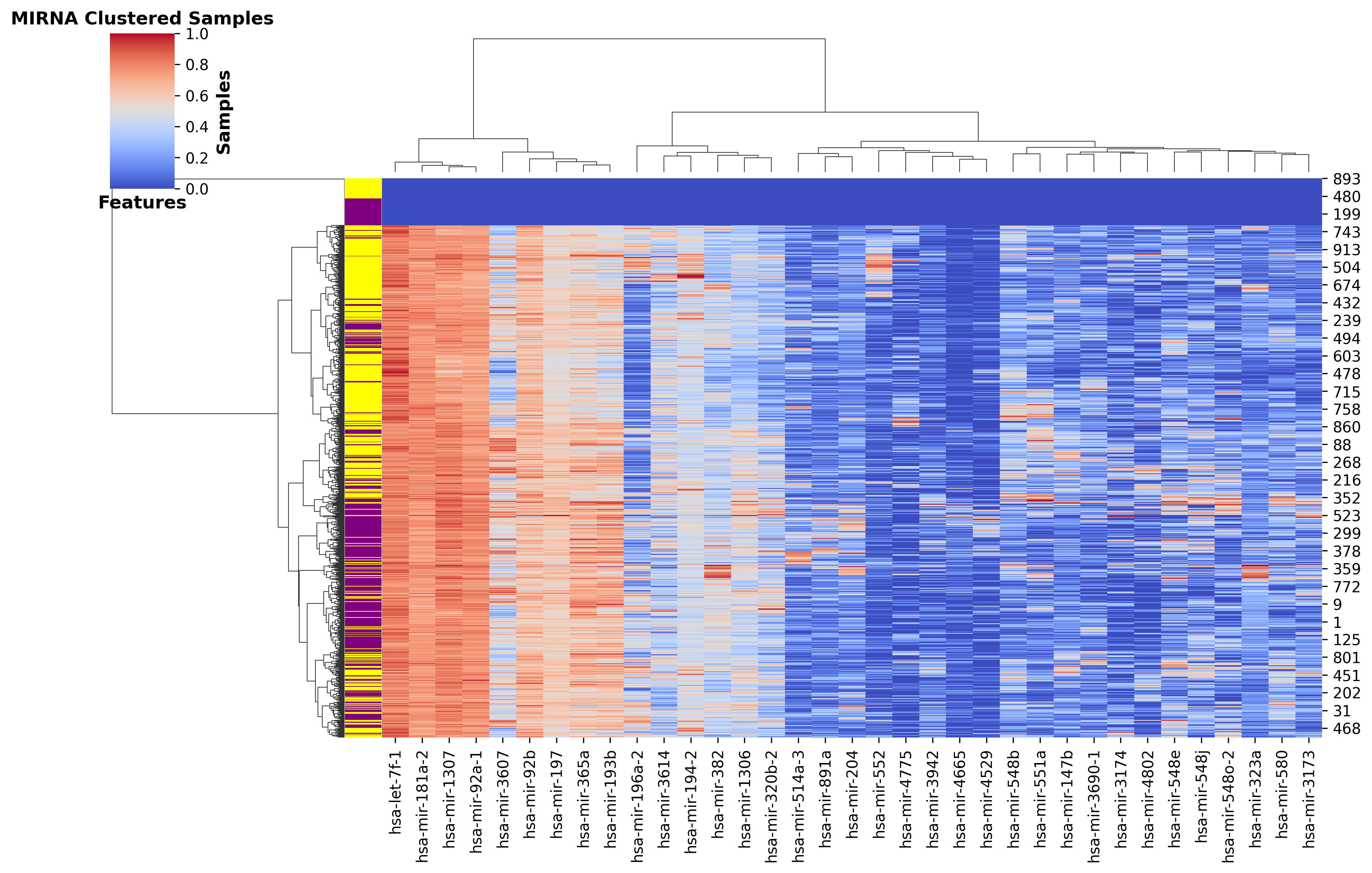}
		\textbf{(h)} 
		\includegraphics[scale=0.12]{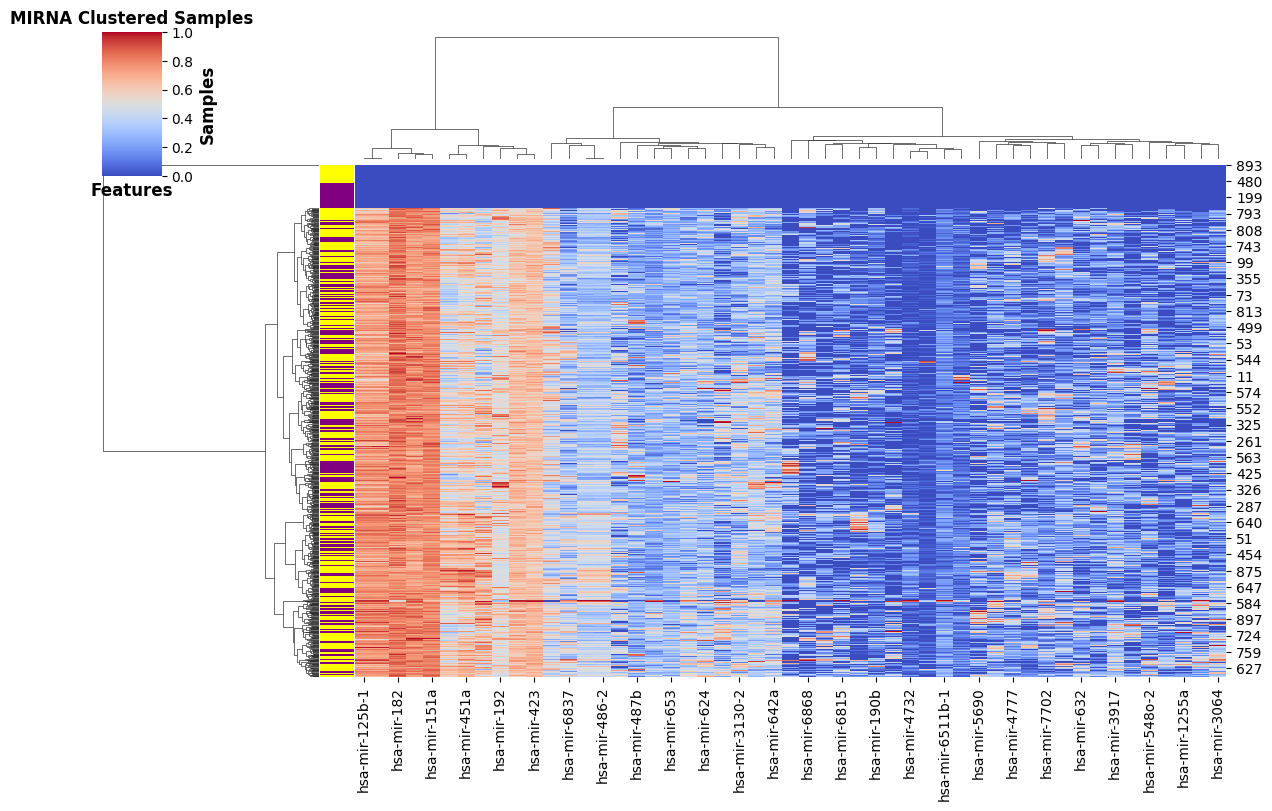}
	\end{tabular}
	\caption{\textbf{Represents the Hierarchical Clustering Heatmap of selected features of each omic subsets (a)-(c) DNA S$_{1}$ to S$_{3}$, (d)-(f) RNA S$_{1}$ to S$_{3}$, and (g)-(h) miRNA S$_{1}$ to S$_{3}$,} }
\end{figure*}

\begin{figure*}[!ht]
	\includegraphics[scale=0.15]{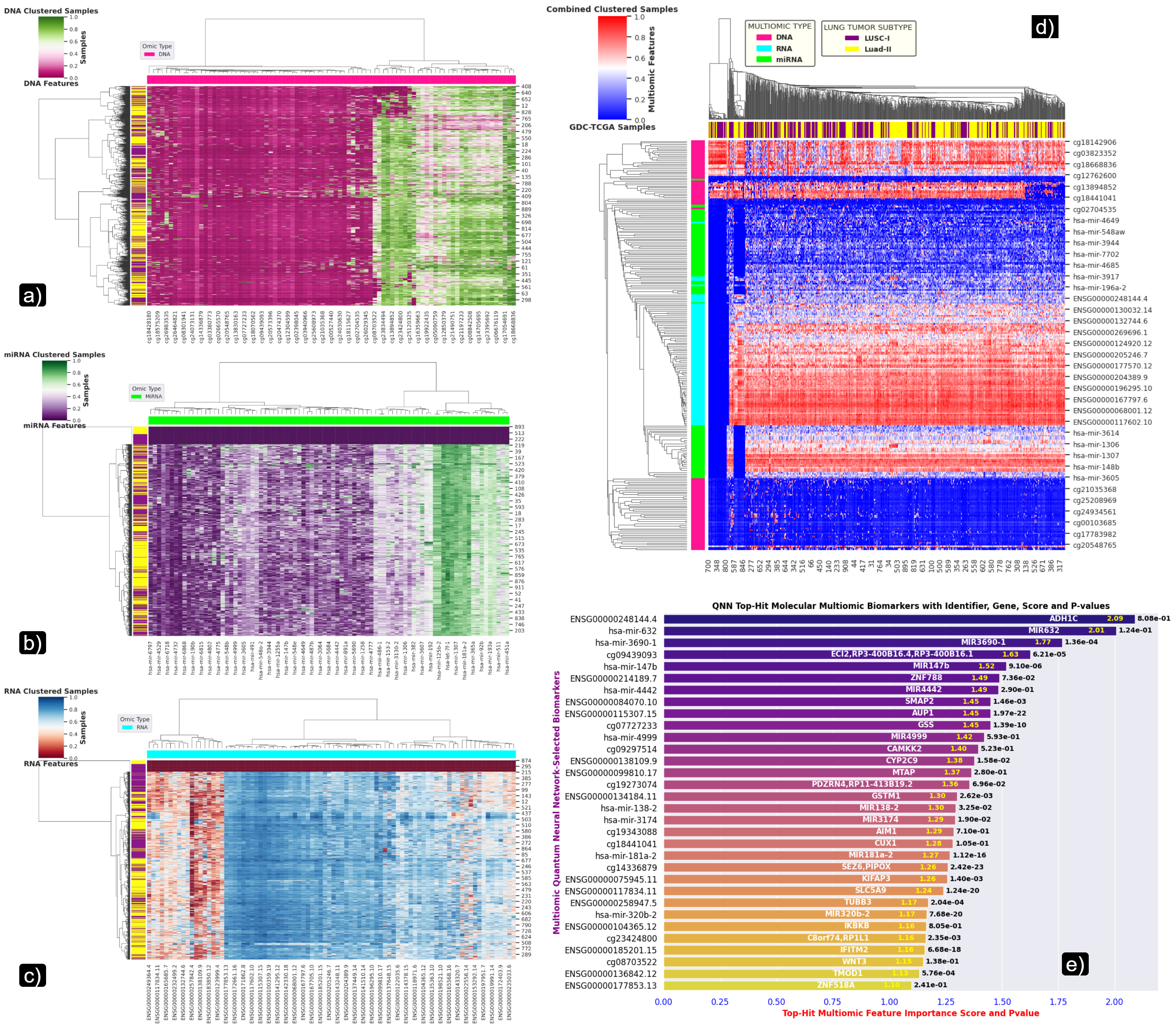}    
	\caption{\textbf{Hierarchical representation of single omics and integrated multi-omic} hierarchical clustering analyses and top-hit feature importance plot. Hierarchical clustering of emerging data results in the effective separation of patients’ groups. (A), whereas for DNAme (B) miRNA-seq (C) RNA-seq data and (D) Integrated all omics as Multi-Omic$_{256}$. Heatmaps illustrate  DNA methylation, gene expression,  miRNA, and patterns of lung feature specific-related genes in the LUAD subtype-II compared to the LUSC subtype-I lung dataset. In each heatmap, the cell lines were organized by hierarchical clustering, with labels colored in purple (subtype-I) and yellow for (subtype-II) highlighting LUSC and LUAD patients, respectively. (E) Outcome of top molecular features of integrated multi-omic$_{256}$ using the QNN1$_{256}$ model}
\end{figure*}

\begin{table*}[!ht]
	\caption{Performance Comparison of Different Feature Selection Methods Across Omics Data Types and Subsets with RF Classifier}
	\begin{scriptsize}
		\begin{tabular}{ |p{1.2 cm} | p{0.7 cm}|p{0.7 cm}|  p{0.9 cm}|  p{0.7 cm}|   p{0.7 cm}|p{0.7 cm}|  p{0.9 cm}|  p{0.7 cm}|  p{0.7 cm}|p{0.7 cm}|  p{0.9 cm}|  p{0.7 cm}|}
			\hline
			\multicolumn{5}{|c|}{DNAme-Subset1} & \multicolumn{4}{c|}{RNA-seq-Subset1} &
			\multicolumn{4}{c|}{miRNA-seq-Subset1} \\
			\hline
			Subset1	&	MI	&	PCA	&	Chi-sq	&	RF	& MI	&	PCA	&	Chi-sq	&	RF &	MI	&	PCA	&	Chi-sq	&	RF	\\
			\hline
			ACC	&	0.99	&	0.841	&	0.968	&	0.998	&	0.9	&	0.75	&	0.79	&	0.84	&	0.84	&	0.77	&	0.76	&	0.87	\\
			Precision	&	1.99	&	0.843	&	0.9677	&	0.99	&	0.91	&	0.75	&	0.76	&	0.84 &	0.88	&	0.79	&	0.77	&	0.9	\\
			Recall	&	2.99	&	0.874	&	0.97	&	0.998	 &	0.916	&	0.82	&	0.89	&	0.86	&	0.82	&	0.8	&	0.81	&	0.86	\\
			F1-Score	&	3.99	&	0.857	&	0.97	&	0.99	&	0.91	&	0.78	&	0.82	&	0.85	&	0.85	&	0.79	&	0.79	&	0.88\\
			\hline
			\multicolumn{5}{|c|}{DNAme-Subset2} & \multicolumn{4}{c|}{RNA-seq-Subset2} & \multicolumn{4}{c|}{miRNA-seq-Subset2}\\
			\hline
			Subset2	&	MI	&	PCA	&	Chi-sq	&	RF	 &	MI	&	PCA	&	Chi-sq	&	RF &	MI	&	PCA	&	Chi-sq	&	RF	\\
			\hline
			ACC	&	0.998	&	0.75	&	0.94	&	0.99 &	0.85	&	0.78	&	0.69	&	0.88	&	0.77	&	0.66	&	0.63	&	0.8		\\
			Precision	&	0.99	&	0.76	&	0.95	&	0.99	&	0.86	&	0.7	&	0.73	&	0.89	&	0.79	&	0.68	&	0.61	&	0.81	\\
			Recall	&	0.99	&	0.8	&	0.93	&	0.998	&	0.88	&	0.78	&	0.71	&	0.91 &	0.78	&	0.72	&	0.88	&	0.82	\\
			F1-Score	&	0.99	&	0.78	&	0.94	&	0.99	&	0.87	&	0.74	&	0.72	&	0.9	&	0.79	&	0.7	&	0.72	&	0.81	\\
			\hline
			\multicolumn{5}{|c|}{DNAme-Subset3} & \multicolumn{4}{c|}{RNA-seq-Subset3} & \multicolumn{4}{c|}{}\\
			\hline
			Subset3	&	MI	&	PCA	&	Chi-sq	&	RF	&	MI	&	PCA	&	Chi-sq	&	RF	& & & &\\
			\hline
			ACC	&	0.996	&	0.881	&	0.942	&	0.995	&	0.79	&	0.68	&	0.56	&	0.85 & - & -& -& -	\\
			Precision	&	0.997	&	0.879	&	0.952	&	0.996	&	0.78	&	0.66	&	0.55	&	0.85	& - & -& -& -	\\
			Recall	&	0.995	&	0.908	&	0.942	&	0.996	&	0.87	&	0.84	&	0.99	&	0.88	& - & -& -& -	\\
			F1-Score	&	0.996	&	0.893	&	0.947	&	0.996 &	0.82	&	0.74	&	0.71	&	0.87	& - & -& -& -	\\
			\hline
		\end{tabular}
	\end{scriptsize}
\end{table*}

\subsection{Experimental Hardware and Software}
\noindent We implemented the experiments on the Gilbreth GPU cluster at Purdue University's Research Computing and Data Systems (RCAC) with a single server, one GPU node, 8 cores per node, an A10 GPU, and 512 GB memory per node \cite{15}. The experiments were performed using TensorFlow 2.7.0 in Python 2.8.0. The quantum machine learning library PennyLane (version 0.28.0) by Xanadu Inc. was employed, with dependencies including appdirs, autograd, autoray, cachetools, networkx, numpy, pennylane-lightning, requests, retworkx, scipy, semantic-version, and toml. Additionally, the fundamental array computing package NumPy (version 1.20.3), the Python plotting package Matplotlib (version 3.4.3), the machine learning and data mining library scikit-learn (version 1.4.2), the statistical data visualization library Seaborn (version 0.11.2), and the data analysis package pandas (version 1.3.4) were utilized in the study.

\subsection{Performance of Feature Engineering Process}
Statistical analyses were conducted using a t-test to calculate the p-value significant and in-significant values for LUSC and LUAD subtypes. Then, each modality was divided into subsets, meaning features were divided into subsets column-wise based on the obtained p-values. For high-dimensional features in DNAme and RNA-seq modalities, the datasets were divided into three subsets based on the significance level: two subsets with p-values less than 0.05 and one subset with p-values greater than 0.05. However, the miRNA-seq modality was divided into two subsets: one included p-values less than 0.05 and the second subset had p-values greater than 0.05 to 0.9. The process of data engineering is shown in Fig 2. Table II depicts the range of p-values used for selecting the most significant and least significant features across three omics. Table III presents the feature selection process using p-value-based data across DNAme (OMIC1), RNA-seq (OMIC2), and miRNA-seq (OMIC3) omics, detailing the number of features selected, feature selection methods employed, and significance thresholds (Feature cluster) (FS Clt).
Table IV provides a summary of p-value-based encoding and simulations conducted across different omic sets: DNAme, RNA-seq, and miRNA-seq . Each set (denoted as $S_{1}$, $S_{2}$, $S_{3}$, and $S_{4}$) shows the threshold p-value used for feature selection and the corresponding number of features selected for each omic type. The table also includes aggregated results for multi-omic data integration, highlighting the total number of features retained across all omics.

\subsection{Performance of Feature Selection Process with RF Classifier}
Further, this process is applied to discover the best and unique features from each omic modality (DNAme, RNA-seq, and miRNA-seq) based on their subsets. The applied feature selection process based on p-value, the OMIC$_{DNAme}$,  OMIC$_{RNA-seq}$, and OMIC$_{miRNA-seq}$ was divided into three subsets. The initial two subsets represent less than 0.05 and the third subset represents a greater than 0.05 p-value. The four machine learning models are applied to select the features. (DNAme subset combined Total 85= DNAme$_{Subset1}$(10) + DNAme$_{Subset2}$(50) + DNAme$_{Subset3}$(25) features are combined to concatenate the final subset of OMIC$_{DNA}$.

RNA subset combined Total 85= RNA-seq$_{Subset1}$(20) + RNA-seq$_{Subset2}$(32) + RNA-seq$_{Subset3}$(34) features are combined to concatenate the final subset of OMIC$_{RNA}$. miRNA-seq subset combined Total 85= miRNA-seq$_{Subset1}$(35) + miRNA-seq$_{Subset2}$(51) features are combined to concatenate the final subset of OMIC$_{miRNA}$. These four methods were used to select the features and then classified with a random forest classifier to see the performance of selected features by each method as shown in Table V. This table presents the performance metrics (Accuracy, Precision, Recall, and F1-Score) of four feature selection models—MI, PCA, Chi-sq, and RF—applied to different subsets of DNAme, RNA-seq, and miRNA-seq data. The Random Forest (RF) model consistently shows the highest performance across most subsets, particularly in DNA, where it nearly achieves perfect scores. MI demonstrates competitive performance across all omics data types and subsets, especially in re- call and F1-Score. Chi-sq generally performs moderately well, while PCA tends to have the lowest performance across the board.

\subsection{\textbf{Performance of Multi-Omic (DNAme/RNA-seq/miRNA-seq) Integration of Subtype-I and Subtype-II with MQML-QNN }}
In this MQML-LungSC QNN framework, we integrated multi-omic DNA-seq/RNA-seq/miRNA-seq cohorts for lung subtype-I and subtype-II diagnostic classification with three different dimensions of features and n qubits encoding. The QNN models are named QNN$_{1}$, QNN$_{2}$ and QNN$_{3}$ based on different settings and hyper-parameters as shown in Table I. 

Among these models, in the QNN1$_{256}$ model, we have used qubit$_{8}$ to encode 256 features using amplitude encoding. Then, a layer is repeated consisting of single-qubit gates (Rz, Ry, Rz) Parameterized rotations 8 \[ R(\theta, \phi, \lambda) = R_z(\lambda) R_y(\theta) R_z(\phi) \] on each qubit, followed by a linear entanglement of 7 Controlled-Z (CZ) gates between adjacent qubits. Finally, the measurement is performed on all qubits, followed by classical neurons and a dense layer for the classification using ADAM optimizer. This QNN1$_{256}$ model, provided the highest classification performance, with an accuracy score of 0.90. In this model, for the LUAD$_{Subtype-II}$ class, the precision, recall, and F1 scores were 0.89, 0.94, and 0.92, respectively, whereas for the LUSC$_{Subtype-I}$ class, the precision, recall, and F1 score were 0.92, 0.85, and 0.88 respectively for the testing multi-omic dataset.

\begin{table}[!ht]
	\begin{scriptsize}
		\caption{Performance of QNN models on differentiating LUAD vs LUSC with different number of features.}
		\begin{tabular}{ |p{1.5 cm} | p{1.0 cm}|p{1.0 cm}|  p{1.0 cm}|  p{1.0 cm}| p{1.0 cm}| p{1.0 cm}|}
			\hline
			\multicolumn{1}{|c|}{Models} & \multicolumn{2}{|c|}{QNN-256} & \multicolumn{2}{|c|}{QNN-64} & \multicolumn{2}{|c|}{QNN-32}  \\
			\hline
			Metrics	&	LUSC	&	LUAD	&	LUSC	&	LUAD	&	LUSC	&	LUAD	\\
			\hline
			Precision	&	0.92	&	0.89	&	0.85	&	0.86	&	0.83	&	0.88	\\
			Recall	&	0.85	&	0.94	&	0.81	&	0.89	&	0.85	&	0.87	\\
			F1-Score	&	0.88	&	0.92	&	0.83	&	0.88	&	0.84	&	0.87	\\
			Class	&	80	&	103	&	79	&	104	&	79	&	104	\\
			Accuracy	&	0.9	&	0.9	&	0.86	&	0.86	&	0.85	&	0.85	\\
			\hline
		\end{tabular}
	\end{scriptsize}
\end{table}

\begin{figure}[!ht]
	\includegraphics[scale=0.38]{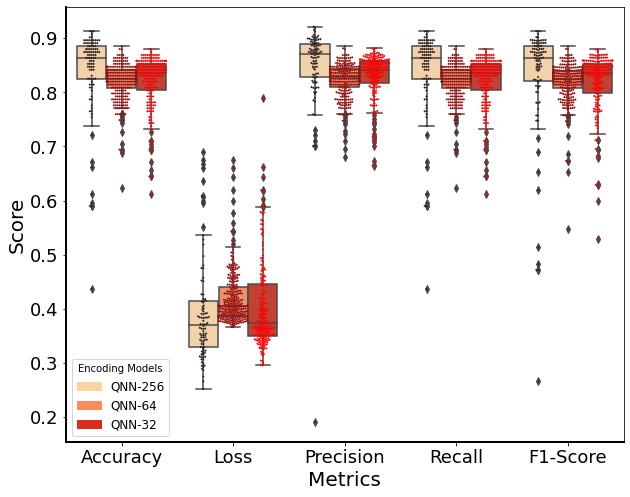}    
	\caption{\textbf{Compairson of QNN models with different  encoding features on integrated multi-omic data based on different classification metrics.}}
\end{figure}

\begin{figure*}
	\centering      
	\includegraphics[scale=0.43]{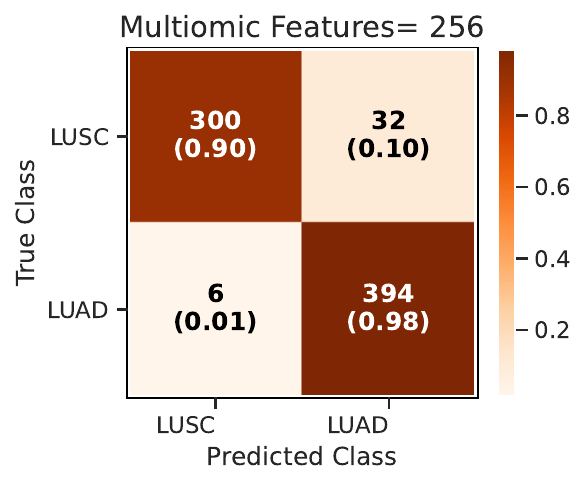}
	\includegraphics[scale=0.43]{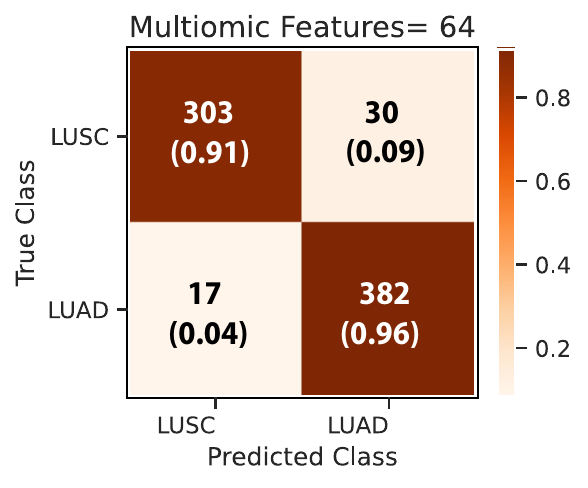}
	\includegraphics[scale=0.43]{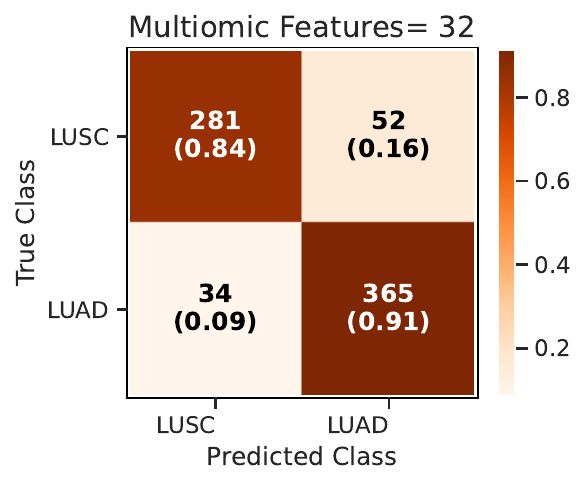}
	\includegraphics[scale=0.36]{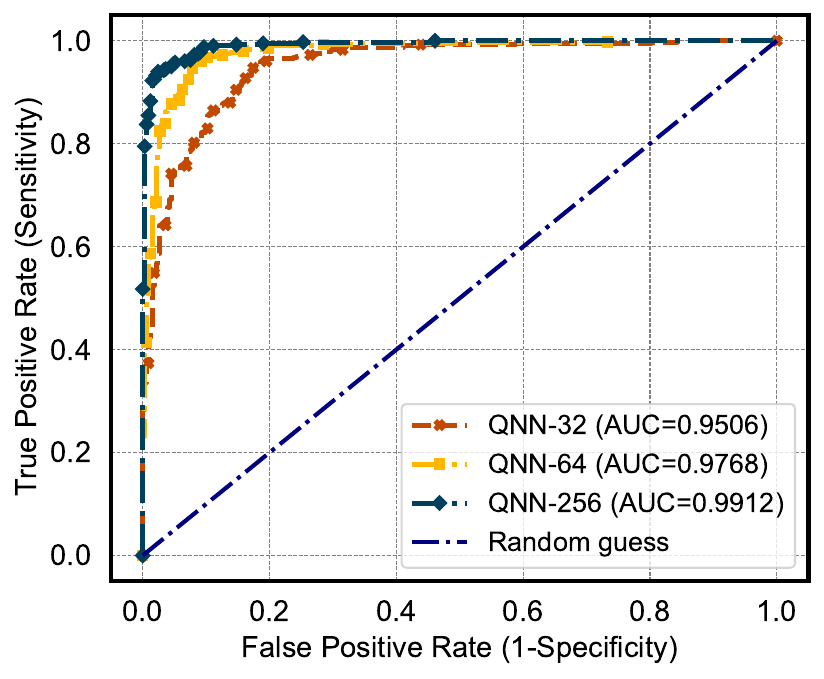} 
	
	\caption{\textbf{Represents the performance of QNN model on (a) 256, (b) 64, (c) 32 features, (d) AUC-ROC on the training dataset respectively} Visualization of Confusion Matrix on training dataset respectively}
	
\end{figure*}

\begin{figure*}
	\centering        
	\includegraphics[scale=0.43]{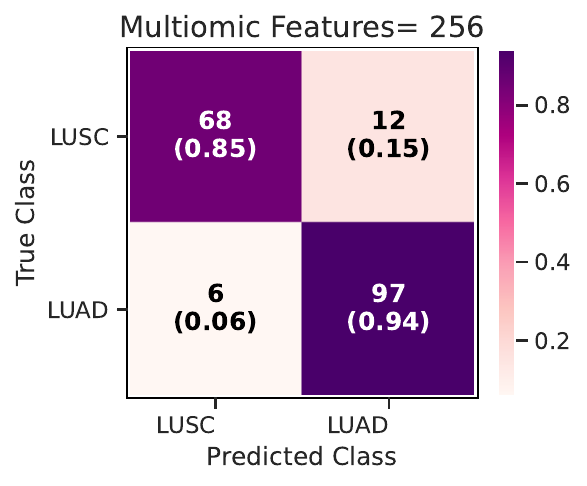}
	\includegraphics[scale=0.43]{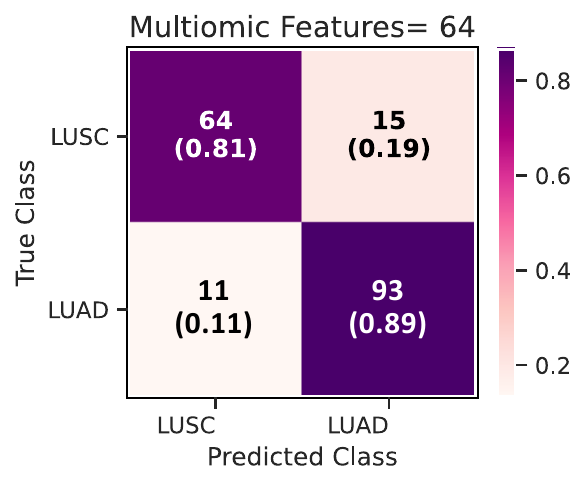}
	\includegraphics[scale=0.43]{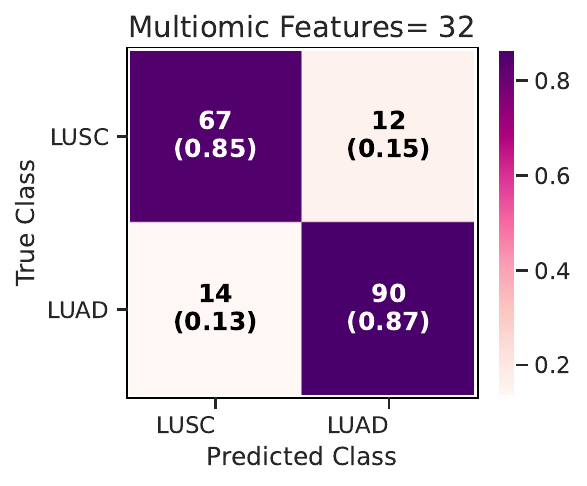}
	\includegraphics[scale=0.36]{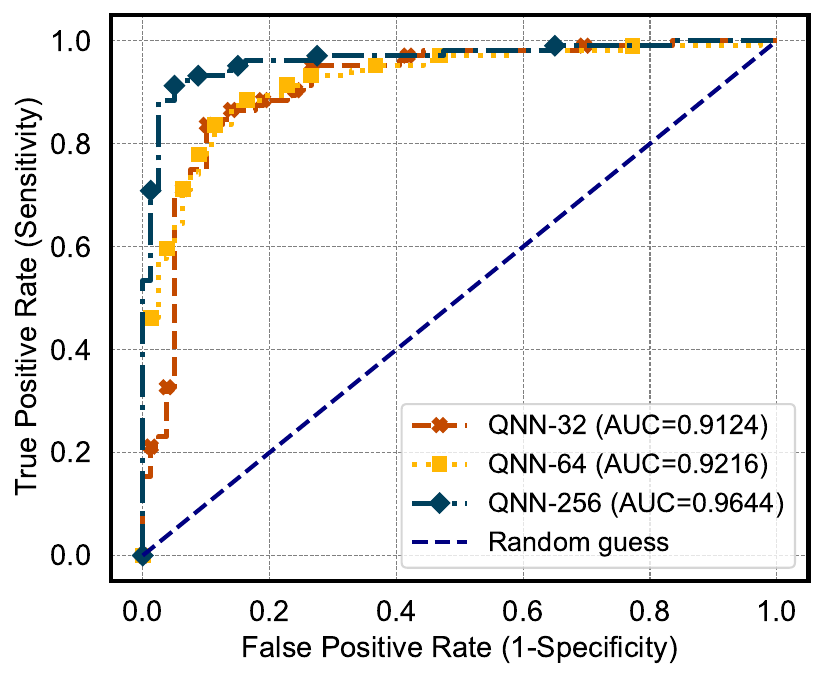}
	
	\caption{\textbf{Represents the performance of QNN model on (a) 256, (b) 64, (c) 32 features (d) AUC-ROC on testing dataset respectively} Visualization of Confusion Matrix on testing dataset respectively}
	
\end{figure*}

The QNN2$_{64}$ model demonstrated secondary predictive performance with an accuracy of 0.86. In the QNN2$_{64}$ model circuit, qubit$_{6}$ used to encode 64 features with amplitude encoding are employed within the feature map and apply 6 rotation gates with 5 CZ gates and a classical dense layer. In the context of the QNN2$_{64}$ model, the precision, recall, and F1 score values for the LUAD$_{Subtype-II}$ class were  0.86, 0.89, and 0.88, respectively. In contrast, for the LUSC$_{Subtype-I}$ class, the corresponding  precision, recall, and F1 scores are 0.85, 0.81, and 0.83, respectively.

Similarly, the QNN3$_{32}$ model exhibited improved performance with an accuracy of 0.85. In the QNN3$_{32}$ model circuit qubit$_{5}$ is used to encode 32 features with amplitude encoding and apply 5 rotation gates with 4 CZ gates using a classical dense layer. In the context of the QNN3$_{32}$ model, the precision, recall, and F1 score values for the LUAD$_{Subtype-II}$ class were 0.88, 0.87, and 0.87 respectively. In contrast, for the LUSC$_{Subtype-I}$ class, the corresponding values for precision, recall, and F1 score are 0.83, 0.85, and 0.84 respectively.

Each subset of modality was combined to make a feature of the Multi-Omic dataset (DNA subset combined Total  85= DNAme$_{Subset1}$(10) +DNAme$_{Subset2}$(50) + DNAme$_{Subset3}$(25), RNA-seq subset combined 86= RNA-seq$_{Subset1}$(20) + RNA-seq$_{Subset2}$(32) + RNA-seq$_{Subset3}$(34) and (miRNA-seq subset combined total = 85 miRNA-seq$_{Subset1}$(35) + miRNA-seq$_{Subset2}$(51).

\begin{table}
	\begin{scriptsize}
		\caption{Top 40 features identified by the QNN$_{256}$ model, including their gene names, P-values, feature importance (FI), Feature-specific based on their mean levels in LUAD and LUSC (i.e. Luad-Tumor, Luad-Normal, Lusc-Tumor and Lusc-Normal)}
		\begin{tabular}{ |p{2.8 cm} | p{1.1 cm}|p{0.8 cm} |p{1.4 cm}|p{1.9 cm}|}
			\hline
			Features	&	P-values	&	FI	&	Gene	&	Feat. Specific	\\
			\hline
			ENSG00000248144.4	&	0.8082	&	2.08	&	ADH1C	&	LUSC Normal	\\
			hsa-mir-632	&	0.1245	&	2.007	&	MIR632	&	LUAD Tumor	\\
			hsa-mir-3690-1	&	0.00013	&	1.768	&	MIR3690-1	&	LUSC Normal	\\
			cg09439093	&	6.21E-05	&	1.625	&	ECI2,	&	LUAD Tumor	\\
			hsa-mir-147b	&	9.10E-06	&	1.518	&	MIR147b	&	LUAD Tumor	\\
			ENSG00000214189.7	&	0.0736	&	1.487	&	ZNF788	&	LUSC Normal	\\
			hsa-mir-4442	&	0.2897	&	1.485	&	MIR4442	&	LUAD Tumor	\\
			ENSG00000084070.10	&	0.0014	&	1.450	&	SMAP2	&	LUSC Normal	\\
			ENSG00000115307.15	&	1.97E-22	&	1.449	&	AUP1	&	LUSC Tumor	\\
			cg07727233	&	1.39E-10	&	1.448	&	GSS	&	LUAD Normal	\\
			hsa-mir-4999	&	0.5929	&	1.421	&	MIR4999	&	LUSC Normal	\\
			cg09297514	&	0.5230	&	1.395	&	CAMKK2	&	LUAD Tumor	\\
			ENSG00000138109.9	&	0.0157	&	1.378	&	CYP2C9	&	LUSC Tumor	\\
			ENSG00000099810.17	&	0.2798	&	1.365	&	MTAP	&	LUSC Normal	\\
			cg19273074	&	0.06963	&	1.357	&	PDZRN4	&	LUSC Normal	\\
			ENSG00000134184.11	&	0.0026	&	1.301	&	GSTM1	&	LUSC Tumor	\\
			hsa-mir-138-2	&	0.03246	&	1.295	&	MIR138-2	&	LUSC Normal	\\
			cg19343088	&	0.7102	&	1.286	&	AIM1	&	LUSC Tumor	\\
			cg18441041	&	0.1053	&	1.283	&	CUX1	&	LUAD Tumor	\\
			hsa-mir-181a-2	&	1.12E-16	&	1.269	&	MIR181a-2	&	LUSC Normal	\\
			cg14336879	&	2.42E-23	&	1.260	&	SEZ6,PIPOX	&	LUAD Tumor	\\
			ENSG00000075945.11	&	0.0014	&	1.257	&	KIFAP3	&	LUSC Normal	\\
			ENSG00000117834.11	&	1.24E-20	&	1.239	&	SLC5A9	&	LUSC Normal	\\
			ENSG00000258947.5	&	0.00020	&	1.174	&	TUBB3	&	LUAD Tumor	\\
			hsa-mir-320b-2	&	7.68E-20	&	1.166	&	MIR320b-2	&	LUSC Tumor	\\
			ENSG00000104365.12	&	0.8046	&	1.163	&	IKBKB	&	LUSC Normal	\\
			cg23424800	&	0.00233	&	1.161	&	C8orf74	&	LUAD Normal	\\
			ENSG00000185201.15	&	6.68E-18	&	1.158	&	IFITM2	&	LUSC Normal	\\
			cg08703522	&	0.1377	&	1.151	&	WNT3	&	LUAD Tumor	\\
			ENSG00000136842.12	&	0.00057	&	1.134	&	TMOD1	&	LUSC Normal	\\
			ENSG00000177853.13	&	0.2412	&	1.099	&	ZNF518A	&	LUSC Normal	\\
			ENSG00000100359.19	&	0.00232	&	1.070	&	SGSM3	&	LUSC Normal	\\
			cg18070562	&	6.01E-20	&	1.042	&	ALOXE3	&	LUAD Normal	\\
			cg18575209	&	3.68E-06	&	1.022	&	ZNF586	&	LUSC Tumor	\\
			ENSG00000115896.14	&	0.000327	&	1.006	&	PLCL1	&	LUSC Normal	\\
			ENSG00000022556.14	&	0.24421	&	1.002	&	NLRP2	&	LUSC Normal	\\
			hsa-mir-193a	&	0.10296	&	0.989	&	MIR193a	&	LUSC Normal	\\
			cg12304599	&	0.19903	&	0.967	&	RERG	&	LUAD Tumor	\\
			cg24041269	&	2.72E-24	&	0.967	&	HLTF	&	LUAD Tumor	\\
			
			\hline
		\end{tabular}
	\end{scriptsize}
\end{table}

Thus, the integration of three types of molecular data (RNA-seq, DNAme, and miRNA-seq) can achieve better results for the patients’ subtype- prediction. MQML-LungSC achieves superior performance when integrating a variable number of modalities while including a smaller number of samples. Our results indicate that the Multi-Omic Quantum Machine Learning Lung Subtype Classification (MQML-LungSC) framework offers superior classification performance with smaller training datasets as significant and non-significant based on p-value, thus providing compelling empirical evidence for the potential future application of unconventional computing approaches in the biomedical sciences.
We observed that as we increased the number of features, we achieved superior performance in diagnostic classification and improved significance in distinguishing between significant and non-significant p-subtypes. Additionally, we identified the best features within the 32 and 64-feature datasets to highlight the top-hit multi-omic features for further investigations.
Fig 7. presents the hierarchical representation and clustering analysis of single omics and integrated multi-omic data. Heatmaps illustrate DNAme, RNA-seq, miRNA-seq, and important features patterns in LUAD subtype-II compared to LUSC subtype-I lung datasets. Hierarchical clustering organizes cell lines, with labels in purple (subtype-I) and yellow (subtype-II) highlighting LUSC and LUAD patients, respectively. (A) DNAme, (B) miRNA-seq, (C) RNA-seq, (D) Integrated (A, B, C) omics as Multi-Omic$_{256}$ features.
(E) Outcome of Top 32 molecular features of integrated Multi-Omic$_{256}$ features using QNN1$_{256}$ model. We discovered sev- eral novel features inDNA-me , miRNA-Seq and RNA-Seq which have a unique diagnostic potential to differentiate LUSC and LUAD subtypes.  Fig 8. presents a comparison of QNN models with different encoding features on integrated multi-omic data based on various classification metrics through a boxplot.

Fig (9-10) represents the performance evaluation of the QNN model using various feature sets. Fig 9.  depict the visualization of confusion matrices for training datasets with 256, 64, and 32 features, respectively. Similarly, Fig 10. displays confusion matrices for testing datasets with the same feature configurations. Additionally, Fig 9 illustrates the visualization of the Area Under the Curve (AUC-ROC) visualization for the QNN model trained on 256, 64, and 32 features, showcasing performance metrics across training and testing datasets.
Table VI presents the performance metrics of QNN models in distinguishing between LUAD and LUSC , using different numbers of features. The table compares precision, recall, F1-score, and accuracy for QNN models trained with 256, 64, and 32 features, respectively. Each metric is reported separately for both LUSC and LUAD classes, demonstrating the model's ability to classify between these lung dataset subtypes across varying feature dimensions.

\subsection{\textbf{Performance of Multi-Omic Model1:QNN-256 and Important Features Identification}}
MQML-$QNN_{256}$ Model achieves superior performance when integrating three modalities while including 915 samples. A Multi-Omic integration final Classification results with all visualizations with a dimension of 256 features shown in Fig. 9. In the integration of $QNN_{256}$, we integrated OMIC$_{DNA}$=85, OMIC$_{RNA}$=86, and OMIC$_{miRNA}$=85 features with 915 sample of subtype-I and subtype-II.  Similar parameter metrics results have also been reported that the MQML-QNN$_{256}$ model provided slightly better accuracy results in the training dataset than in testing results. As evident from Table VI and Table X, the MQML-QNN$_{256}$ provided the best performance for the diagnostic subtype-I and subtype-II Loss= (0.1475 and 0.2527), Accuracy= (0.95 and 0.90), AUC-ROC= (0.94 and 0.96) in training and testing respectively. 

\noindent \textbf{Research Process for Distinguishing LUAD and LUSC Subtypes for QNN$_{256}$ model:}}

We identified the 32 Top-hit features with significant and insignificant features by the proposed MQML-QNN$_{256}$ model1: ADH1C, MIR632, MIR3690-1 ECI2, RP3-400B16.4, MIR147b, ZNF788, MIR4442, SMAP2, AUP1, GSS, MIR4999, CAMKK2, CYP2C9, MTAP, PDZRN4, RP11-413B19.2, GSTM1, MIR138-2, MIR3174, AIM1, CUX1, MIR181a-2, SEZ6, PIPOX, KIFAP3, SLC5A9, TUBB3, MIR320b-2, IKBKB, C8orf74, RP1L1, IFITM2, WNT3, TMOD1 and many more.A detailed summary of the information on top-hit features is provided in Table VII. Identification of novel molecular features such as ADH1C, MIR632, and MIR3690-1 is crucial for advancing cancer research, particularly in distinguishing between different subtypes like LUAD and LUSC in lung subtypes. These features, identified through the MQML-QNN$_{256}$ model, offer significant insights into disease mechanisms and treatment responses. ADH1C  \cite{16}  MiR-632  \cite{17} \cite{18} \cite{19}, IFITM2  \cite{20}, \cite{21} GSTM1, TMOD1  \cite{22}, Wnt3 \cite{23}, miR-147b \cite{24}  \cite{25}, miR-4442 \cite{26} PDZRN4 \cite{27}, CUX1  \cite{28} \cite{29}.

Given a trained MQML-QNN model for classifying lung dataset subtypes (LUAD and LUSC), the most important features for classification can be identified by calculating feature importance scores from model weights and associating features with LUAD or LUSC based on their mean expression levels. 

\noindent (i) \textbf{Identify Significant Features:} Let \( \{X_i\} \) be the set of features used in the (MQML-LungSC) model for classifying LUAD and LUSC.

\noindent (ii) \textbf{Compute Feature Importance:} For each feature \( i \), calculate the feature importance score \( \text{Importance}_i \) as the average of absolute weights: 

$\text{Importance}_i = \frac{1}{n} \sum_{j=1}^n |w_{ij}|$

where \( w_{ij} \) is the weight for feature \( i \) in instance \( j \), and \( n \) is the number of instances.

\noindent (iii) \textbf{Calculate Mean Expression Levels:} Let \( \mu_i^{LUAD} \) and \( \mu_i^{LUSC} \) represent the mean expression levels of feature \( i \) in LUAD and LUSC samples, respectively:

$\mu_i^{LUAD} = \frac{1}{n_{LUAD}} \sum_{j \in \text{LUAD}} X_{ij}$ 

$\mu_i^{LUSC} = \frac{1}{n_{LUSC}} \sum_{j \in \text{LUSC}} X_{ij}$

where \( X_{ij} \) is the expression level of feature \( i \) in sample \( j \), and \( n_{LUAD} \) and \( n_{LUSC} \) are the numbers of LUAD and LUSC samples, respectively. 

(iv) \textbf{Determine Feature Association:} Classify feature \( i \) based on its mean expression levels:
\[
\text{Association}_i = 
\begin{cases}
\text{LUAD} & \text{if } \mu_i^{LUAD} > \mu_i^{LUSC} \\
\text{LUSC} & \text{otherwise}
\end{cases}
\] 
\noindent (iv) \textbf{Integrate Results:} Combine the feature importance, mean expression levels, and association results into a final table for features identification (BI):
\[
\text{BI}_i = \left( \text{Importance}_i, \mu_i^{LUAD}, \mu_i^{LUSC}, \text{Association}_i \right)
\]

By calculating the feature importance scores and comparing the mean expression levels of features in LUAD and LUSC, we can systematically identify the key features for classifying these subtypes. This thorough analysis allows for a clearer understanding of the differences between LUAD and LUSC, enhancing diagnostic and therapeutic strategies.

\begin{table}
\begin{scriptsize}
	\caption{List of 32 top-hit features and their P-values on using QNN-64 model2.}
	\begin{tabular}{ |p{3.2 cm} | p{2.5 cm}|p{2 cm}|}
		\hline
		Feature	&	Gene Name	&	P-Value	\\
		\hline
		cg14705695	&	TYR	&	0.2329	\\
		hsa-mir-323a	&	MIR323a	&	8.52E-10	\\
		hsa-mir-193b	&	MIR193b	&	2.32E-23	\\
		ENSG00000167004.11	&	PDIA3	&	0.00584	\\
		hsa-mir-6718	&	MIR6718	&	0.01123	\\
		cg02398045	&	HLTF,HLTFAS1	&	0.00509	\\
		hsa-mir-1258	&	MIR1258	&	0.8056	\\
		cg25016127	&	HRAS,LRRC56	&	0.4157	\\
		ENSG00000130032.14	&	PRRG3	&	1.48E-23	\\
		hsa-mir-3174	&	MIR3174	&	0.01899	\\
		ENSG00000138109.9	&	CYP2C9	&	0.01577	\\
		hsa-mir-519a-2	&	MIR519a-2	&	0.09878	\\
		hsa-mir-6815	&	MIR6815	&	0.49621	\\
		cg06808983	&	G6PC3	&	0.4518	\\
		ENSG00000258947.5	&	TUBB3	&	0.00020	\\
		hsa-mir-552	&	MIR552	&	3.86E-21	\\
		hsa-mir-182	&	MIR182	&	0.0068	\\
		ENSG00000177570.12	&	SAMD12	&	0.01211	\\
		ENSG00000166770.9	&	ZNF667-AS1	&	0.4329	\\
		cg01025842	&	FBXL8,TRADD	&	1.41E-05	\\
		ENSG00000114378.15	&	HYAL1	&	0.00010	\\
		hsa-mir-6868	&	MIR6868	&	0.0387	\\
		cg21035368	&	ERCC1	&	0.7402	\\
		ENSG00000143248.11	&	RGS5	&	1.85E-19	\\
		cg08842508	&	PABPC1P2	&	0.2452	\\
		ENSG00000169660.14	&	HEXDC	&	4.13E-05	\\
		cg02665570	&	LYPLAL1	&	0.01026	\\
		hsa-mir-4685	&	MIR4685	&	0.0471	\\
		hsa-mir-4665	&	MIR4665	&	1.41E-24	\\
		cg25208969	&	ARHGAP23	&	1.24E-09	\\
		cg06676119	&	PDE1A	&	0.09512	\\
		ENSG00000213949.7	&	ITGA1	&	4.22E-23	\\
		\hline
	\end{tabular}
\end{scriptsize}
\end{table}

\begin{figure}[!ht]
\centering 
\includegraphics[scale=0.23]{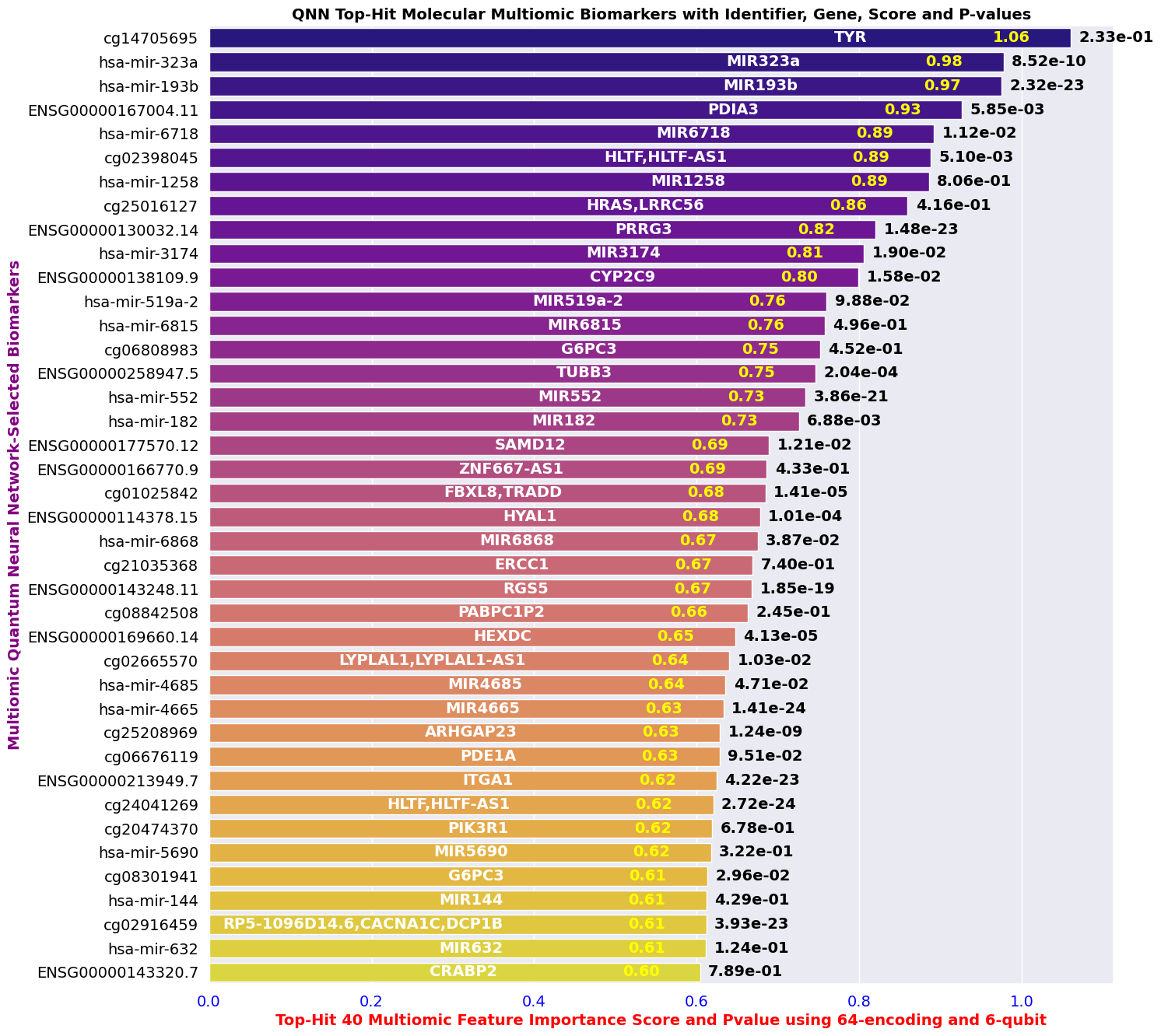} 
\caption{\textbf{Top-Hit Comparison features} Visualization of (b) 64 QNN model 2}
\end{figure}

\begin{figure}[!ht]
	\centering
	\includegraphics[scale=0.3]{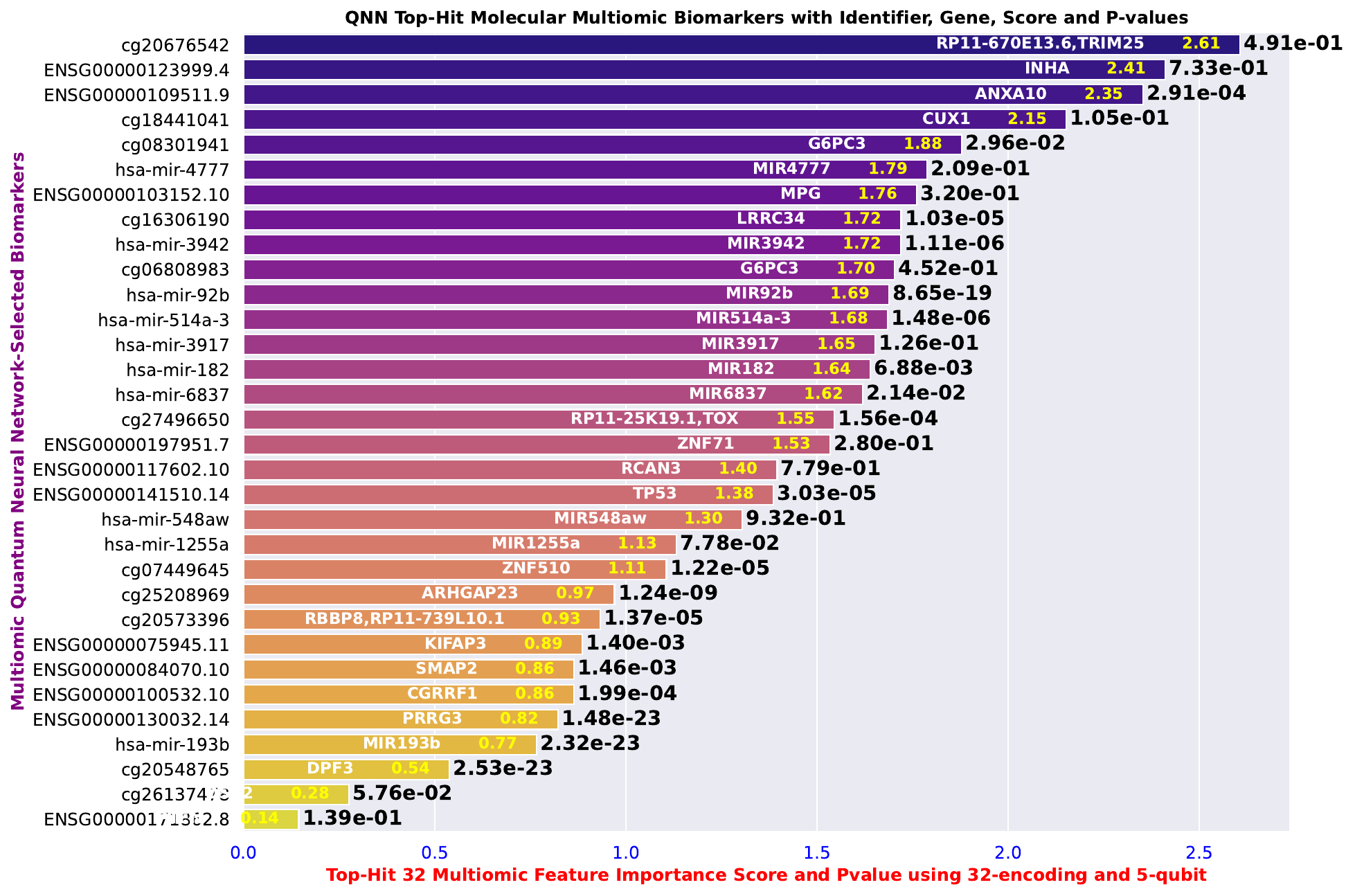}
	\caption{\textbf{Top-Hit Comparison features} Visualization of (c) 32 of QNN model}
\end{figure}

\begin{table}
	\begin{scriptsize}
		\caption{List of 32 top-hit features and their P-values on using QNN-32 model3.}
		\begin{tabular}{ |p{3.2 cm} | p{2.5 cm}|p{2 cm}|}
			\hline
			Feature	&	Gene Name	&	P-Value	\\
			\hline
			cg20676542	&	TRIM25	&	0.4906	\\
			ENSG00000123999.4	&	INHA	&	0.7333	\\
			ENSG00000109511.9	&	ANXA10	&	0.00029	\\
			cg18441041	&	CUX1	&	0.10537	\\
			cg08301941	&	G6PC3	&	0.02957	\\
			hsa-mir-4777	&	MIR4777	&	0.2086	\\
			ENSG00000103152.10	&	MPG	&	0.32002	\\
			cg16306190	&	LRRC34	&	1.03E-05	\\
			hsa-mir-3942	&	MIR3942	&	1.11E-06	\\
			cg06808983	&	G6PC3	&	0.4518	\\
			hsa-mir-92b	&	MIR92b	&	8.65E-19	\\
			hsa-mir-514a-3	&	MIR514a-3	&	1.48E-06	\\
			hsa-mir-3917	&	MIR3917	&	0.12600	\\
			hsa-mir-182	&	MIR182	&	0.0068	\\
			hsa-mir-6837	&	MIR6837	&	0.02140	\\
			cg27496650	&	TOX	&	0.000156	\\
			ENSG00000197951.7	&	ZNF71	&	0.2799	\\
			ENSG00000117602.10	&	RCAN3	&	0.7790	\\
			ENSG00000141510.14	&	TP53	&	3.03E-05	\\
			hsa-mir-548aw	&	MIR548aw	&	0.9316	\\
			hsa-mir-1255a	&	MIR1255a	&	0.0777	\\
			cg07449645	&	ZNF510	&	1.22E-05	\\
			cg25208969	&	ARHGAP23	&	1.24E-09	\\
			cg20573396	&	RBBP8	&	1.37E-05	\\
			ENSG00000075945.11	&	KIFAP3	&	0.00140	\\
			ENSG00000084070.10	&	SMAP2	&	0.0014	\\
			ENSG00000100532.10	&	CGRRF1	&	0.0001	\\
			ENSG00000130032.14	&	PRRG3	&	1.48E-23	\\
			hsa-mir-193b	&	MIR193b	&	2.32E-23	\\
			cg20548765	&	DPF3	&	2.53E-23	\\
			cg26137478	&	OSR2	&	0.0575	\\
			ENSG00000171862.8	&	PTEN	&	0.1390	\\
			\hline
		\end{tabular}
	\end{scriptsize}
\end{table}

\subsection{\textbf{Performance of Multi-Omic Model2: QNN-64 and Important Features Identification}}

In the integration of $QNN_{64}$, we integrated OMIC$_{DNA}$=22, OMIC$_{RNA}$=21, and OMIC$_{miRNA}$=21 features with 915 samples of subtype-I and subtype-II.  
A multi-omic integration final classification results with all visualizations with a dimension of 64 features are shown in Fig. 9-10.
The MQML-QNN$_{64}$ model provided slightly better accuracy results than MQML-QNN$_{32}$ in testing results. As evident from Table VI and Table X, the MQML-QNN$_{64}$ provided the best performance for the diagnostic subtype-I and subtype-II Loss= (0.21108 and 0.28467), Accuracy= (0.92 and 0.86), AUC-ROC= (0.97 and 0.92) in training and testing respectively. 
We identified the 32 Top-hit features with significant and insignificant features by the proposed MQML-QNN$_{64}$ model2: TYR, MIR323a, MIR193b, PDIA3, MIR6718, HLTF, HLTF-AS1, MIR1258, HRAS, LRRC56, PRRG3, MIR3174, CYP2C9, MIR519a-2, MIR6815, G6PC3, TUBB3, MIR552, MIR182, SAMD12, ZNF667-AS1, FBXL8, TRADD, HYAL1, MIR6868, ERCC1, RGS5, PABPC1P2, HEXDC, LYPLAL1, LYPLAL1-AS1, MIR4685, MIR4665, ARHGAP23, PDE1A, ITGA1 and many more. A detailed summary of top-hit features information is provided in Table VIII and Fig 11. 

\subsection{ \textbf{Performance of Multi-Omic Model3: QNN-32 and Important Features Identification}}
A multi-omic integration final classification results with all visualizations with a dimension of 32 features are shown in Fig. 9-10. In the integration of $QNN_{32}$, we integrated OMIC$_{DNA}$=11, OMIC$_{RNA}$=10, and OMIC$_{miRNA}$=11 features with 915 sample of subtype-I and subtype-II.  
The MQML-QNN$_{32}$ model provided slightly better confusion matrix subtype classification i.e. close to QNN$_{256}$ but slightly better than QNN$_{64}$ in testing results than training results. As evident from Table X, the MQML-QNN$_{32}$ provided the best performance for the diagnostic subtype-I and subtype-II. Loss= (0.21108 and 0.28467), Accuracy= (0.92 and 0.86), AUC-ROC= (0.97 and 0.92) in training and testing respectively. 
We identified the 32 Top-hit features with significant and insignificant features by the proposed MQML-QNN$_{32}$ model3: RP11-670E13.6, TRIM25
INHA, ANXA10, CUX1, G6PC3, MIR4777, MPG, LRRC34, MIR3942, G6PC3, MIR92b, MIR514a-3, MIR3917, MIR182, MIR6837, RP11-25K19.1,TOX, ZNF71, RCAN3, TP53, MIR548aw, MIR1255a, ZNF510, ARHGAP23, RBBP8,RP11-739L10.1, KIFAP3, SMAP2, CGRRF1, PRRG3, MIR193b, DPF3, OSR2, PTEN and many more. 

A detailed summary of the information on top-hit features is provided in Table IX and Fig 12.  Fig 14 shows the visualization representation with a violin plot, box plot, dot plot, and swarm plot of QNN-256 model top-12 molecular multi-omic features.

\subsection{ \textbf{Comparison with Classical Machine Learning Models with using Multi-Omic Datasets}}

In this study, with the aim of enhancing diagnostic lung subtype performance using GDC-TCGA datasets, a novel hybrid-QNN feature selection and diagnostic classification method was first proposed. Subsequently, four machine-learning models were employed to investigate the impact of feature selection methods on classification performance.
The classification performance of (MQML-LungSC) was compared to classical classifier algorithms as well as benchmark classification algorithms namely;  MLP, SVM, and LR as shown in Fig 13. Training and testing set performance of classical machine learning algorithms and proposed QNN on subtype diagnostic classification. Table X shows that (MQML-LungSC) outperforms all benchmark classification algorithms when trained on different dimensional representations of multiple modalities. This demonstrates the predictive power of the Hybrid Classical QNN learning architecture.

This study compares the performance of our proposed QNN models with classical machine learning models LR, MLP, and SVM—at different dimensions (256, 64, and 32) on a multi-omic dataset. Key metrics include accuracy, loss, and AUC.
The performance of QNN models, specifically $QNN_{256}$, $QNN_{64}$, and $QNN_{32}$, was compared against classical machine learning models such as  (LR), (MLP), and  (SVM) across different dimensions (256, 64, and 32). The evaluation metrics included Accuracy, Loss, and AUC for both training and testing phases.

\noindent \textbf{Performance of $QNN_{256}$:} $QNN_{256}$ achieved superior performance with a training accuracy of 0.95 and testing accuracy of 0.905. It also outperformed others with a lower loss of 0.1475 during training and 0.2527 during testing. The AUC was notably high at 0.99 for training and 0.96 for testing, indicating excellent classification capability. These metrics indicate that $QNN_{256}$ not only classifies data with high accuracy but also maintains a low error rate and a high capability to distinguish between classes. Compared to the best classical model, $SVM_{256}$, which achieved an accuracy of 0.93, loss of 0.2235, and AUC of 0.96, $QNN_{256}$ outperforms it in all aspects. This highlights the superiority of QNN in handling complex datasets with greater precision and efficiency.

\begin{table}
	\begin{scriptsize}
		\caption{Performance comparison with different classical machine learning models on multi-omic. }
		\begin{tabular}{ |p{1 cm} | p{1.1 cm}|p{1 cm}|  p{0.8 cm}|  p{1.2cm}| p{1.0 cm}| p{0.8 cm}|}
			\hline
			\multicolumn{1}{|c|}{} & \multicolumn{3}{|c|}{Training} & \multicolumn{3}{|c|}{Testing}  \\
			\hline
			
			Models	&	Accuracy	&	Loss	&	AUC	&	Accuracy	&	Loss	&	AUC	\\
			LR$_{256}$	&	0.748	&	0.5752	&	0.9	&	0.748	&	0.5869	&	0.89	\\
			LR$_{64}$	&	0.844	&	0.5032	&	0.91	&	0.797	&	0.5181	&	0.89	\\
			LR$_{32}$	&	0.76	&	0.5112	&	0.86	&	0.78	&	0.5121	&	0.86	\\
			\hline
			
			MLP$_{256}$	&	0.89	&	0.376	&	0.96	&	0.874	&	0.4101	&	0.95	\\
			MLP$_{64}$	&	0.78	&	0.5696	&	0.88	&	0.77	&	0.5787	&	0.84	\\
			MLP$_{32}$	&	0.75	&	0.5591	&	0.82	&	0.74	&	0.5715	&	0.81	\\
			\hline
			
			SVM$_{256}$	&	\textbf{0.93}	&	0.2235	&	0.96	&	0.87	&	0.2658	&	\textbf{0.96}	\\
			SVM$_{64}$	&	0.88	&	0.3131	&	0.93	&	0.84	&	\textbf{0.3545}	&	\textbf{0.92}	\\
			SVM$_{32}$	&	0.83	&	0.4215 	&	0.89	&	0.8	&	0.4704	&	0.87	\\
			\hline
			
			QNN$_{256}$	&	\textbf{0.95}	&	\textbf{0.1475}	&	\textbf{0.99}	&	\textbf{0.905}	&	\textbf{0.2527}	&	\textbf{0.96}	\\
			QNN$_{64}$	&	0.92	&	0.21108	&	\textbf{0.97}	&	\textbf{0.86}	&	0.3849	&	\textbf{0.92}	\\
			QNN$_{32}$	&	\textbf{0.88}	&	0.2846	&	\textbf{0.95}	&	\textbf{0.86}	&	\textbf{0.3863}&	\textbf{0.92}	\\
			\hline
		\end{tabular}
	\end{scriptsize}
\end{table}

\begin{figure}[!ht]
	\centering
	\includegraphics[scale=0.28]{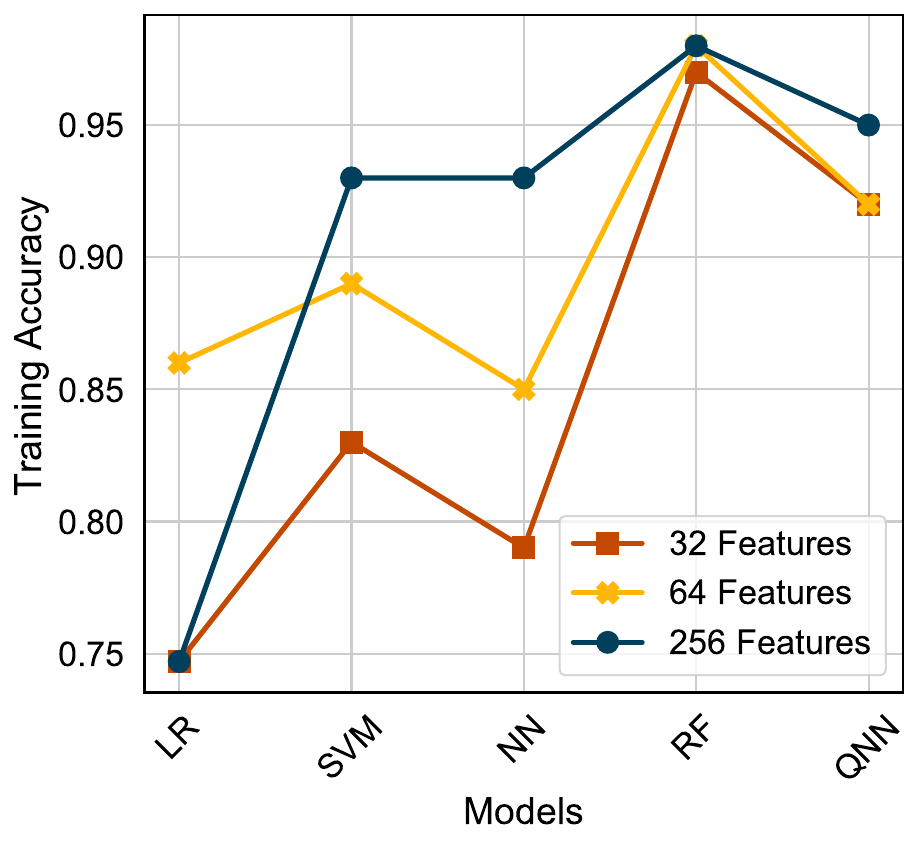} 
	\includegraphics[scale=0.28]{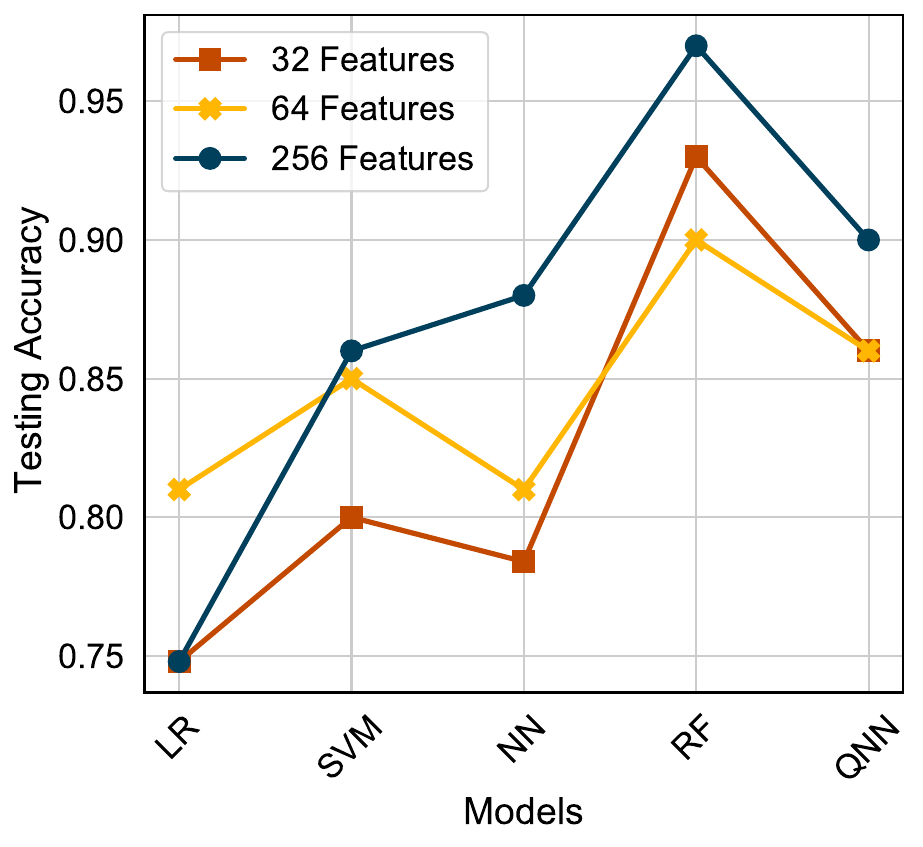}         
	\caption{\textbf{Training and testing set performance of classical machine learning algorithms and proposed QNN on subtype diagnostic classification}. }
\end{figure}

\begin{figure*}[!ht]
	\centering
	\begin{tabular}{c}
		\includegraphics[scale=0.28]{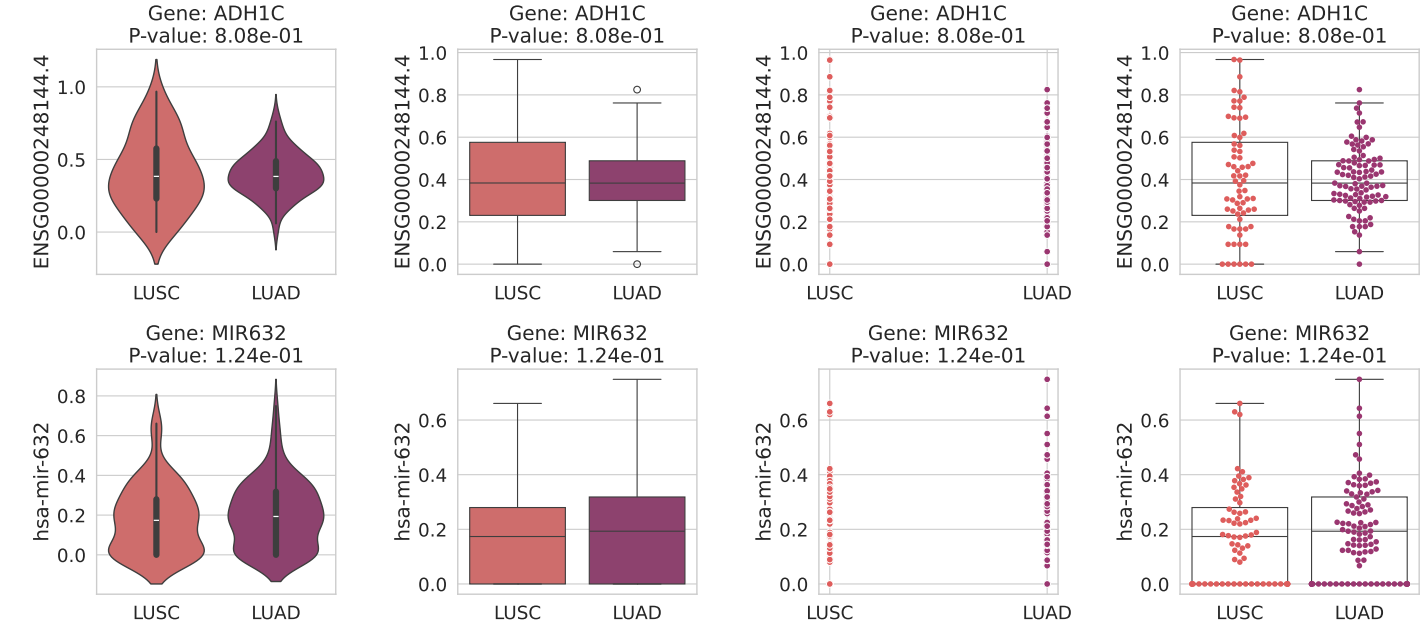}     
		\includegraphics[scale=0.28]{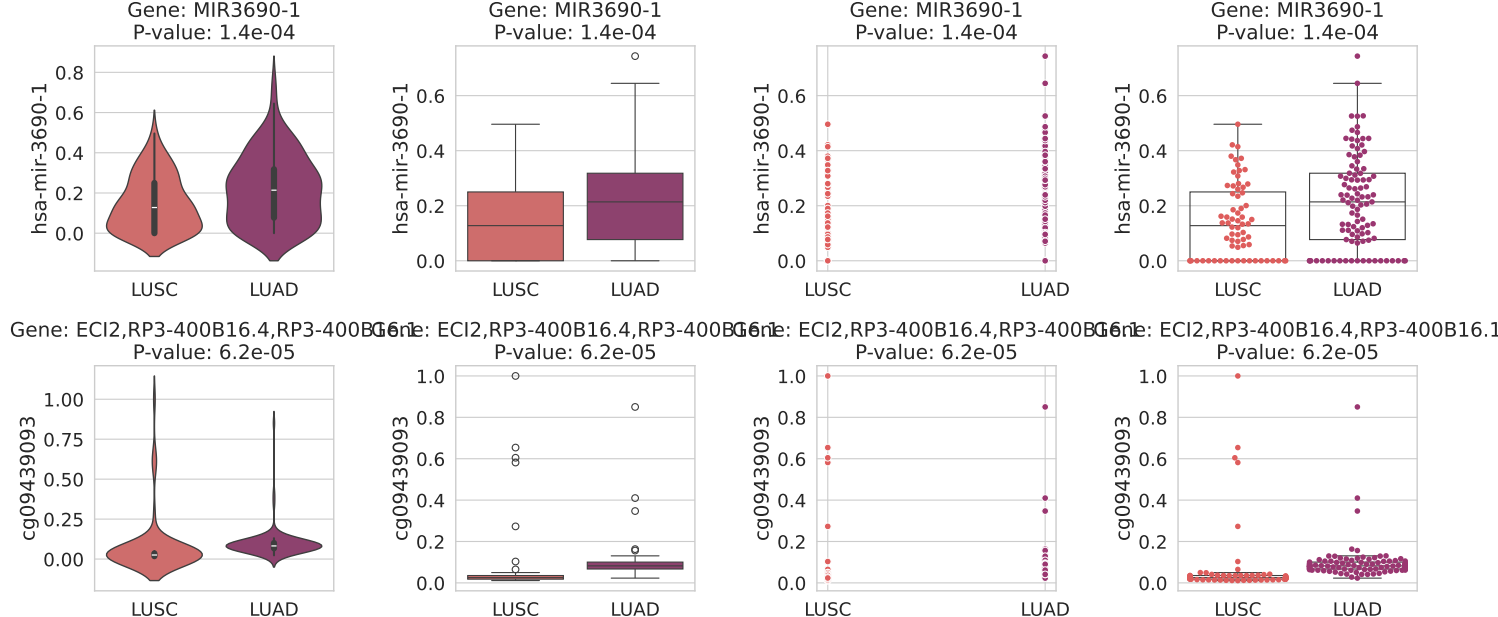} \\
		\includegraphics[scale=0.28]{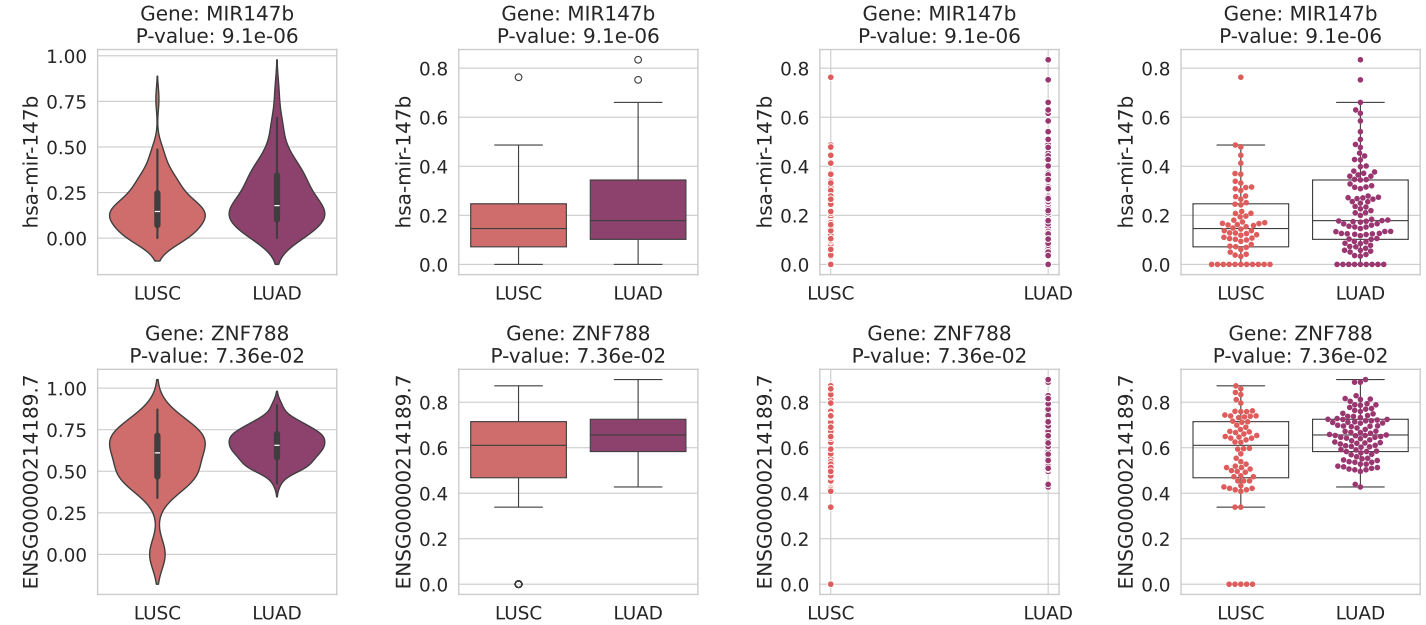} 
		\includegraphics[scale=0.28]{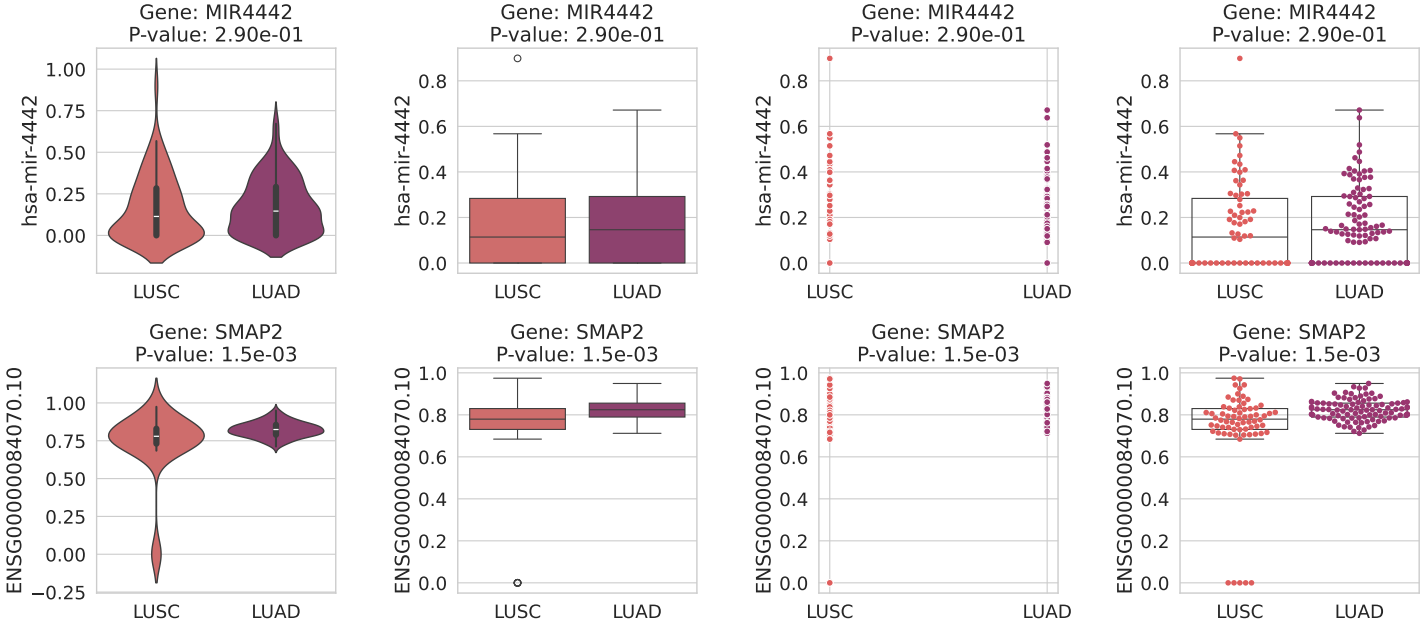} \\
		\includegraphics[scale=0.28]{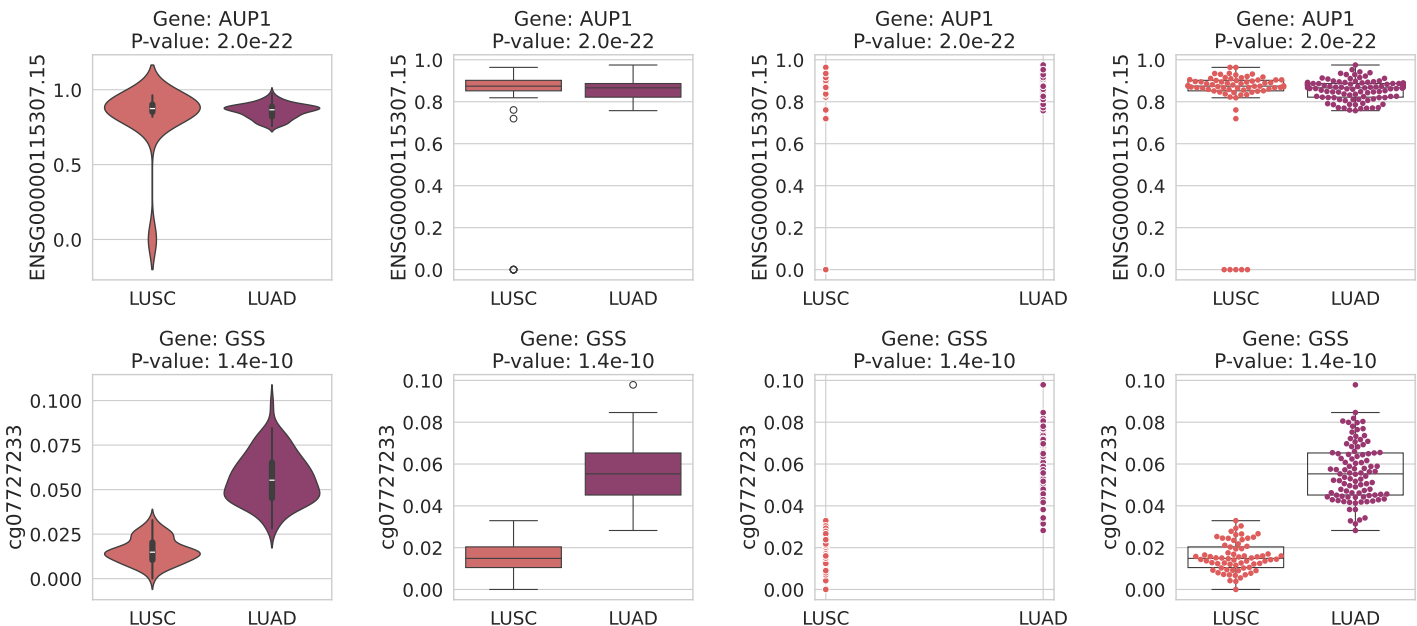} 
		\includegraphics[scale=0.28]{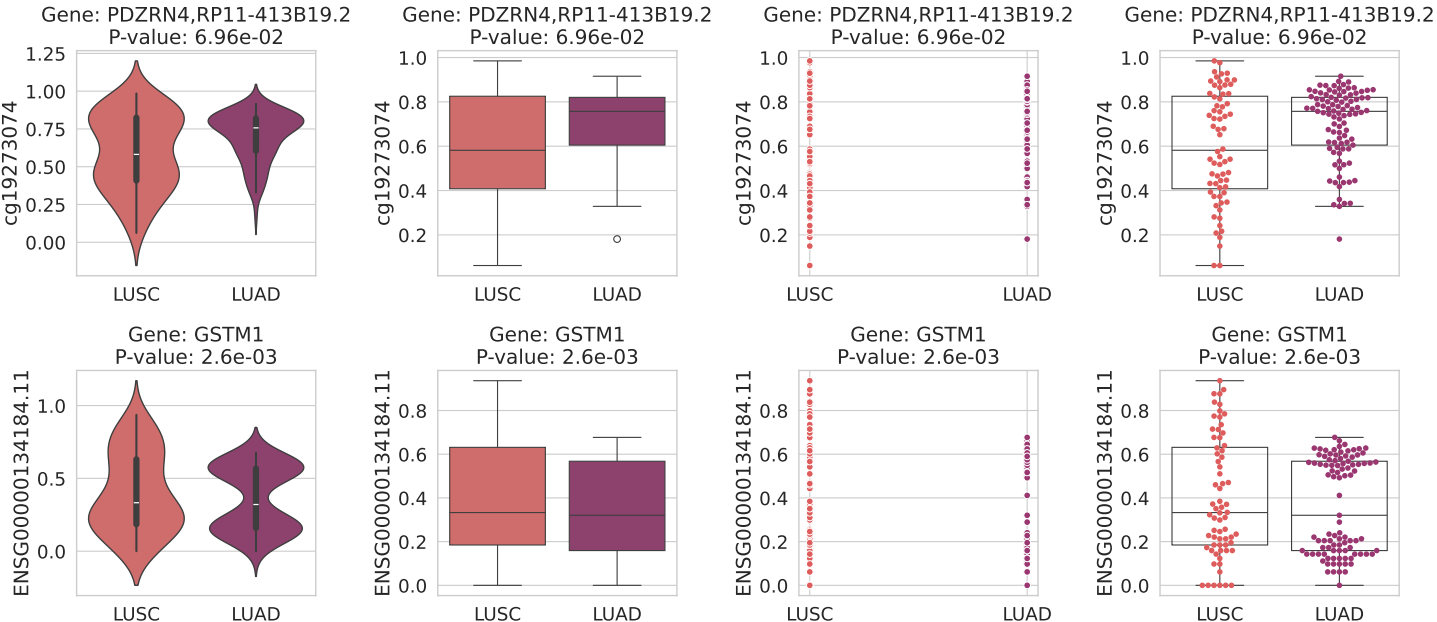} \\     
		\includegraphics[scale=0.28]{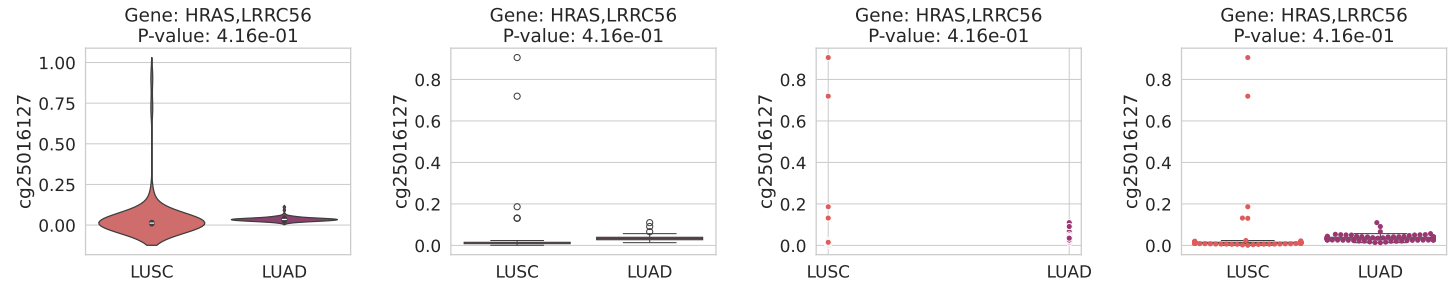}
		\includegraphics[scale=0.28]{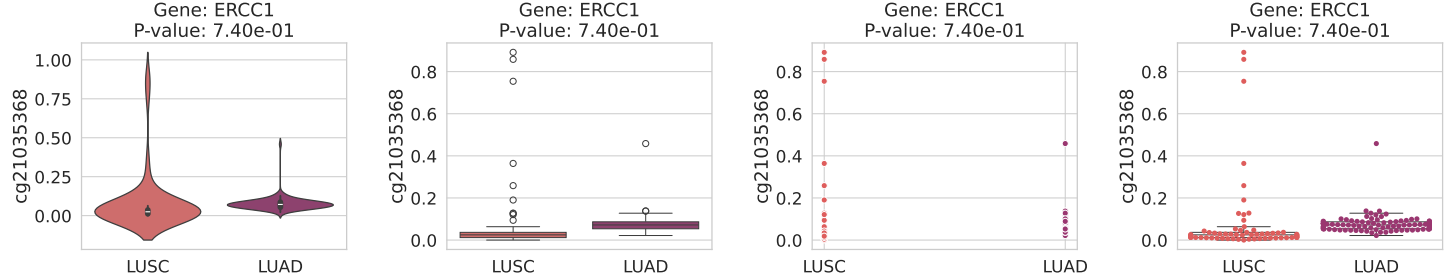}\\
		\includegraphics[scale=0.28]{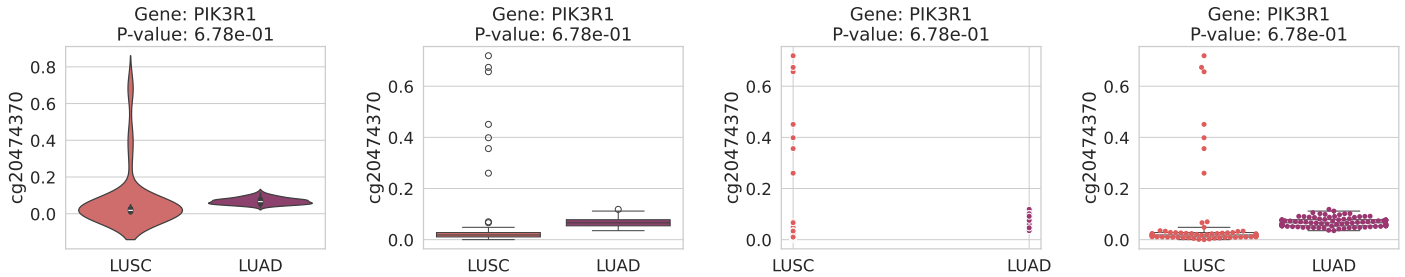}
		\includegraphics[scale=0.28]{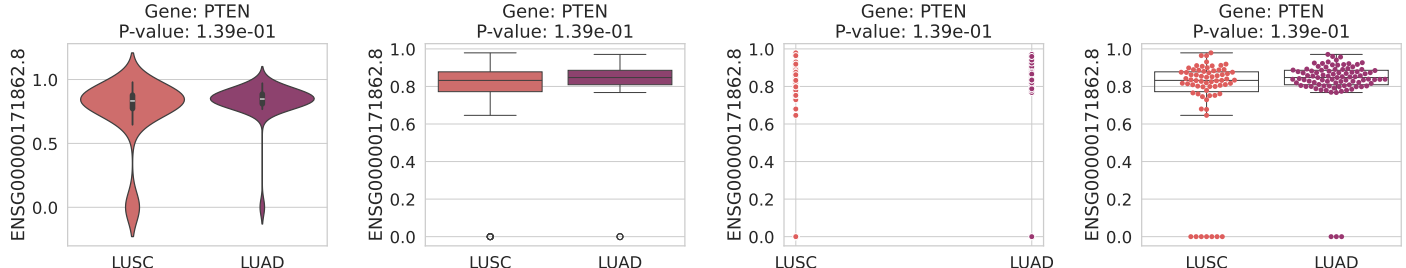}
		
	\end{tabular}
	\caption{\textbf{Visualization of Multi-Omic-Molecular features} Top-16 hit features visualization representation with a violin plot, box plot, dot plot, and swarm plot of QNN-256 model for diagnostic subtype-I (LUSC) and subtype-II(LUAD) lung classification with identifier name, gene/feature name and P-value }
\end{figure*}

\noindent \textbf{Performance of $QNN_{64}$: } showed strong performance with a training accuracy of 0.92 and a testing accuracy of 0.86. The training loss was 0.21108, while the testing loss was higher at 0.3849. The AUC values were also competitive, with 0.97 for training and 0.92 for testing.
These results are competitive with the $SVM_{64}$, which has an accuracy of 0.88, a loss of 0.3131, and an AUC of 0.93. While $SVM_{64}$ performs well, $QNN_{64}$ surpasses it in accuracy and AUC, although the loss is relatively higher. The comparison underscores that $QNN_{64}$ maintains a significant edge over classical models in terms of classification accuracy and the ability to distinguish between classes.

\noindent \textbf{Performance of $QNN_{32}$: } For the $QNN_{32}$ model, the performance metrics were still superior to many classical models. $QNN_{32}$ achieved an with a training accuracy of 0.88 and a testing accuracy of 0.86. The training loss was 0.2846, and the testing loss was 0.3863. The AUC values were 0.95 for training and 0.92 for testing, indicating reliable performance.
In comparison, the $SVM_{32}$ model obtained an accuracy of 0.83, a loss of 0.4215, and an AUC of 0.89. The results indicate that even at a lower dimensional configuration, $QNN_{32}$ outperforms traditional models, particularly in accuracy and AUC, though the loss metric indicates a slightly higher error rate compared to its higher-dimensional counterparts.

Across all configurations, the QNN consistently outperformed traditional classifiers. $QNN_{256}$ was the most effective, followed by $QNN_{64}$ and $QNN_{32}$. Traditional models like SVM and MLP showed competitive performance, especially $SVM_{256}$ with a high AUC of 0.96. However, the logistic regression models consistently lagged behind, with $LR_{64}$ achieving an accuracy of 0.748, a loss of 0.5752, and an AUC of 0.9, underscoring their limitations in complex data scenarios. The MLP models, while showing reasonable performance, were not able to match the accuracy and AUC of the QNN models, particularly in lower dimensions.

The dimension-wise analysis further reveals the effectiveness of quantum neural networks. $QNN_{256}$ emerged as the best performer with the highest accuracy (0.95), the lowest loss (0.1475), and the highest AUC (0.99). This model outshined all other models across every metric. Among the 64-dimensional models, $QNN_{64}$ was superior with an accuracy of 0.92 and an AUC of 0.97, although $SVM_{64}$ showed a slightly lower loss (0.3131 vs. 0.21108). In the 32-dimensional category, $QNN_{32}$ led with an accuracy of 0.88 and an AUC of 0.95, proving its robustness even at lower dimensions. These results affirm the potential of QNNs to provide high-performance classification across various dimensions, outperforming classical counterparts consistently.

\begin{figure*}[!ht]
	\centering
	\begin{tabular}{c}
		\includegraphics[scale=0.24]{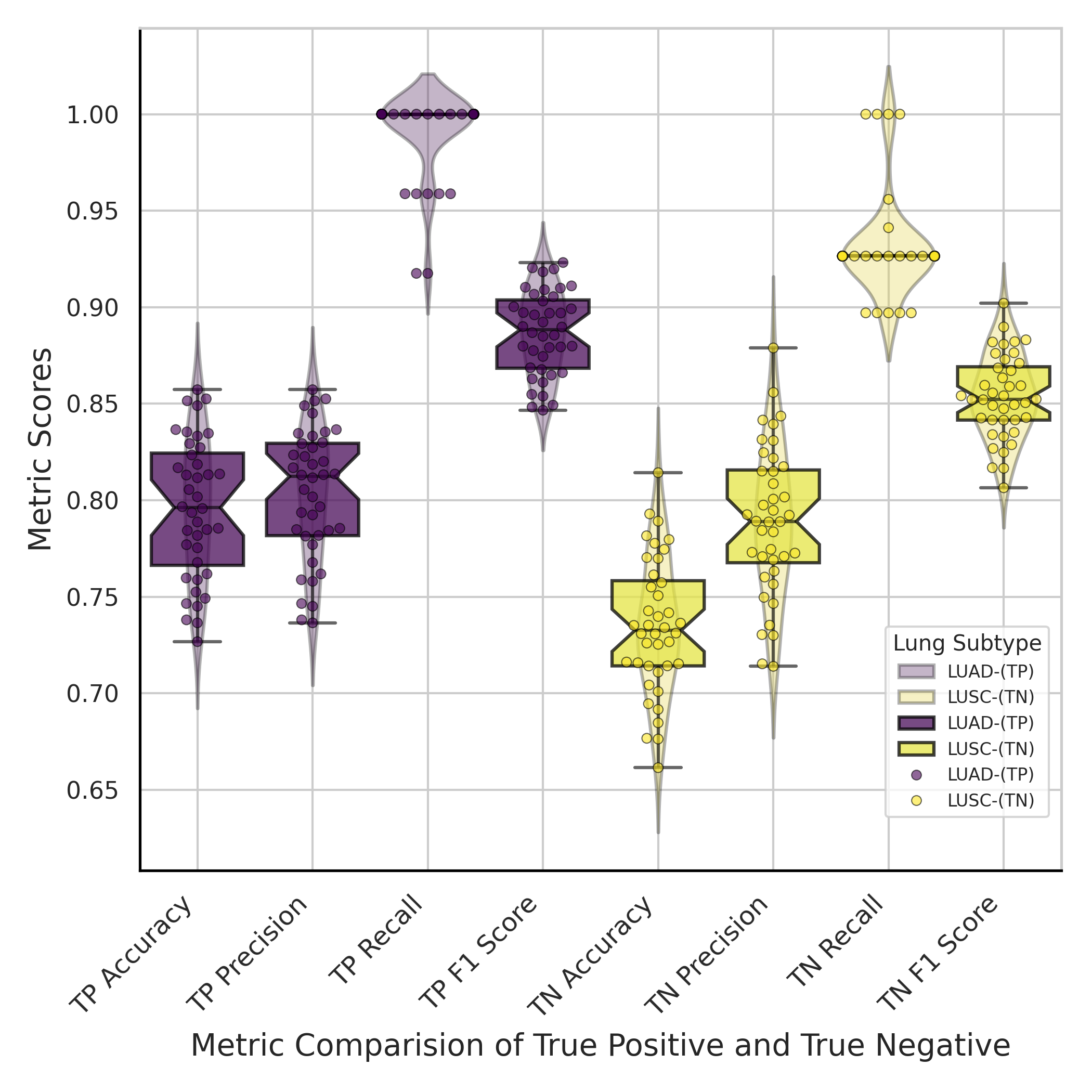}
		\includegraphics[scale=0.24]{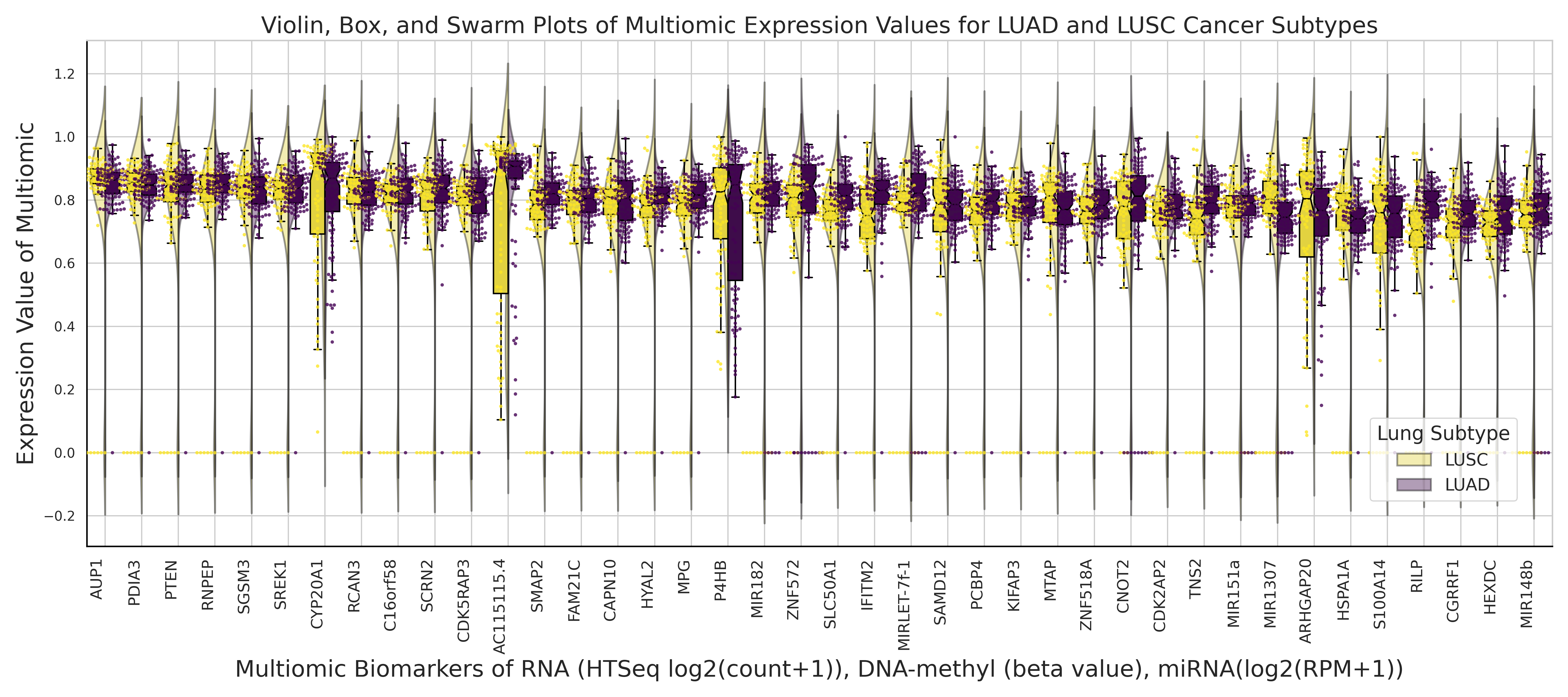} \\ 
		\includegraphics[scale=0.24]{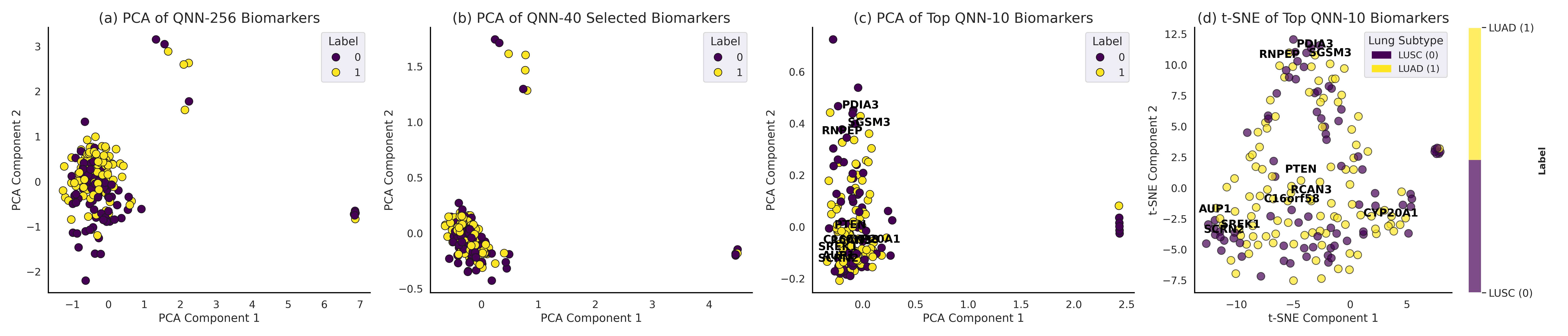}   \\
		\includegraphics[scale=0.24]{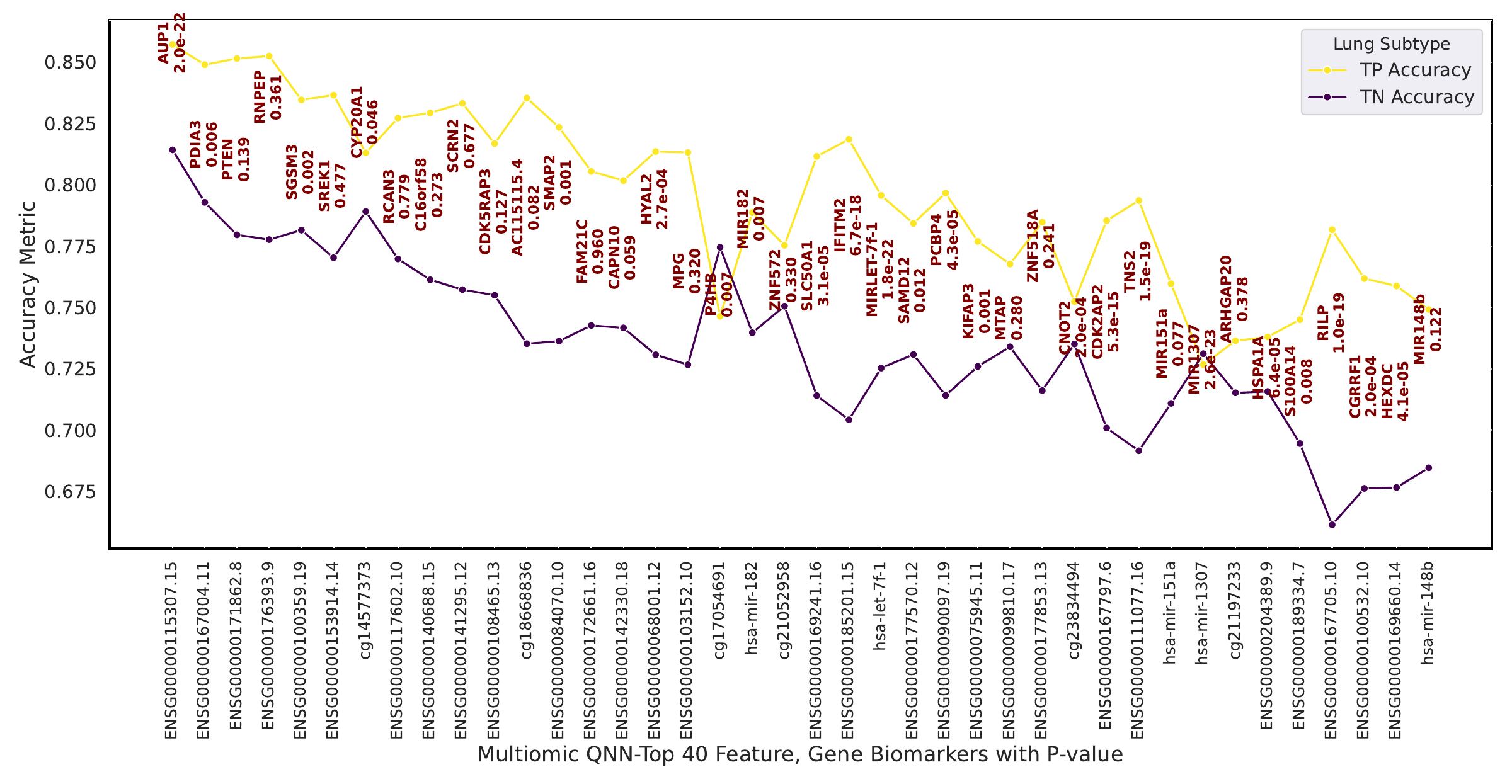}   
		\includegraphics[scale=0.2]{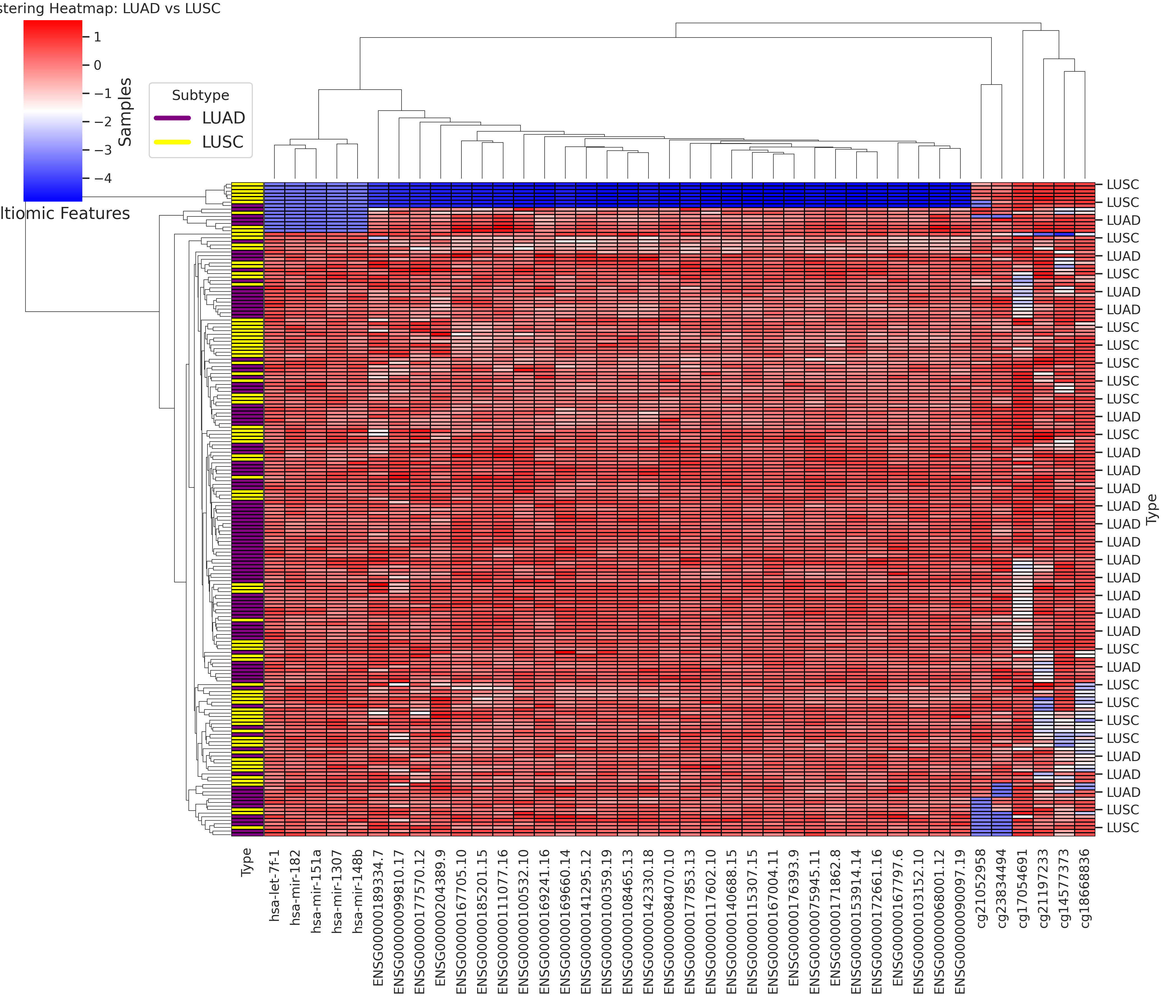}   
		
	\end{tabular}
	\caption{\textbf{Visualization of Multi-omic Molecular features with $QNN_{256}$ Predictions} Left Side: First row depicts the metric performance comparison of (LUAD)-TP and (LUSC)-TN, Right side: Visualization representation of 40 features via Violin, Box, and Swarm plots of Multi-omic expression values for LUAD and LUSC subtypes. In the Second Row: (a) Pca representation of $QNN_{256}$ model with 256 features before predictions, (b) After selecting 40-features for $QNN_{256}$ with PCA, (c) After selecting the top-best representation of 10-features for $QNN_{256}$ with PCA and (d) TSNE representation that separated the LUAD and LUSC group of subtypes very well.   }
\end{figure*}

\begin{figure*}[!ht]
	\centering
	\begin{tabular}{c}
		\includegraphics[scale=0.16]{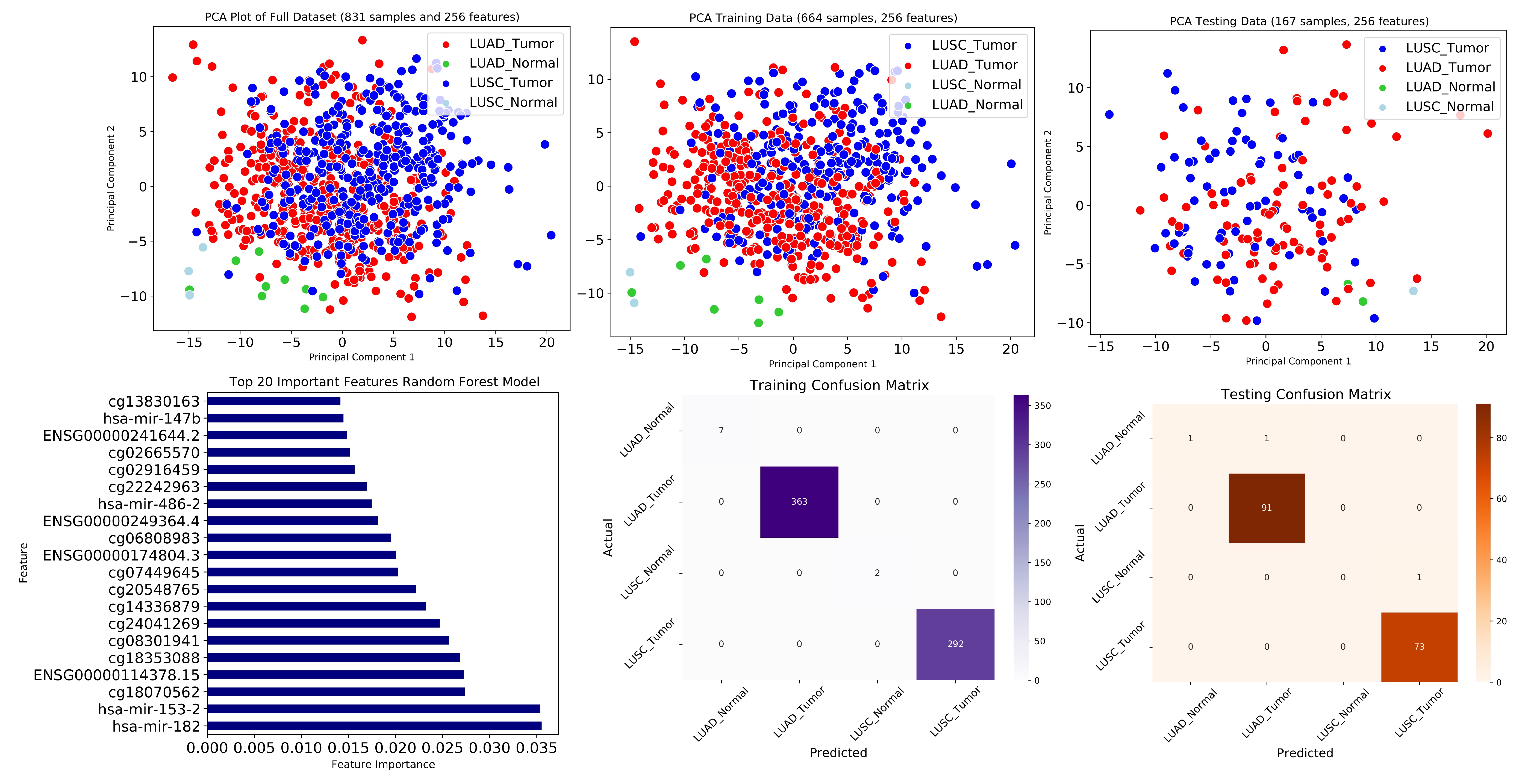}   
	\end{tabular}
	\caption{Visualization of 256-feature classification using a Random Forest model on multi-omic molecular features ($ML_{256}$ predictions). Top section: (a) The first row shows a PCA plot of the 256 features, displaying the training and testing datasets. (b) The second row presents the confusion matrix for both the training and testing datasets. (c) Bottom section: A visualization of the 20 best features identified by the Random Forest model, distinguishing between LUAD-tumor, LUAD-normal, LUSC-tumor, and LUSC-normal subtypes.}
\end{figure*}

\begin{figure*}[!ht]
	\centering
	\begin{tabular}{c}
		\includegraphics[scale=0.26]{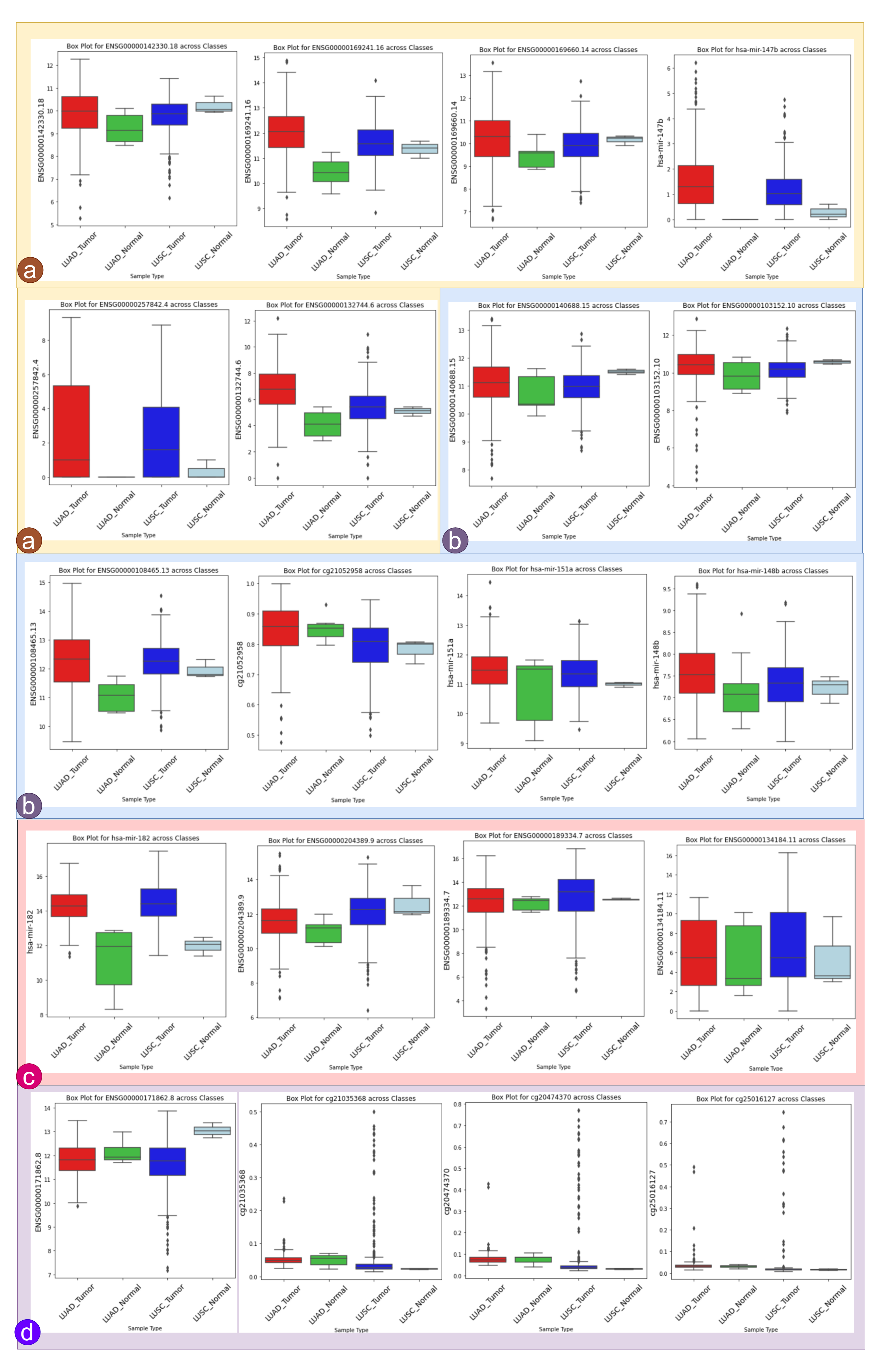}   
	\end{tabular}
	\caption{\textbf{Visualization of 21 out of 256 most and least significant genes based on p-values for LUAD and LUSC cohorts, distinguishing between tumor and normal samples.} (a) depicts the most significant LUAD-specific genes with a yellow outline, (b) depicts the least significant LUAD-specific genes with a blue outline, and (c) depicts the most significant LUSC-specific genes with a pink outline, (d) depicts the less significant LUSC-specific genes with a purple outline. Box plots show multiomic level values for LUAD-tumor, LUAD-normal, LUSC-tumor, and LUSC-normal subtypes.}
\end{figure*}

\begin{table*}
	\begin{scriptsize}
		\caption{List of 21 top-hit features and their P-values on using QNN-256 and RF model analysis}
		\begin{tabular}{ |p{2.8 cm} | p{1.8 cm}|p{1.4 cm}|p{2.1  cm} | p{1.2 cm}| p{2.2 cm}| }
			\hline		
			Identifier Names	&	Gene	&	Pvalue	&	Feature-specific	&	FI	&	Outcome	\\
			\hline
			ENSG00000140688.15	&	C16orf58	&	0.2731	&	LUAD-Tumor	&	0.1029	&	Less Significant	\\
			ENSG00000108465.13	&	CDK5RAP3	&	0.12661	&	LUAD-Tumor	&	0.1369	&	Less Significant	\\
			ENSG00000103152.10	&	MPG	&	0.3200	&	LUAD-Tumor	&	0.039	&	Less Significant	\\
			cg21052958	&	ZNF572	&	0.330273	&	LUAD-Tumor	&	0.1258	&	Less Significant	\\
			hsa-mir-151a	&	MIR151a	&	0.07682	&	LUAD-Tumor	&	0.097	&	Less Significant	\\
			hsa-mir-148b	&	MIR148b	&	0.12183	&	LUAD-Tumor	&	0.1413	&	Less Significant	\\
			hsa-mir-153-2	&	MIR153-2	&	0.1332	&	LUAD-Tumor	&	0.09582	&	Less Significant	\\
			ENSG00000142330.18	&	CAPN10	&	0.05855	&	LUAD-Tumor	&	0.1578	&	Most Significant	\\
			ENSG00000169241.16	&	SLC50A1	&	3.07E-05	&	LUAD-Tumor	&	0.0363	&	Most Significant	\\
			ENSG00000169660.14	&	HEXDC	&	4.13E-05	&	LUAD-Tumor	&	0.137	&	Most Significant	\\
			hsa-mir-147b	&	MIR147b	&	9.10E-06	&	LUAD-Tumor	&	1.51868	&	Most Significant	\\
			ENSG00000257842.4	&	NOVA1-AS1	&	0.0378	&	LUAD-Tumor	&	0.11304	&	Most Significant	\\
			ENSG00000132744.6	&	ACY3	&	5.57E-24	&	LUAD-Tumor	&	0.15902	&	Most Significant	\\
			hsa-mir-182	&	MIR182	&	0.00688	&	LUSC-Tumor	&	0.09	&	Most Significant	\\
			ENSG00000204389.9	&	HSPA1A	&	6.38E-05	&	LUSC-Tumor	&	0.0848	&	Most Significant	\\
			ENSG00000189334.7	&	S100A14	&	0.0077	&	LUSC-Tumor	&	0.10036	&	Most Significant	\\
			ENSG00000134184.11	&	GSTM1	&	0.00261	&	LUSC Tumor	&	1.3014	&	Most Significant	\\
			ENSG00000171862.8	&	PTEN	&	0.13904	&	LUSC Tumor	&	0.88	&	Less Significant	\\
			cg21035368	&	ERCC1	&	0.740247569	&	LUSC Tumor	&	0.67	&	Less Significant	\\
			cg20474370	&	PIK3R1	&	0.678445797	&	LUSC Tumor	&	0.62	&	Less Significant	\\
			cg25016127	&	HRAS	&	0.415771079	&	LUSC Tumor	&	0.86	&	Less Significant	\\
			
			\hline
		\end{tabular}
	\end{scriptsize}
\end{table*}

\section{DISCUSSION}
\noindent In this paper, we have proposed a (MQML-LungSC)-QNN framework based on quantum neural networks using three different dimensions. With the development of bio-informatics methods, researchers have been able to cover interesting features for LUAD and LUSC, and several relevant papers have been published. To detect biologically relevant markers, numerous studies have aimed to enhance traditional machine-learning algorithms or develop new ones for molecular feature discovery. However, few have employed overlapping machine learning or feature selection methods for cancer classification, best features identification, or gene expression analysis. The author \cite{6} the study proposed using overlapping traditional feature selection for cancer classification and biomarker discovery. The genes selected by the overlapping method were then validated using random forest. Gene expression analysis was subsequently performed to further investigate biological differences between LUAD and LUSC and identified 18 potentially novel features with high discriminating values between LUAD and LUSC.  The application of (MQML-LungSC) has been benchmarked on Lung subtype datasets from the GDC-TCGA, namely; LUAD and LUSC. Cancer is widely regarded as a highly heterogeneous disease however, (MQML-LungSC) was able to accurately classify lung dataset sub-types from integrated omic data.
(MQML-LungSC) identified the optimal combination of modalities which resulted in greater patient coverage while maintaining a state-of-the-art classification performance compared to its different dimensions of features as shown in Fig 4-5-6-7. We identified the top 32 features from model 1 $QNN_{256}$ features, another top 32 features from model 2 $QNN_{64}$ features, and the top 32 features from model 3 $QNN_{32}$ features using feature selection method and QNN diagnostic classifier that have potential to differentiate the subtype-I and subtype-II.
In comparing the performance of our proposed models, $QNN_{256}$, $QNN_{64}$, and $QNN_{32}$, it is evident that $QNN_{256}$ exhibits superior results across several metrics both in training and testing phases. 
Overall, the QNN models, particularly $QNN_{256}$, outperformed the classical models across most evaluation metrics. $QNN_{256}$ demonstrated the highest accuracy, lowest loss, and superior AUC, making it the most robust model in this comparison. QNN-64 and QNN-32 also showed strong results, particularly in terms of AUC, validating the effectiveness of quantum neural networks for multi-omic data classification tasks.

\subsection{\textbf{Simulation Analysis on $QNN_{256}$ Model to Find the Best Features Based on Performance Metrics}}
In this study, we aimed to identify key features (genes) that contribute to the accurate classification of lung dataset subtypes (LUAD and LUSC).
With the help of $QNN_{256}$ model, we employed a comprehensive approach to identifying the top 40 best features from multi-omic data by leveraging performance metrics of a predictive model. The process began by predicting test samples and evaluating the model's performance using a confusion matrix, from which we derived key metrics such as accuracy, precision, recall, and F1-score for true positive (TP) and true negative (TN) cases. We then calculated the mean feature values for TP and TN samples, followed by deviations from the overall mean. We computed aggregate scores by normalizing and combining TP and TN metrics to rank features. This led to identifying the top features with the highest deviations and best performance metrics. The final selection of the top 40 features was based on these aggregate scores, providing a robust set of best features for further analysis. This methodology ensures a balanced consideration of feature performance in distinguishing between LUAD and LUSC labels. Fig 15. illustrates the results of $QNN_{256}$ predictions on multi-omic-molecular features. The left side displays a performance comparison between LUAD-TP and LUSC-TN metrics, while the right side shows the visualization of 40 important features through violin, box, and swarm plots for LUAD and LUSC subtypes. The second row includes: (a) PCA representation of $QNN_{256}$ with 256 features before predictions; (b) PCA after selecting 40 features; (c) PCA with the top 10 features; and (d) t-SNE representation effectively separating LUAD and LUSC subtypes. In this analysis 40 features genes along with their performance metrics in distinguishing between LUAD (True Positive) and LUSC (True Negative) cases. Each gene is evaluated based on its p-value, accuracy, precision, recall, and F1-score for LUAD and LUSC classifications. This comprehensive overview highlights the effectiveness of each features, with emphasis on those showing both high statistical significance and strong classification performance, providing valuable insights into their potential utility in cancer diagnostics.
Table XI presents based on the above $QNN_{256}$ analysis and random forest (RF) model analysis, a total of 21 biomarkers were identified for LUAD and LUSC tumors. Among them, 7 biomarkers were found to be less significant in LUAD tumors, while 6 biomarkers were classified as most significant for LUAD. For LUSC tumors, 4 biomarkers were identified as most significant and 4 biomarkers were identified as less significant. This distribution underscores the differing levels of biomarker relevance in LUAD and LUSC, suggesting potential targets for further investigation in both cancer types.
In the LUAD dataset, the most significant genes include SLC50A1 (p-value = 3.07E-05), HEXDC (p-value = 4.13E-05), MIR147b (p-value = 9.10E-06), and ACY3 (p-value = 5.57E-24), all of which exhibit strong associations with LUAD tumors. On the other hand, less significant genes in LUAD include C16orf58 (p-value = 0.2731), CDK5RAP3 (p-value = 0.12661), and MPG (p-value = 0.320029). In the LUSC dataset, significant genes such as MIR182 (p-value = 0.00688) and GSTM1 (p-value = 0.002615542) stand out, while genes like PTEN (p-value = 0.13904) and ERCC1 (p-value = 0.740247569) are among the less significant.
In Fig. 16, the analysis involved combining RNA, DNA, and miRNA data, resulting in a dataset of 831 samples and 256 (selected features). Labels for the sample types (LUAD and LUSC, tumor and normal) were encoded into numeric classes for model training. Before applying machine learning, PCA was performed to visualize the variance in the dataset. The data was then split into training (80\%) and testing (20\%) sets, with 664 samples used for training and 167 for testing. A Random Forest classifier was trained using class-balanced weights, yielding a high testing accuracy of 98.8\%. The model performed well for LUAD and LUSC tumor classes, but had difficulty with LUSC normal due to the small sample size, reflected in zero recall for that class. Feature importance analysis revealed the top 20 most significant features, which were visualized. Multiple plots, including PCA scatter plots, confusion matrices, and box plots, were generated to assess model performance and the distribution of key features across the sample types.

\subsection{\textbf{Model Compilation Parameters}}

In this paper, we have created the function \texttt{MetricsCallback} to evaluate the performance of a machine learning model during training. The evaluation metrics include accuracy, precision, recall, and F1-score. These metrics are widely used in classification tasks to assess the model's ability to classify instances belonging to different classes correctly. All methods use the Adaptive Moment Estimation (Adam) optimizer; the learning rate is set to 0.01, and the batch size is 16. The experiments are conducted using the Pennylane quantum programming framework (version 0.28.0) in Python 2.8.0. The details of the classification metrics are given

\textbf{Accuracy}: It is used to evaluate how often the predictions match the actual labels and defined as the ratio of correctly predicted instances to the total instances:

\begin{equation}
	\text{Accuracy} = \frac{1}{N} \sum_{i=1}^{N} \mathbf{1}(\hat{y}_i = y_i) 
\end{equation}

\noindent where \(\mathbf{1}(\cdot)\) is the indicator function that returns 1 if the condition inside is true and 0 otherwise.

\textbf{Precision Score}: It is defined as the ratio of true positive predictions to the total number of positive predictions made by the model as
\begin{equation}
	\text{Precision} = \frac{{\sum_{i=1}^{N} TP_i}}{{\sum_{i=1}^{N} (TP_i + FP_i)}} 
\end{equation}
where \( TP_i \) represents the number of true positive predictions for class \( i \) and \( FP_i \) represents the number of false positive predictions for class \( i \).

\textbf{Recall Score}: It is also known as sensitivity, and measures the ability of the model to correctly identify positive instances out of all actual positive instances. Mathematically, it is represented as:
\begin{equation} \text{Recall} = \frac{{\sum_{i=1}^{N} TP_i}}{{\sum_{i=1}^{N} (TP_i + FN_i)}}
\end{equation}
where \( FN_i \) represents the number of false negative predictions for class \( i \).

\textbf{F1-score}: It is the harmonic mean of precision and recall, providing a balance between the two metrics. It is calculated using the formula:
\begin{equation} F_1 = 2 \times \frac{{\text{Precision} \times \text{Recall}}}{{\text{Precision} + \text{Recall}}}  \end{equation}

\section{Conclusion}
\noindent 
To the best of our knowledge, this study is the first to investigate the difference between LUSC and LUAD using the GDC-TCGA data within a hybrid classical-quantum classification model. In the classical part of the study, a feature selection method was used to determine the best and unique subset of multi-omic molecular features using the GDC-TCGA dataset.  The proposed model involves applying a QNN with classical dense layers for diagnosing lung subtype-I and subtype-II. This model's performance was compared with various classical machine learning classifiers, including Logistic Regression, Support Vector Machine, and Random Forest. Bench-marking with Quantum neural network ($QNN_{256}$), and the comparison included all existing models using 256, 64 and 32 features of integrated multi-omics data. The empirical results demonstrate that Quantum Neural Networks, particularly $QNN_{256}$, offer significant improvements over classical machine learning models. The superior performance in terms of accuracy, loss, and AUC suggests that QNNs are highly effective in handling complex data and providing accurate classifications. The $QNN_{256}$ model, with its exceptional metrics, stands out as the most promising, indicating the potential of quantum-based approaches to revolutionize machine learning. The overall comparison highlights the potential of quantum machine learning to surpass traditional methods, providing a compelling case for further exploration and development in this field. The dimension-wise analysis reaffirms the robustness and efficacy of QNNs, particularly at higher dimensions, where they consistently outperform classical models across all metrics. We employed two approaches to identify best features. First, we trained $QNN_{256}$, $QNN_{64}$, and $QNN_{32}$ models separately and determined the top 32 features for each model based on the QNN model weights. Also, we visualized the top-12 hit features using various plots. This included violin, box, dot, and swarm plots from the $QNN_{256}$ model for diagnostic classification of lung subtypes (LUSC and LUAD). Each plot displayed the identifier name, gene/feature name, and P-value for effective comparison.
In the second approach, the $QNN_{256}$ model was used to predict labels for 183 test samples, followed by evaluation using confusion matrices to derive key metrics like accuracy, precision, recall, and F1-score for true positive (TP) and true negative (TN) cases. We calculated mean feature values and deviations for TP and TN samples, then ranked features based on normalized aggregate scores. The top 40 features were selected from these scores, ensuring a comprehensive assessment of feature performance for distinguishing LUAD and LUSC subtypes. In this research, we distinguished between LUAD and LUSC subtypes by integrating feature importance with mean expression levels. First, we extracted feature importance from the model by averaging the absolute weights of each feature. Next, we calculated the mean expression levels of features in LUAD and LUSC samples using the equations.

\textbf{In the future,} we aim to enhance our work by focusing on deeper biological insights as well as quantum. In \textbf{Phase 1,} we will identify LUAD-specific and LUSC-specific genes by first applying a p-value t-test between tumor and normal samples, followed by differential analysis to further distinguish LUAD-specific genes and LUSC-specific genes. This will help identify additional biomarkers with potential prognostic value for cancer-specific between two subtypes. 
\textbf{Phase 2} will involve developing a quantum classifier to distinguish between tumor and normal samples, as well as between LUAD type-1 and LUSC type-2 tumors. This will classify the four sample groups and identify the top hit biomarkers specific to each cancer subtype. This approach will also highlight the significant p-values for LUAD- and LUSC-specific genes, enabling a clearer distinction between the two cancer types. Additionally, we will demonstrate the quantum advantage in a biologically meaningful context. Also, we plan to use these biomarkers for survival analysis to identify high-risk and low-risk clinical patients. We will also explore biological implications through Gene Ontology (GO) analysis and Kaplan-Meier survival curves. Additionally, we aim to compare correlations between classical and quantum analysis methods to understand their effectiveness in cancer research.
\subsection*{Data availability}
\noindent All data generated and/or analyzed during the current study are included in this article. The data in this paper were obtained from The Cancer Genome Atlas (GDC-TCGA) Research Network.https://www.cancer.gov/tcga)
The underlying code for this study is not publicly available but may be made available to qualified researchers on reasonable request from the corresponding author.

Single Omics: (1) DNA-Methylation (2) MicroRNA-seq (3) RNA-seq
Source name and link: UCSC XENA GDC-TCGA Dataset
UCSC Xena: GDC Xena Hub. The hub hosts v18.0 release from the GDC Data Portal. 

Open and Select OMICs for Download:
Github Link download: https://xenabrowser.net/datapages/

(1) DNA: Illumina Human Methylation 450 (n=503) GDC Hub, unit= beta value
Download: 
(2) miRNA-seq: stem-loop expression miRNA Expression Quantification (n=564) GDC Hub, unit= log2(RPM+1)
(3) RNA-seq: gene expression RNAseq, HTSeq - Counts (n=585) GDC Hub, unit= log2(count+1)
(4) Phenotype and clinical data
Download Phenotype clinical: TCGA-LUAD GDC phenotype
Download Survival data: TCGA-LUAD survival

\section*{Conflict of Interest}
\noindent The authors declare that they have no conflicts of interest.

\section*{Acknowledgements}
\noindent We would like to acknowledge the financial support of Purdue Quantum Science and Engineering Institute and U.S. Department of Energy (Office of Basic Energy Sciences) under Award No. DE-SC0019215.

\section*{Author Contributions}
S.K. and H.G. conceived and supervised the project.  M.K.S and A.S.B. designed the framework, and performed the simulations.  M.K.S. wrote the initial manuscript and contributed to the development and visualization. The Data feature engineering phase was conducted by M.I. and M.K.S. Resources and Funding acquisition by S.K. All authors contributed to analyzing the results and  finalizing the manuscript.

\section*{Data and software Availability}
The authors confirm that the data supporting the findings of this study are available from the corresponding author, upon reasonable request.

\bibliography{arxiv}

	
\end{document}